\documentclass{article}
\newcommand{\emotionprobepreprint}{}

\makeatletter
\AddToHook{package/iclr2027_conference/after}{%
  \iclrfinalcopy
  \renewcommand{\@maketitle}{%
    \vbox{\hsize\textwidth
      \lhead{Preprint. Under review.}%
      {\LARGE\scshape\@title\par}%
      \hbox to\textwidth{\hfil
        \begin{tabular}[t]{c}
          \rule{\z@}{24pt}\@author
        \end{tabular}%
        \hfil}%
      \vskip 0.3in minus 0.1in}}%
}
\makeatother

\ifdefined\emotionprobepreprint
\else
\documentclass{article} % For LaTeX2e
\fi
\usepackage{iclr2027_conference,times}

\usepackage{amsmath,amsfonts,bm}

\def\eqref#1{equation~\ref{#1}}
\def\1{\bm{1}}

\DeclareMathAlphabet{\mathsfit}{\encodingdefault}{\sfdefault}{m}{sl}
\SetMathAlphabet{\mathsfit}{bold}{\encodingdefault}{\sfdefault}{bx}{n}

\usepackage{hyperref}
\usepackage{url}
\usepackage{booktabs}
\usepackage{array}
\usepackage{multirow}
\usepackage{graphicx}
\usepackage{tcolorbox}

\title{Do Emotion Concepts Generalize Across Sources, Modalities, and Architectures in Vision-Language Models?}

\author{%
\textbf{Bohao Xing\textsuperscript{1}, Xin Liu\textsuperscript{2,1}\thanks{Corresponding author.},
Kaishen Yuan\textsuperscript{3}, Deng Li\textsuperscript{1}, Rong Gao\textsuperscript{1},} \\
\textbf{Guoying Zhao\textsuperscript{4,5}, Xiaolan Fu\textsuperscript{2},
Heikki K\"alvi\"ainen\textsuperscript{1,6}} \\[3pt]
\textsuperscript{1}Lappeenranta-Lahti University of Technology LUT \\
\textsuperscript{2}Shanghai Jiao Tong University \\
\textsuperscript{3}The Hong Kong University of Science and Technology (Guangzhou) \\
\textsuperscript{4}University of Oulu \quad
\textsuperscript{5}ELLIS Institute Finland \quad
\textsuperscript{6}Brno University of Technology \\[3pt]
\texttt{bohao.xing@lut.fi} \quad \texttt{linuxsino@gmail.com}
}

\begin{document}

\maketitle

\begin{abstract}
Recent studies suggest that large language models encode emotion concepts as structured internal representations, but most existing work focuses on text and a single architecture. Therefore, we ask, do emotion concepts generalize across sources, modalities, and architectures in vision--language models (VLMs)? To address this, we construct CMES (Cross-Modal Emotion Stimuli), a multi-source collection of emotion-conditioned stories, real facial expressions, synthetic portraits, and synthetic emotion-evoking scenes. For each stimulus source, we extract a separate set of six Ekman emotion vectors from each of three VLMs. We report four main findings 
as follows:
1) Image-derived emotion vectors form a low-dimensional geometry similar to that of text-derived vectors. Valence is relatively stable across sources, while arousal varies more. 2) Text- and image-derived emotion vectors have modest cosine similarity but still show held-out cross-modal correspondence. Text-derived vectors can also steer image interpretation. 3) Cross-architecture correspondence remains even when native cosine is near zero. Transformations estimated from generic ImageNet activations recover both correspondence and causal transfer without using the six emotion vectors or their labels. 4) After aligning representations across architectures, we construct a shared emotion subspace that preserves affective geometry and selective steering effects. The corresponding consensus emotion vectors also generalize to a held-out fourth architecture at two model sizes. These results suggest that emotion representations can share relational structure and causal effects across sources, modalities, and architectures, even when individual vector directions differ.
\end{abstract}

\section{Introduction}

As large language models (LLMs) develop, more studies are focusing on emotion understanding~\citep{cheng2024emotion,lian2025affectgpt,xing2026emo}. Meanwhile, recent interpretability work suggests that LLMs encode emotion concepts as structured internal representations. Using text-based stimuli, prior work has identified emotion vectors whose low-dimensional organization resembles human affective structure~\citep{anthropic2026emotion,reichman2026emotions,wang2025circuits,ben2026where}. Individual neurons and attention heads also carry emotion-related information~\citep{wang2025circuits,zheng2026neurons}. In addition, interventions along emotion vectors can change emotional judgments and generated behavior~\citep{zou2023repe,anthropic2026emotion,reichman2026emotions}. Recent work also traces emotion-related information flow within individual VLMs~\citep{zhang2026interpreting}. However, most existing evidence is based on text and a single model. It remains unclear how emotion representations change across stimulus sources and modalities, and whether their geometry and causal effects are preserved across architectures.

Researchers have also studied whether steering vectors can be transferred across models. Activation steering can modify refusal behavior and instruction following within individual models~\citep{arditi2024refusal,lee2025programming,stolfo2025improving}. More recent studies transfer steering vectors for concepts and behavioral attributes across independently trained text LLMs through learned transformations~\citep{huang2025cross,chen2026transferring,agarwal2026cross}. However, these results leave open how related emotion concepts are organized as a group across text and image sources, and whether their relational geometry and causal effects generalize across architectures. Further related work is discussed in Appendix~\ref{app:related-work}.

\begin{figure*}[t]
\centering
\begin{minipage}[t]{0.418\textwidth}
    \centering
    \includegraphics[width=\linewidth]{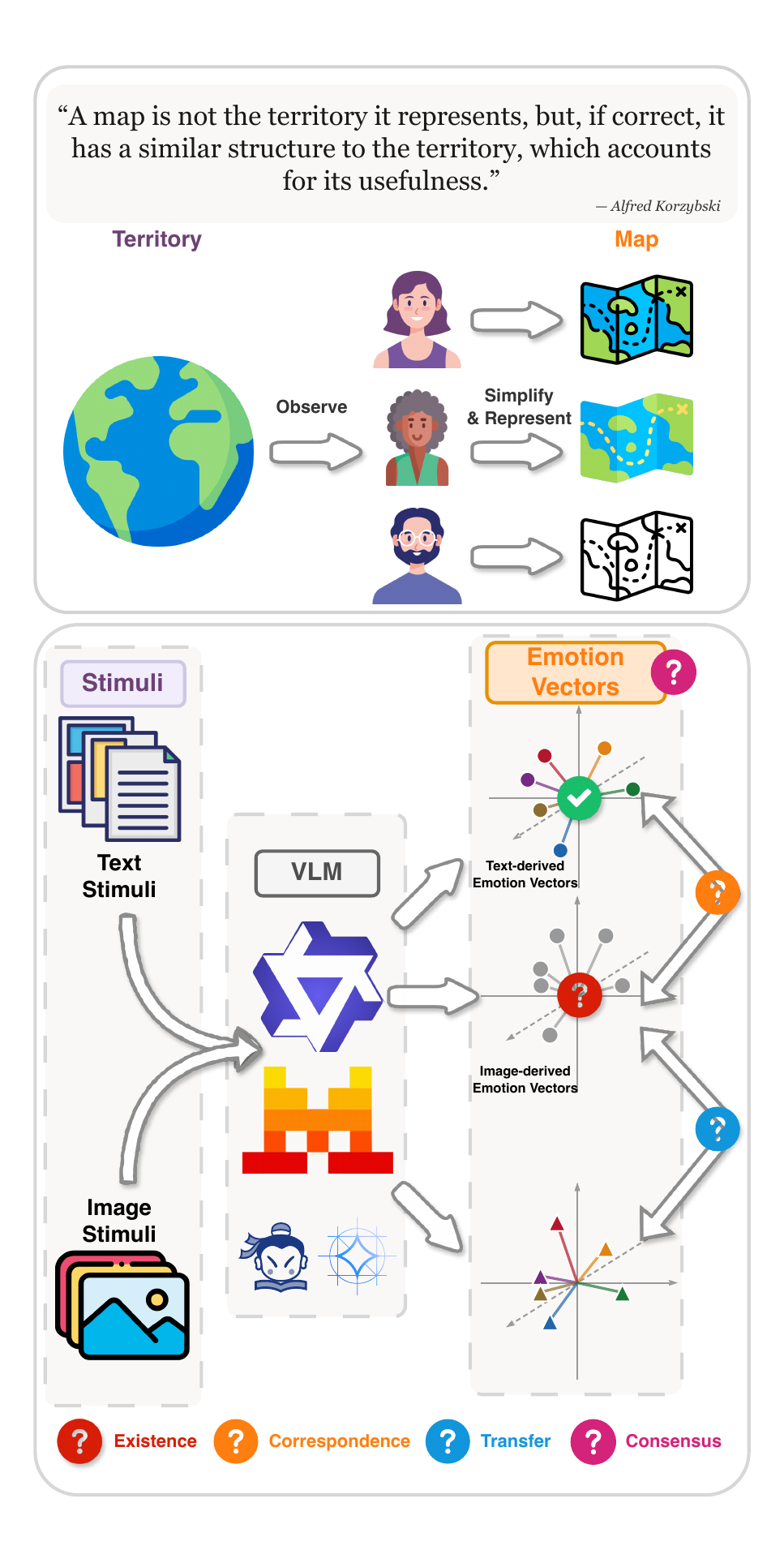}
\end{minipage}
\hspace{-2em}
\begin{minipage}[t]{0.518\textwidth}
    \centering
    \includegraphics[width=\linewidth]{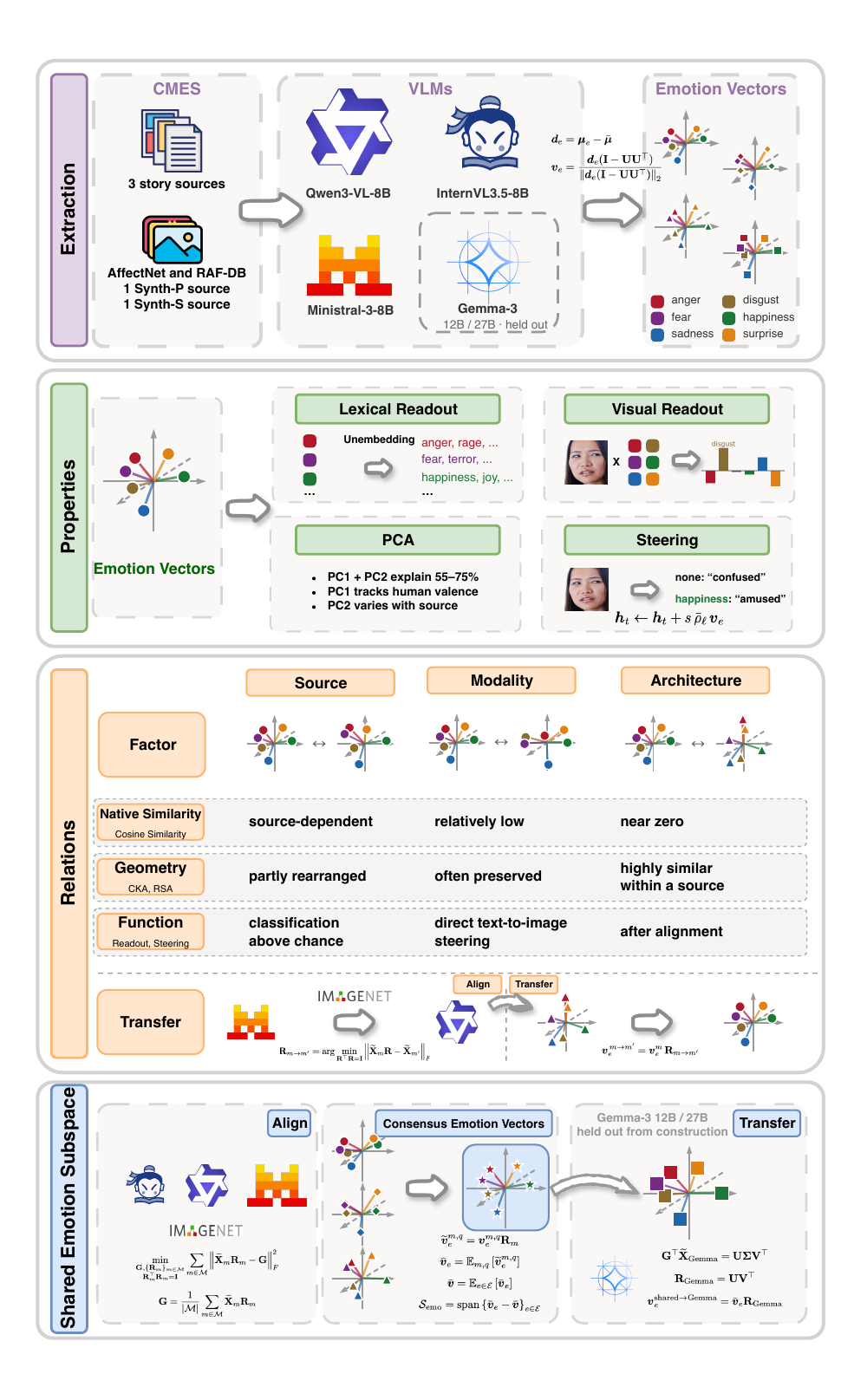}
\end{minipage}
\vspace{-2em}
\caption{Motivation and study overview. Left: The map--territory analogy and our motivation. Right: Emotion-vector analysis across sources, modalities, and architectures, shared-subspace construction, and held-out Gemma-3 evaluation.}
\vspace{-1em}
\label{fig:overview}
\end{figure*}

As shown on the left side of Figure~\ref{fig:overview}, these gaps lead to four questions. 1) Existence. Do images also produce structured emotion vectors, and how does their geometry depend on stimulus source? 2) Correspondence. Do text- and image-derived vectors correspond, and can text-derived vectors steer image interpretation? 3) Transfer. Do these relations and causal effects extend across architectures? 4) Consensus. Can we construct a shared emotion subspace that preserves this structure and function and generalizes to a held-out architecture? The map--territory analogy summarizes our underlying idea. Different internal representations may take different forms while preserving some common structure.

The right side of Figure~\ref{fig:overview} summarizes our workflow. To address these questions, we first construct CMES (Cross-Modal Emotion Stimuli), a collection of text and image stimuli organized around the same emotion categories. We use Ekman's six basic emotions, anger, disgust, fear, happiness, sadness, and surprise~\citep{ekman1992}, which are shared by the facial-expression datasets in CMES. Following the story-based protocol of prior work~\citep{anthropic2026emotion}, CMES includes three independently generated text-story sources. It also includes two real facial-expression datasets~\citep{mollahosseini2017affectnet,li2017rafdb}, synthetic facial portraits, and synthetic emotion-evoking scenes generated with Qwen-Image~\citep{wu2025qwenimaget}. Multiple sources within each modality help distinguish source effects from modality effects. We perform the core analyses with Qwen3-VL-8B~\citep{bai2025qwen3}, InternVL3.5-8B~\citep{wang2025internvl3}, and Ministral-3-8B~\citep{liu2026ministral}. Further details are provided in Appendix~\ref{sec:setup}.

Our results answer these questions in sequence. \textit{First}, image inputs produce structured emotion vectors that can be interpreted through lexical and visual readouts. Their low-dimensional structure broadly resembles human valence--arousal organization. PC1 shows relatively stable alignment with human valence across sources, while PC2's alignment with arousal varies more. Across the three core models, changing stimulus source affects the six-emotion geometry more than changing architecture.
\textit{Second}, text- and image-derived emotion vectors show cross-modal correspondence despite low cosine similarity. Before alignment, matched text--image vectors within the same model have cosine similarity of only $0.15$--$0.31$. Yet an alignment fitted on five emotions often generalizes to the held-out sixth in leave-one-emotion-out (LOEO) evaluation. Text-derived vectors also selectively steer the interpretation of face images when injected at image-token positions. Their effects are similar to those of image-derived vectors from the same model. The contrast between low cosine similarity and preserved steering effects motivates us to test whether emotion vectors can also transfer across architectures.
\textit{Third}, emotion vectors from different models also correspond after alignment. Matched Ministral-3 and Qwen3-VL emotion vectors have near-zero cosine similarity in their native residual spaces, but LOEO correspondence remains well above permutation controls. A transformation fitted to paired activations on neutral-face or generic ImageNet images recovers substantial correspondence without using the six emotion vectors or their labels. The mapped vectors also transfer causal effects across models.
\textit{Finally}, after aligning the three core architectures, we construct a shared emotion subspace from the centered consensus emotion vectors. It preserves the broad valence--arousal organization of the six emotions. The consensus vectors produce selective steering, and their lexical readouts generally show words related to the corresponding emotions. Although Gemma-3-12B and Gemma-3-27B are excluded from consensus construction, the mapped consensus vectors show correspondence with their emotion vectors and causally steer both variants despite their different widths and depths. 
Together, these results suggest that emotion representations can share relational structure and causal effects across sources, modalities, and architectures. In the map--territory view, different VLMs form different internal maps that preserve part of the same emotion structure.
Our main contributions are summarized 
%below.
as follows: 
\begin{itemize}
\vspace{-0.5em}
\item We introduce CMES, a collection of seven text and image sources covering the same six basic emotions. It supports comparisons of emotion representations and their causal effects across stimulus sources, modalities, and architectures.
\vspace{-0.5em}
\item We identify structured image-derived emotion vectors with a low-dimensional organization similar to text-derived vectors. Their geometry varies more across sources than across architectures, while valence alignment is more stable than arousal alignment.
\vspace{-0.5em}
\item We show that text- and image-derived emotion vectors within the same model can retain correspondence and similar steering effects despite low cosine similarity.
\vspace{-0.5em}
\item We show that mappings learned from paired ImageNet images recover cross-architecture correspondence and enable causal transfer despite near-zero native cosine similarity.
\vspace{-0.5em}
\item We construct a shared emotion subspace across three architectures. Its consensus vectors retain the original vectors' functional properties and generalize to held-out Gemma-3.
\end{itemize}

\section{CMES and Emotion-Vector Extraction}
\label{sec:cmes}

CMES covers six basic emotions (anger, disgust, fear, happiness, sadness, and surprise), denoted by $\mathcal E$, across seven sources. These include stories generated by Gemini (T1), Qwen3-Max (T2), and Qwen3-VL (T3), real faces from AffectNet and RAF-DB, and synthetic portraits (Synth-P) and synthetic emotion-evoking scenes (Synth-S). Real-face vectors use training images, while functional evaluations use separate validation or test splits.

For each model and source, we use residual activations at approximately two-thirds of the model's depth unless otherwise stated. For text, we average activations over story tokens from position~50 onward, following~\citet{anthropic2026emotion}. For images, we average over all visual tokens. We then average across examples of each emotion $e\in\mathcal E$ to obtain its class mean $\bm\mu_e$. We subtract the mean across the six emotions, then remove background directions estimated from neutral stimuli and normalize to obtain
\begin{equation}
\bm d_e
=
\bm\mu_e
-
\frac{1}{|\mathcal E|}
\sum_{e'\in\mathcal E}\bm\mu_{e'},
\qquad
\bm v_e
=
\frac{
\bm d_e(\mathbf I-\mathbf U\mathbf U^\top)
}{
\left\|
\bm d_e(\mathbf I-\mathbf U\mathbf U^\top)
\right\|_2
},
\label{eq:emovec}
\end{equation}
where $\mathbf U$ contains the leading neutral PCA~\citep{jolliffe2016PCA} directions needed to explain at least $50\%$ of the variance. This gives six vectors per model and source. Full setup and protocols are provided in Appendix~\ref{sec:setup}.

\section{Emotion Geometry and Representational Correspondence}
\label{sec:vectors}

\subsection{Visual Emotion Vectors Form Structured Low-Dimensional Representations}

\begin{figure}[t]
\centering
\includegraphics[width=0.95\linewidth]{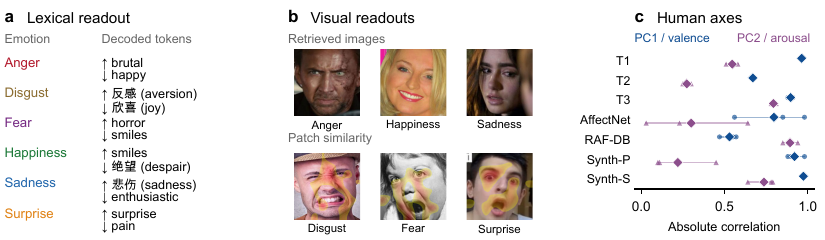}
\vspace{-1em}
\caption{Emotion-vector readouts and human-axis agreement. (a) Selected promoted ($\uparrow$) and suppressed ($\downarrow$) tokens for Qwen3-VL AffectNet vectors, with English translations. (b) Selected Qwen3-VL AffectNet retrievals (top) and patch-readout examples with the top $30\%$ of patch similarities highlighted (bottom). (c) Absolute correlations with Russell--Mehrabian (R\&M) ratings~\citep{russell1977evidence}. Small markers show models, diamonds show means, and lines show ranges.}
\label{fig:main-visual}
\end{figure}

We first explore whether image-derived emotion vectors are interpretable through lexical and visual readouts. We pass each $\bm v_e$ through the model's final normalization and unembedding, then decode the highest- and lowest-scoring tokens. High-scoring tokens generally describe the target emotion, while low-scoring tokens often express opposite emotions, as illustrated in Figure~\ref{fig:main-visual}a. Figure~\ref{fig:main-visual}b shows that ranking images by cosine similarity between their pooled activations and each emotion vector retrieves examples of the corresponding class, while patch-level similarity often highlights expressive facial regions within the images. More details are provided in Appendices~\ref{app:logitlens} and~\ref{app:qualitative}.

\begin{figure*}[t]
\centering
\vspace{-1em}
\includegraphics[width=0.95\textwidth]{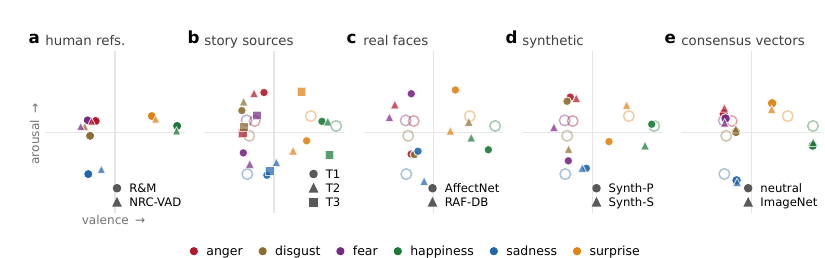}
\vspace{-1em}
\caption{Leading geometry of the six emotion vectors. (a) Human valence--arousal references from R\&M and the NRC Valence--Arousal--Dominance (NRC-VAD) lexicon~\citep{mohammad2018nrc}. (b--d) Qwen3-VL PCA projections from stories, real faces, and synthetic images. (e) Three-model consensus vectors. Faint circles mark R\&M positions. Display alignment is described in Appendix~\ref{app:metrics}.}
\vspace{-1em}
\label{fig:main-pca}
\end{figure*}

We next use PCA to explore the main dimensions of variation among the six emotion vectors. Across the model--source sets, PC1 explains $30$--$46\%$ of the variance, and PC1$+$PC2 explain $55$--$75\%$. We then correlate the six emotions' PC1 and PC2 coordinates with human valence and arousal ratings, respectively. As shown in Figure~\ref{fig:main-visual}c, agreement with human valence is generally stronger and more stable, whereas agreement with human arousal varies more across stimulus sources. This variation is not simply a text--image difference, as both modalities include sources with weak and strong arousal agreement. Figure~\ref{fig:main-pca} shows how the two-dimensional emotion configurations vary across text and image sources, while the consensus configuration introduced in Section~\ref{sec:demix} appears closer to the human reference. Full comparisons across vector sets and human ratings are provided in Appendix~\ref{app:axes}.

\subsection{Stimulus Source Changes Relational Geometry More Than Architecture}

\begin{figure*}[t]
\centering
\includegraphics[width=0.95\textwidth]{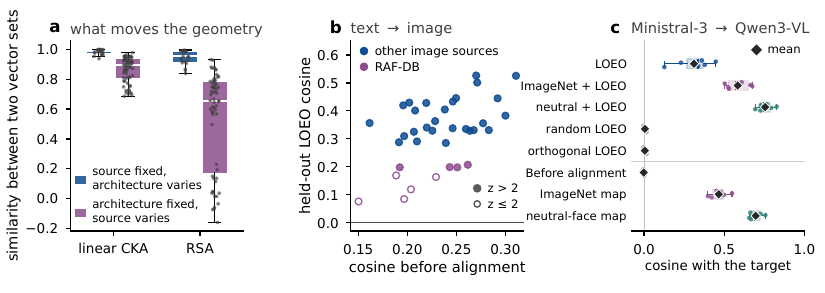}
\vspace{-1.5em}
\caption{Representational correspondence. (a) CKA and RSA with source or architecture fixed. (b) Text--image cosine before alignment versus LOEO. Filled markers indicate permutation $z>2$; open markers indicate $z\leq2$. (c) Ministral-3$\rightarrow$Qwen3-VL. Dots show source means before alignment and with LOEO, individual AffectNet emotions after mapping, and source/target replacement means for controls. Diamonds show means; boxes show quartiles and medians.}
\label{fig:main-correspondence}
\end{figure*}

The PCA visualization shows only the leading two-dimensional plane. We therefore use linear CKA~\citep{kornblith2019similarity} and RSA~\citep{kriegeskorte2008rsa} to compare the relationships among all six emotion vectors in the full representation space. To assess how source and architecture affect these relationships, we compare pairs that differ in only one of these factors. As shown in Figure~\ref{fig:main-correspondence}a, linear CKA ranges from $0.94$ to $1.00$ with median $0.99$ for the same source across different architectures. For different sources within the same architecture, CKA ranges from $0.69$ to $0.98$ with median $0.90$. The corresponding RSA medians are $0.95$ and $0.66$. Thus, within the three architectures studied here, changing stimulus source affects relational geometry more than changing architecture. RAF-DB accounts for much of the strongest source-specific change. Across the three architectures, its vectors place disgust closer to happiness and fear closer to anger. AffectNet and the other sources more often associate disgust with anger and fear with surprise, consistent with confusion patterns reported in human facial-expression recognition~\citep{mo2021confusion}. The ground-truth labels of retrieved images show similar differences between AffectNet and RAF-DB (Table~\ref{tab:topk-composition} in Appendix~\ref{app:qualitative}). Together, these results suggest that stimulus source changes the relationships among some emotions, rather than simply weakening emotion information. Detailed geometry and retrieval comparisons are provided in Appendices~\ref{app:qualitative}, \ref{app:setlevel}, and~\ref{app:sim-within}.

\subsection{Cross-Modal Correspondence Survives Low Cosine Similarity}

We next explore the similarity between text- and image-derived vectors for the same emotion within each VLM. These vectors share residual coordinates, yet their matched-emotion cosine similarity is only $0.15$--$0.31$ across the $36$ source/model comparisons. Together with the above geometric results, this suggests that cosine similarity may not fully capture cross-modal correspondence, even within the same model, motivating our cross-modal steering experiments in Section~\ref{sec:steering}. For a stricter test of generalization, we use LOEO alignment to fit an orthogonal transformation~\citep{schonemann1966generalized} on five matched emotions and evaluate the sixth. As shown in Figure~\ref{fig:main-correspondence}b, held-out cosine ranges from $0.29$ to $0.53$ for AffectNet and the two synthetic image sources. All $27$ comparisons exceed our permutation-based threshold of $z=2$. This suggests that the vectors preserve relationships among emotions across modalities, allowing alignment learned from five emotions to generalize to the sixth.
RAF-DB shows weaker LOEO correspondence with both text and other image sources. Text--image LOEO cosine ranges from $0.08$ to $0.21$, with only four of nine comparisons exceeding the permutation threshold, whereas all image--image comparisons exceed it. This weaker correspondence may reflect the different relationships among emotions observed for RAF-DB in the previous subsection. Overall, even when cosine similarity before alignment is low, a mapping learned from five emotions can generalize to the sixth, suggesting that text- and image-derived vectors preserve relationships among emotions. More details are provided in Appendix~\ref{app:sim-modal}.

\subsection{Cross-Architecture Correspondence Survives Near-Orthogonal Native Coordinates}

We further explore whether emotion-vector correspondence extends across model architectures. Figure~\ref{fig:main-correspondence}c shows the Ministral-3-to-Qwen3-VL comparison as an example. Across the seven stimulus sources, cosine similarity before alignment stays near zero. In contrast, LOEO cosine ranges from $0.13$ to $0.45$ with mean $0.31$. Matched-norm random and orthogonal LOEO controls instead give near-zero held-out cosine, suggesting that the observed correspondence depends on the emotion relationships rather than on alignment alone.
These results suggest that emotion vectors retain similar relationships across models despite their different coordinate bases. A map learned from five emotions can still predict the correspondence for the sixth. This motivates us to explore mappings learned without using the six emotion vectors or their labels. The neutral-face and ImageNet results in Figure~\ref{fig:main-correspondence}c preview such mappings, fitted to paired image activations as described in Section~\ref{sec:crossarch}. Across the six AffectNet emotion pairs, neutral-face mapping gives cosine of $0.66$--$0.76$ with mean $0.70$, while ImageNet mapping gives $0.40$--$0.55$ with mean $0.47$. Together, these results suggest that mappings learned from activations of generic images can recover cross-model emotion correspondence without fitting to the six emotion vectors. More details are provided in Appendix~\ref{app:sim-arch}.

\section{Causal Function Across Modalities}
\label{sec:steering}

Representational correspondence alone does not show that the vectors affect model behavior. We therefore intervene on activations while each VLM reads held-out face images. At layer $\ell$, we add a scaled unit emotion vector at every image-token position according to
\begin{equation}
\bm h_t
\leftarrow
\bm h_t
+
s\,\bar\rho_\ell\,\bm v_e,
\qquad
t\in\mathcal I_{\mathrm{img}},\quad e\in\mathcal E,
\label{eq:steer}
\end{equation}
where $\bm h_t$ is the residual activation at token position $t$ in layer $\ell$, $\bar\rho_\ell$ is the mean residual norm at that layer, and $s$ controls intervention strength and sign. Unless stated otherwise, we intervene at the same layer $\ell$ used to extract $\bm v_e$, with $s=+0.5$ for positive interventions and $s=-0.5$ for negative interventions. Relative to the unsteered model, we measure the change in log-probability of the injected emotion (\emph{diag}) and the mean change over the five competing emotions (\emph{off}). Positive selective steering should increase the injected-emotion readout and suppress competitors, with the pattern reversing for negative $s$.

\subsection{Emotion Vectors Selectively Steer Image Interpretation}

We first explore whether image-derived emotion vectors can selectively steer image interpretation. As shown in Figure~\ref{fig:main-causal}a, injection in Qwen3-VL on AffectNet increases the matched emotion's log-probability by $4.96$ and decreases that of competing emotions by $1.43$ on average, with the pattern reversing under negative injection. InternVL3.5 and Ministral-3 show a similar pattern, as shown in Figure~\ref{fig:main-causal}b. In contrast, matched-norm random and emotion-orthogonal directions produce near-zero changes and do not reproduce the separation between matched and competing emotions. More details are provided in Appendix~\ref{app:causal-within}.

\subsection{Text-Derived Vectors Functionally Substitute for Image-Derived Vectors}

\begin{figure*}[t]
\centering
\includegraphics[width=0.95\textwidth]{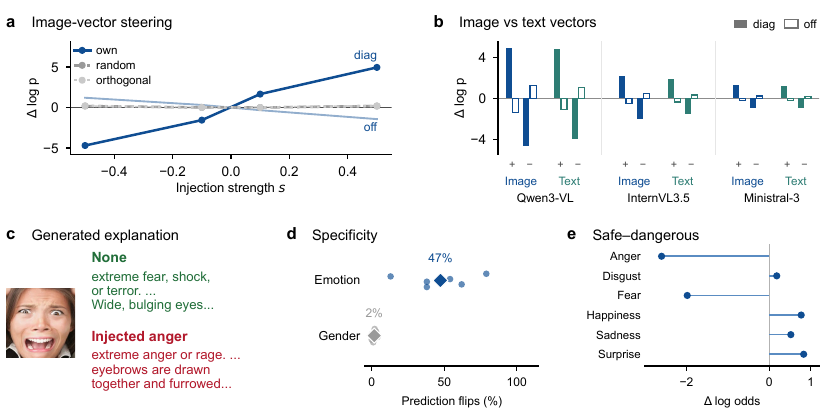}
\vspace{-1em}
\caption{Causal effects within models. (a) Qwen3-VL steering and controls. (b) Image- versus T3 text-derived vectors under positive ($+$) and negative ($-$) injection, with filled/open bars showing diag/off. (c) Qwen3-VL continuation excerpts before and after anger injection. (d) Emotion and gender prediction flips across six injections (dots) and their means (diamonds). (e) Changes in safe--dangerous log odds, with positive values favoring safe. }
\vspace{-1em}
\label{fig:main-causal}
\end{figure*}

We next replace the image-derived vectors with text-derived vectors from the same model, keeping all other settings unchanged. Despite text--image cosine similarity near $0.2$--$0.3$, the two vector types produce similar steering effects, as shown in Figure~\ref{fig:main-causal}b. The two vector types also produce similar effects in InternVL3.5 and Ministral-3 and suppress competing emotions in all three models. Detailed results are provided in Appendix~\ref{app:causal-modal}. Together with the results in Section~\ref{sec:vectors}, these findings suggest that high cosine similarity is not required for either similar relationships among emotions or similar causal effects across modalities.

Additionally, we test whether steering changes explanations beyond the first emotion word while preserving coherent output. As shown in Figure~\ref{fig:main-causal}c, continuations describe visual cues in ways that match the injected emotion. To test selectivity, we also compare emotion and gender readouts in Qwen3-VL. As shown in Figure~\ref{fig:main-causal}d, steering changes the emotion prediction on $47\%$ of evaluated RAF-DB images on average, compared with $2\%$ for gender. Steering also shifts the related safe--dangerous prediction toward dangerous for anger and fear and toward safe for happiness, as shown in Figure~\ref{fig:main-causal}e. To further test whether these vectors contribute to normal emotion recognition, we remove the centered span of the six source-specific vectors. This impairs recognition substantially more than removing a matched-rank random subspace (Table~\ref{tab:steer-ablate} in Appendix~\ref{app:causal-ablate}). Together, these tests suggest that emotion vectors selectively affect how models interpret emotions in images. More details are provided in Appendices~\ref{app:causal-specificity}, \ref{app:causal-ablate}, and~\ref{app:causal-beyond}.

\section{Cross-Architecture Transfer}
\label{sec:crossarch}

The preceding experiments suggest that emotion vectors have similar relationships and produce similar causal effects across modalities within a model. The cross-architecture LOEO results in Section~\ref{sec:vectors} also provide initial evidence that a mapping fitted directly to five matched pairs of emotion vectors can generalize to the sixth emotion. We therefore explore whether we can learn such a mapping without using emotion information, and whether it can also support causal transfer across architectures.

\subsection{Estimating the Coordinate Transformation Without the Six Emotion Vectors}

We explore whether activations from generic ImageNet images can align different models without using images from emotion datasets. For comparison, we also use activations from neutral faces in an emotion dataset. Let $\widetilde{\mathbf X}_m$ and $\widetilde{\mathbf X}_{m'}$ contain centered activations from source model $m$ and target model $m'$, with each row pair obtained from the same image. Then, we fit an orthogonal transformation and apply it to emotion vectors from each stimulus source $q$ as follows:
\begin{equation}
\mathbf R_{m\rightarrow m'}
=
\arg\min_{\mathbf R^\top\mathbf R=\mathbf I}
\left\|
\widetilde{\mathbf X}_m\mathbf R
-
\widetilde{\mathbf X}_{m'}
\right\|_F,
\qquad
\bm v_e^{\,m\rightarrow m',q}
=
\bm v_e^{m,q}\mathbf R_{m\rightarrow m'}.
\label{eq:neutralmap}
\end{equation}
As shown in Figure~\ref{fig:main-correspondence}c, the Ministral-3$\rightarrow$Qwen3-VL mapping learned from $3{,}000$ ImageNet validation images raises mean matched-emotion cosine on AffectNet from near zero to $0.47$. For comparison, a mapping learned from the same number of neutral faces reaches $0.70$. The recovery with ImageNet suggests that generic visual inputs can support cross-model emotion correspondence without using images from emotion datasets. Additional LOEO alignment further improves held-out correspondence for image-derived vectors (Figure~\ref{fig:main-correspondence}c) and text-derived vectors (Table~\ref{tab:map-loeo} in Appendix~\ref{app:sim-arch}).

\subsection{Mapped Emotion Vectors Transfer Causally Across Architectures}

\begin{figure}[t]
\centering
\includegraphics[width=\linewidth]{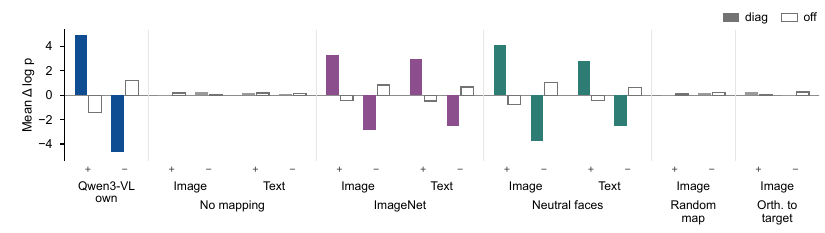}
\vspace{-2em}
\caption{Cross-architecture causal transfer on AffectNet. Ministral-3 image and T2 text vectors are injected into Qwen3-VL, with Qwen3-VL's own image vectors as a reference. Each condition shows positive ($+$) and negative ($-$) injections. Filled/open bars show diag/off effects.}
\label{fig:main-functional}
\vspace{-1em}
\end{figure}

We next test whether this recovered correspondence supports causal transfer. As shown in Figure~\ref{fig:main-functional}, unmapped Ministral-3 image-derived vectors produce essentially no selective effect in Qwen3-VL on AffectNet. Neutral-face and ImageNet mappings restore selective steering, increasing the matched emotion's log-probability by $4.11$ and $3.33$, respectively, while suppressing competing emotions. We further test transfer across both architectures and modalities by applying these mappings learned from images. For model pairs involving Ministral-3, mapped text vectors also produce selective bidirectional steering despite cosine similarity of only $0.18$--$0.29$. Figure~\ref{fig:main-functional} shows this transfer for Ministral-3$\rightarrow$Qwen3-VL using T2 vectors. This suggests that mappings learned from image activations can also transfer the causal effects of vectors extracted from text. More details are provided in Appendix~\ref{app:causal-arch}.

Together, these results suggest that differences in model coordinates can hide similarities between emotion vectors and limit their direct transfer. A mapping learned from paired image activations can recover both vector correspondence and selective causal effects without fitting to the six emotion vectors or their labels. In addition, these mappings further improve LOEO results, providing more evidence of shared relationships among emotions across models. 

\section{A Shared Emotion Subspace}
\label{sec:demix}

The above analyses compare pairs of models or sources. We now combine them to construct a shared emotion subspace across the three core architectures and seven stimulus sources, then evaluate the structure and causal effects it preserves.
We first extend the pairwise alignment in Equation~\ref{eq:neutralmap} to the set of three core models $\mathcal M$ using generalized Procrustes alignment. We jointly fit orthogonal transformations $\mathbf R_m$ and a shared template $\mathbf G$, which is the mean of the aligned activations across models. We then align the emotion vectors,
\begin{equation}
\min_{\substack{
\mathbf G,\{\mathbf R_m\}_{m\in\mathcal M}\\
\mathbf R_m^\top\mathbf R_m=\mathbf I,\ \forall m\in\mathcal M
}}
\sum_{m\in\mathcal M}
\left\|\widetilde{\mathbf X}_m\mathbf R_m-\mathbf G\right\|_F^2,
\quad
\mathbf G=\frac{1}{|\mathcal M|}\sum_{m\in\mathcal M}\widetilde{\mathbf X}_m\mathbf R_m,
\quad
\widetilde{\bm v}_e^{m,q}=\bm v_e^{m,q}\mathbf R_m.
\label{eq:shared-alignment}
\end{equation}
This places all three models in a common frame without choosing one as the reference. We average the aligned vectors over models and sources to obtain the consensus vector $\bar{\bm v}_e$ for each emotion $e\in\mathcal E$, then average them across emotions,
\begin{equation}
\bar{\bm v}_e
=
\mathbb{E}_{m,q}
\left[
\widetilde{\bm v}_e^{m,q}
\right],
\qquad
\bar{\bm v}
=
\mathbb{E}_{e\in\mathcal E}
\left[
\bar{\bm v}_e
\right].
\end{equation}
We then define the shared emotion subspace as
\begin{equation}
\mathcal{S}_{\mathrm{emo}}
=
\operatorname{span}
\left\{
\bar{\bm v}_e-\bar{\bm v}
\right\}_{e\in\mathcal E}.
\end{equation}
Centering removes the mean, so $\mathcal{S}_{\mathrm{emo}}$ spans the shared vectors that distinguish the six emotions.

We first use variance decomposition to explore how much variation is shared across models and sources. After alignment, shared differences among emotions remain, with little model-wide variation but substantial source variation and interactions. We map $\mathcal{S}_{\mathrm{emo}}$ back into each model's coordinates and compute cosine similarity between each original vector and its orthogonal projection onto this subspace. Under neutral-face alignment, the mean cosine over all seven sources and six emotions ranges from $0.59$ to $0.64$ across the three models. ImageNet alignment gives mean cosine of $0.57$--$0.63$, along with a similar emotion share in the variance decomposition. More details are provided in Appendix~\ref{app:decomp}.

We next test whether the part of each original vector inside the shared subspace retains its causal effects. We split each Qwen3-VL emotion vector into this shared projection and the orthogonal remainder outside $\mathcal{S}_{\mathrm{emo}}$, then inject each part separately (Equation~\ref{eq:emo-residual} in Appendix~\ref{app:demix-causal}). As shown in Figure~\ref{fig:main-consensus}a, the shared projection increases the matched emotion's log-probability slightly more than the original vector. The remainder produces only a very small effect. Random directions outside the shared subspace and remainders with shuffled emotion labels show little selective steering. In addition, we map the consensus vectors $\bar{\bm v}_e$ back into Qwen3-VL's coordinates and find that they produce selective steering comparable to its original vectors. More details are provided in Table~\ref{tab:dpca-causal} in Appendix~\ref{app:demix-causal}.

We also explore whether the consensus vectors retain lexical meaning and steer generated continuations. When mapped back to each model's native coordinates, consensus vectors show words related to the target emotion in their lexical readouts, as shown in Figure~\ref{fig:main-consensus}b. Figure~\ref{fig:main-consensus}c shows that Qwen3-VL continuations also follow the injected emotion and describe supporting visual cues. Together, these examples suggest that consensus vectors can produce clearer emotion-related readouts and steer continuations more effectively than the original emotion vectors. Figure~\ref{fig:main-pca}e also illustrates their agreement with the human valence--arousal reference under both alignment settings. To further test whether the consensus extends beyond the models used to construct it, we evaluate Gemma-3-12B and include Gemma-3-27B as a model-size ablation. Neither contributes to consensus construction, yet the mapped consensus vectors show correspondence with each variant's own vectors and retain causal effects in both, as shown in Figure~\ref{fig:main-consensus}d. These results support a shared emotion subspace whose consensus vectors retain lexical meaning and transfer causal effects to a held-out architecture at both model sizes. More details are provided in Appendices~\ref{app:demix-logitlens} and~\ref{app:gemma}.

\begin{figure}[t]
\centering
\includegraphics[width=\linewidth]{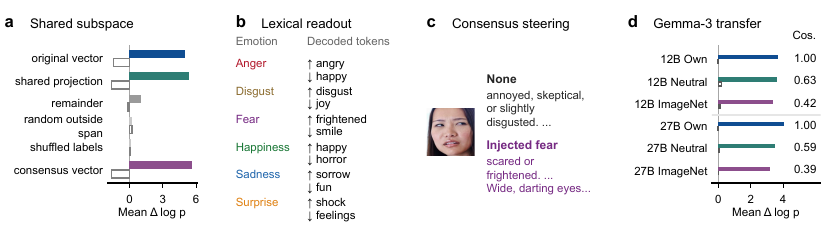}
\vspace{-1.5em}
\caption{Shared-subspace effects and consensus transfer. (a) Qwen3-VL interventions. (b) Selected promoted ($\uparrow$) and suppressed ($\downarrow$) tokens for Ministral-3 under ImageNet alignment. (c) Qwen3-VL continuations before and after consensus fear injection. (d) Gemma's own vectors versus neutral-face/ImageNet-mapped consensus vectors at $s=+0.1$. Filled/open bars in (a, d) show diag/off. Cos. denotes mean cosine similarity to the corresponding Gemma model's own AffectNet vectors.}
\label{fig:main-consensus}
\vspace{-1em}
\end{figure}

\section{Conclusions}

Our results suggest that emotion concepts in VLMs generalize through shared relationships among emotions and transferable causal effects, even when individual vector directions differ across sources, modalities, and architectures. Within a model, text- and image-derived vectors show held-out correspondence despite low cosine similarity, and text-derived vectors directly steer image interpretation. Across architectures, held-out correspondence can persist even when cosine similarity is near zero. Mappings learned from paired image activations then recover vector correspondence and causal transfer without using emotion information. Cosine similarity before alignment therefore gives an incomplete picture of this generalization. We further construct a shared emotion subspace across models and sources. This subspace preserves agreement with human valence and arousal, while projecting Qwen3-VL's original vectors onto it retains most of their selective steering effect. Qualitative examples also suggest that the consensus vectors produce clearer emotion-related lexical readouts and steer continuations more effectively than the original vectors. Moreover, they show correspondence and causal effects in Gemma-3-12B and Gemma-3-27B, although neither model contributes to their construction. We therefore use the map--territory analogy to summarize our findings: VLMs form different internal maps of emotion, yet these maps preserve some common structure and transferable causal effects.

\newpage

\bibliographystyle{iclr2027_conference}
\bibliography{iclr2027_conference}

@misc{anthropic2026emotion,
title={Emotion Concepts and their Function in a Large Language Model}, 
author={Nicholas Sofroniew and Isaac Kauvar and William Saunders and Runjin Chen and Tom Henighan and Sasha Hydrie and Craig Citro and Adam Pearce and Julius Tarng and Wes Gurnee and Joshua Batson and Sam Zimmerman and Kelley Rivoire and Kyle Fish and Chris Olah and Jack Lindsey},
year={2026},
eprint={2604.07729},
archivePrefix={arXiv},
primaryClass={cs.AI},
url={https://arxiv.org/abs/2604.07729}, 
}

@misc{ben2026where,
title={Where Do Models Find Happiness? Emotion Vectors in Open-Source {LLMs}},
author={Sinie van der Ben and Raphaël Baur and Yannick Metz and Mennatallah El-Assady},
year={2026},
eprint={2606.26987},
archivePrefix={arXiv},
primaryClass={cs.CL},
url={https://arxiv.org/abs/2606.26987}, 
}

@article{kobak2016demixed,
article_type = {journal},
title = {Demixed principal component analysis of neural population data},
author = {Kobak, Dmitry and Brendel, Wieland and Constantinidis, Christos and Feierstein, Claudia E and Kepecs, Adam and Mainen, Zachary F and Qi, Xue-Lian and Romo, Ranulfo and Uchida, Naoshige and Machens, Christian K},
editor = {van Rossum, Mark CW},
volume = 5,
year = 2016,
month = {apr},
pub_date = {2016-04-12},
pages = {e10989},
citation = {eLife 2016;5:e10989},
doi = {10.7554/eLife.10989},
url = {https://doi.org/10.7554/eLife.10989},
journal = {eLife},
issn = {2050-084X},
publisher = {eLife Sciences Publications, Ltd},
}

@article{ekman1992,
author = {Paul Ekman},
title = {An argument for basic emotions},
journal = {Cognition and Emotion},
volume = {6},
number = {3-4},
pages = {169--200},
year = {1992},
publisher = {Routledge},
doi = {10.1080/02699939208411068}
}

@article{ekman1971constants,
  title={Constants across cultures in the face and emotion.},
  author={Ekman, Paul and Friesen, Wallace V},
  journal={Journal of personality and social psychology},
  volume={17},
  number={2},
  pages={124},
  year={1971},
  publisher={American Psychological Association}
}

@inproceedings{bansal2021stitching,
author = {Bansal, Yamini and Nakkiran, Preetum and Barak, Boaz},
title = {Revisiting model stitching to compare neural representations},
year = {2021},
isbn = {9781713845393},
publisher = {Curran Associates Inc.},
address = {Red Hook, NY, USA},
booktitle = {Proceedings of the 35th International Conference on Neural Information Processing Systems},
articleno = {18},
numpages = {12},
series = {NIPS '21}
}

@inproceedings{moschella2023relative,
title={Relative representations enable zero-shot latent space communication},
author={Luca Moschella and Valentino Maiorca and Marco Fumero and Antonio Norelli and Francesco Locatello and Emanuele Rodol{\`a}},
booktitle={The Eleventh International Conference on Learning Representations },
year={2023},
url={https://openreview.net/forum?id=SrC-nwieGJ}
}

@inproceedings{huh2024platonic,
title = 	 {Position: The Platonic Representation Hypothesis},
author =       {Huh, Minyoung and Cheung, Brian and Wang, Tongzhou and Isola, Phillip},
booktitle = 	 {Proceedings of the 41st International Conference on Machine Learning},
pages = 	 {20617--20642},
year = 	 {2024},
editor = 	 {Salakhutdinov, Ruslan and Kolter, Zico and Heller, Katherine and Weller, Adrian and Oliver, Nuria and Scarlett, Jonathan and Berkenkamp, Felix},
volume = 	 {235},
series = 	 {Proceedings of Machine Learning Research},
month = 	 {21--27 Jul},
publisher =    {PMLR},
url = 	 {https://proceedings.mlr.press/v235/huh24a.html}
}

@misc{zhao2025hierarchical,
  title={Emergence of Hierarchical Emotion Organization in Large Language Models}, 
  author={Maya Okawa and Bo Zhao and Eric J. Bigelow and Rose Yu and Tomer Ullman and Ekdeep Singh Lubana and Hidenori Tanaka},
  year={2026},
  eprint={2507.10599},
  archivePrefix={arXiv},
  primaryClass={cs.CL},
  url={https://arxiv.org/abs/2507.10599}, 
}

@inproceedings{wang2025circuits,
title={Do {LLM}s {\textquotedblleft}Feel{\textquotedblright}? Emotion Circuits Discovery and Control},
author={Chenxi Wang and Yixuan Zhang and Ruiji Yu and Yufei Zheng and Lang Gao and Zirui Song and Zixiang Xu and Gus Xia and Huishuai Zhang and Dongyan Zhao and Xiuying Chen},
booktitle={Forty-third International Conference on Machine Learning},
year={2026},
url={https://openreview.net/forum?id=a8N0nRG3jA}
}

@inproceedings{reichman2026emotions,
title={Emotions Where Art Thou: Understanding and Characterizing the Emotional Latent Space of Large Language Models},
author={Benjamin Reichman and Adar Avsian and Larry Heck},
booktitle={The Fourteenth International Conference on Learning Representations},
year={2026},
url={https://openreview.net/forum?id=72TN9UAtNI}
}

@inproceedings{zheng2026neurons,
    title = "Are Emotion and Rhetoric Neurons in {LLM}? Neuron Recognition and Adaptive Masking for Emotion-Rhetoric Prediction Steering",
    author = "Zheng, Li  and
      Zhang, Xin  and
      He, Shuyi  and
      Li, Fei  and
      Teng, Chong  and
      Yang, Jiang-Ming  and
      Ji, Donghong  and
      Li, Zhuang",
    editor = "Liakata, Maria  and
      Moreira, Viviane P.  and
      Zhang, Jiajun  and
      Jurgens, David",
    booktitle = "Proceedings of the 64th Annual Meeting of the {A}ssociation for {C}omputational {L}inguistics (Volume 1: Long Papers)",
    month = jul,
    year = "2026",
    address = "San Diego, California, United States",
    publisher = "Association for Computational Linguistics",
    url = "https://aclanthology.org/2026.acl-long.193/",
    doi = "10.18653/v1/2026.acl-long.193",
    pages = "4206--4216",
    ISBN = "979-8-89176-390-6"
}

@article{russell1980circumplex,
author   = {Posner, Jonathan and Russell, James A. and Peterson, Bradley S.},
title    = {The circumplex model of affect: an integrative approach to affective neuroscience, cognitive development, and psychopathology},
journal  = {Development and Psychopathology},
year     = {2005},
volume   = {17},
number   = {3},
pages    = {715--734},
doi      = {10.1017/S0954579405050340},
pmid     = {16262989},
pmcid    = {PMC2367156},
issn     = {0954-5794},
language = {English}
}

@misc{zou2023repe,
title={Representation Engineering: A Top-Down Approach to {AI} Transparency},
author={Andy Zou and Long Phan and Sarah Chen and James Campbell and Phillip Guo and Richard Ren and Alexander Pan and Xuwang Yin and Mantas Mazeika and Ann-Kathrin Dombrowski and Shashwat Goel and Nathaniel Li and Michael J. Byun and Zifan Wang and Alex Mallen and Steven Basart and Sanmi Koyejo and Dawn Song and Matt Fredrikson and J. Zico Kolter and Dan Hendrycks},
year={2025},
eprint={2310.01405},
archivePrefix={arXiv},
primaryClass={cs.LG},
url={https://arxiv.org/abs/2310.01405}, 
}

@inproceedings{li2017rafdb,
  author={Li, Shan and Deng, Weihong and Du, JunPing},
  booktitle={2017 IEEE Conference on Computer Vision and Pattern Recognition (CVPR)}, 
  title={Reliable Crowdsourcing and Deep Locality-Preserving Learning for Expression Recognition in the Wild}, 
  year={2017},
  volume={},
  number={},
  pages={2584-2593},
  doi={10.1109/CVPR.2017.277}
}

@article{mollahosseini2017affectnet,
author={Mollahosseini, Ali and Hasani, Behzad and Mahoor, Mohammad H.},
journal={ IEEE Transactions on Affective Computing },
title={{ AffectNet: A Database for Facial Expression, Valence, and Arousal Computing in the Wild }},
year={2019},
volume={10},
number={01},
ISSN={1949-3045},
pages={18-31},
doi={10.1109/TAFFC.2017.2740923},
url = {https://doi.ieeecomputersociety.org/10.1109/TAFFC.2017.2740923},
publisher={IEEE Computer Society},
address={Los Alamitos, CA, USA},
month=jan
}

@article{russell1977evidence,
title = {Evidence for a three-factor theory of emotions},
journal = {Journal of Research in Personality},
volume = {11},
number = {3},
pages = {273-294},
year = {1977},
issn = {0092-6566},
doi = {https://doi.org/10.1016/0092-6566(77)90037-X},
url = {https://www.sciencedirect.com/science/article/pii/009265667790037X},
author = {James A Russell and Albert Mehrabian}
}

@inproceedings{mohammad2018nrc,
title = "Obtaining Reliable Human Ratings of Valence, Arousal, and Dominance for 20,000 {E}nglish Words",
author = "Mohammad, Saif",
editor = "Gurevych, Iryna  and
  Miyao, Yusuke",
booktitle = "Proceedings of the 56th Annual Meeting of the Association for Computational Linguistics (Volume 1: Long Papers)",
month = jul,
year = "2018",
address = "Melbourne, Australia",
publisher = "Association for Computational Linguistics",
url = "https://aclanthology.org/P18-1017/",
doi = "10.18653/v1/P18-1017",
pages = "174--184"
}

@inproceedings{kornblith2019similarity,
title = 	 {Similarity of Neural Network Representations Revisited},
author =       {Kornblith, Simon and Norouzi, Mohammad and Lee, Honglak and Hinton, Geoffrey},
booktitle = 	 {Proceedings of the 36th International Conference on Machine Learning},
pages = 	 {3519--3529},
year = 	 {2019},
editor = 	 {Chaudhuri, Kamalika and Salakhutdinov, Ruslan},
volume = 	 {97},
series = 	 {Proceedings of Machine Learning Research},
month = 	 {09--15 Jun},
publisher =    {PMLR},
url = 	 {https://proceedings.mlr.press/v97/kornblith19a.html}
}

@article{kriegeskorte2008rsa,
  author  = {Kriegeskorte, Nikolaus and Mur, Marieke and Bandettini, Peter},
  title   = {Representational similarity analysis -- connecting the branches of systems neuroscience},
  journal = {Frontiers in Systems Neuroscience},
  year    = {2008},
  volume  = {2},
  pages   = {4},
  doi     = {10.3389/neuro.06.004.2008},
  pmid    = {19104670},
  pmcid   = {PMC2605405}
}

@article{smith2005transmitting,
author  = {Smith, Marie L. and Cottrell, Garrison W. and Gosselin, Fr{\'e}d{\'e}ric and Schyns, Philippe G.},
title   = {Transmitting and decoding facial expressions},
journal = {Psychological Science},
year    = {2005},
volume  = {16},
number  = {3},
pages   = {184--189},
doi     = {10.1111/j.0956-7976.2005.00801.x},
pmid    = {15733197},
issn    = {0956-7976}
}

@inproceedings{arditi2024refusal,
author = {Arditi, Andy and Obeso, Oscar and Syed, Aaquib and Paleka, Daniel and Panickssery, Nina and Gurnee, Wes and Nanda, Neel},
title = {Refusal in language models is mediated by a single direction},
year = {2024},
isbn = {9798331314385},
publisher = {Curran Associates Inc.},
address = {Red Hook, NY, USA},
booktitle = {Proceedings of the 38th International Conference on Neural Information Processing Systems},
articleno = {4322},
numpages = {47},
location = {Vancouver, BC, Canada},
series = {NIPS '24}
}

@inproceedings{lee2025programming,
title={Programming Refusal with Conditional Activation Steering},
author={Bruce W. Lee and Inkit Padhi and Karthikeyan Natesan Ramamurthy and Erik Miehling and Pierre Dognin and Manish Nagireddy and Amit Dhurandhar},
booktitle={The Thirteenth International Conference on Learning Representations},
year={2025},
url={https://openreview.net/forum?id=Oi47wc10sm}
}

@inproceedings{stolfo2025improving,
title={Improving Instruction-Following in Language Models through Activation Steering},
author={Alessandro Stolfo and Vidhisha Balachandran and Safoora Yousefi and Eric Horvitz and Besmira Nushi},
booktitle={The Thirteenth International Conference on Learning Representations},
year={2025},
url={https://openreview.net/forum?id=wozhdnRCtw}
}

@inproceedings{huang2025cross,
    title = "Cross-model Transferability among Large Language Models on the Platonic Representations of Concepts",
    author = "Huang, Youcheng  and
      Huang, Chen  and
      Feng, Duanyu  and
      Lei, Wenqiang  and
      Lv, Jiancheng",
    editor = "Che, Wanxiang  and
      Nabende, Joyce  and
      Shutova, Ekaterina  and
      Pilehvar, Mohammad Taher",
    booktitle = "Proceedings of the 63rd Annual Meeting of the Association for Computational Linguistics (Volume 1: Long Papers)",
    month = jul,
    year = "2025",
    address = "Vienna, Austria",
    publisher = "Association for Computational Linguistics",
    url = "https://aclanthology.org/2025.acl-long.185/",
    doi = "10.18653/v1/2025.acl-long.185",
    pages = "3686--3704",
    ISBN = "979-8-89176-251-0"
}

@inproceedings{chen2026transferring,
 author = {Chen, Alan and Merullo, Jack and Stolfo, Alessandro and Pavlick, Ellie},
 booktitle = {Advances in Neural Information Processing Systems},
 doi = {10.52202/085713-1621},
 editor = {D. Belgrave and C. Zhang and H. Lin and R. Pascanu and P. Koniusz and M. Ghassemi and N. Chen},
 pages = {48531--48563},
 publisher = {Curran Associates, Inc.},
 title = {Transferring Linear Features Across Language Models With Model Stitching},
 url = {https://proceedings.neurips.cc/paper_files/paper/2025/file/4569a868e7aa891248832ec08445d071-Paper-Conference.pdf},
 volume = {38, Main Conference},
 year = {2025}
}

@misc{agarwal2026cross,
      title={Cross-Architecture Steering Transfer in Language Models: A Systematic Empirical Study}, 
      author={Ayushi Agarwal},
      year={2026},
      eprint={2608.05164},
      archivePrefix={arXiv},
      primaryClass={cs.CL},
      url={https://arxiv.org/abs/2608.05164}, 
}

@misc{bai2025qwen3,
      title={{Qwen3-VL} Technical Report},
      author={Shuai Bai and Yuxuan Cai and Ruizhe Chen and Keqin Chen and Xionghui Chen and Zesen Cheng and Lianghao Deng and Wei Ding and Chang Gao and Chunjiang Ge and Wenbin Ge and Zhifang Guo and Qidong Huang and Jie Huang and Fei Huang and Binyuan Hui and Shutong Jiang and Zhaohai Li and Mingsheng Li and Mei Li and Kaixin Li and Zicheng Lin and Junyang Lin and Xuejing Liu and Jiawei Liu and Chenglong Liu and Yang Liu and Dayiheng Liu and Shixuan Liu and Dunjie Lu and Ruilin Luo and Chenxu Lv and Rui Men and Lingchen Meng and Xuancheng Ren and Xingzhang Ren and Sibo Song and Yuchong Sun and Jun Tang and Jianhong Tu and Jianqiang Wan and Peng Wang and Pengfei Wang and Qiuyue Wang and Yuxuan Wang and Tianbao Xie and Yiheng Xu and Haiyang Xu and Jin Xu and Zhibo Yang and Mingkun Yang and Jianxin Yang and An Yang and Bowen Yu and Fei Zhang and Hang Zhang and Xi Zhang and Bo Zheng and Humen Zhong and Jingren Zhou and Fan Zhou and Jing Zhou and Yuanzhi Zhu and Ke Zhu},
      year={2025},
      eprint={2511.21631},
      archivePrefix={arXiv},
      primaryClass={cs.CV},
      url={https://arxiv.org/abs/2511.21631}, 
}

@misc{wang2025internvl3,
      title={{InternVL3.5}: Advancing Open-Source Multimodal Models in Versatility, Reasoning, and Efficiency},
      author={Weiyun Wang and Zhangwei Gao and Lixin Gu and Hengjun Pu and Long Cui and Xingguang Wei and Zhaoyang Liu and Linglin Jing and Shenglong Ye and Jie Shao and Zhaokai Wang and Zhe Chen and Hongjie Zhang and Ganlin Yang and Haomin Wang and Qi Wei and Jinhui Yin and Wenhao Li and Erfei Cui and Guanzhou Chen and Zichen Ding and Changyao Tian and Zhenyu Wu and Jingjing Xie and Zehao Li and Bowen Yang and Yuchen Duan and Xuehui Wang and Zhi Hou and Haoran Hao and Tianyi Zhang and Songze Li and Xiangyu Zhao and Haodong Duan and Nianchen Deng and Bin Fu and Yinan He and Yi Wang and Conghui He and Botian Shi and Junjun He and Yingtong Xiong and Han Lv and Lijun Wu and Wenqi Shao and Kaipeng Zhang and Huipeng Deng and Biqing Qi and Jiaye Ge and Qipeng Guo and Wenwei Zhang and Songyang Zhang and Maosong Cao and Junyao Lin and Kexian Tang and Jianfei Gao and Haian Huang and Yuzhe Gu and Chengqi Lyu and Huanze Tang and Rui Wang and Haijun Lv and Wanli Ouyang and Limin Wang and Min Dou and Xizhou Zhu and Tong Lu and Dahua Lin and Jifeng Dai and Weijie Su and Bowen Zhou and Kai Chen and Yu Qiao and Wenhai Wang and Gen Luo},
      year={2025},
      eprint={2508.18265},
      archivePrefix={arXiv},
      primaryClass={cs.CV},
      url={https://arxiv.org/abs/2508.18265}, 
}

@misc{liu2026ministral,
      title={Ministral 3}, 
      author={Alexander H. Liu and Kartik Khandelwal and Sandeep Subramanian and Victor Jouault and Abhinav Rastogi and Adrien Sadé and Alan Jeffares and Albert Jiang and Alexandre Cahill and Alexandre Gavaudan and Alexandre Sablayrolles and Amélie Héliou and Amos You and Andy Ehrenberg and Andy Lo and Anton Eliseev and Antonia Calvi and Avinash Sooriyarachchi and Baptiste Bout and Baptiste Rozière and Baudouin De Monicault and Clémence Lanfranchi and Corentin Barreau and Cyprien Courtot and Daniele Grattarola and Darius Dabert and Diego de las Casas and Elliot Chane-Sane and Faruk Ahmed and Gabrielle Berrada and Gaëtan Ecrepont and Gauthier Guinet and Georgii Novikov and Guillaume Kunsch and Guillaume Lample and Guillaume Martin and Gunshi Gupta and Jan Ludziejewski and Jason Rute and Joachim Studnia and Jonas Amar and Joséphine Delas and Josselin Somerville Roberts and Karmesh Yadav and Khyathi Chandu and Kush Jain and Laurence Aitchison and Laurent Fainsin and Léonard Blier and Lingxiao Zhao and Louis Martin and Lucile Saulnier and Luyu Gao and Maarten Buyl and Margaret Jennings and Marie Pellat and Mark Prins and Mathieu Poirée and Mathilde Guillaumin and Matthieu Dinot and Matthieu Futeral and Maxime Darrin and Maximilian Augustin and Mia Chiquier and Michel Schimpf and Nathan Grinsztajn and Neha Gupta and Nikhil Raghuraman and Olivier Bousquet and Olivier Duchenne and Patricia Wang and Patrick von Platen and Paul Jacob and Paul Wambergue and Paula Kurylowicz and Pavankumar Reddy Muddireddy and Philomène Chagniot and Pierre Stock and Pravesh Agrawal and Quentin Torroba and Romain Sauvestre and Roman Soletskyi and Rupert Menneer and Sagar Vaze and Samuel Barry and Sanchit Gandhi and Siddhant Waghjale and Siddharth Gandhi and Soham Ghosh and Srijan Mishra and Sumukh Aithal and Szymon Antoniak and Teven Le Scao and Théo Cachet and Theo Simon Sorg and Thibaut Lavril and Thiziri Nait Saada and Thomas Chabal and Thomas Foubert and Thomas Robert and Thomas Wang and Tim Lawson and Tom Bewley and Tom Edwards and Umar Jamil and Umberto Tomasini and Valeriia Nemychnikova and Van Phung and Vincent Maladière and Virgile Richard and Wassim Bouaziz and Wen-Ding Li and William Marshall and Xinghui Li and Xinyu Yang and Yassine El Ouahidi and Yihan Wang and Yunhao Tang and Zaccharie Ramzi},
      year={2026},
      eprint={2601.08584},
      archivePrefix={arXiv},
      primaryClass={cs.CL},
      url={https://arxiv.org/abs/2601.08584}, 
}

@misc{wu2025qwenimaget,
      title={Qwen-Image Technical Report}, 
      author={Chenfei Wu and Jiahao Li and Jingren Zhou and Junyang Lin and Kaiyuan Gao and Kun Yan and Sheng-ming Yin and Shuai Bai and Xiao Xu and Yilei Chen and Yuxiang Chen and Zecheng Tang and Zekai Zhang and Zhengyi Wang and An Yang and Bowen Yu and Chen Cheng and Dayiheng Liu and Deqing Li and Hang Zhang and Hao Meng and Hu Wei and Jingyuan Ni and Kai Chen and Kuan Cao and Liang Peng and Lin Qu and Minggang Wu and Peng Wang and Shuting Yu and Tingkun Wen and Wensen Feng and Xiaoxiao Xu and Yi Wang and Yichang Zhang and Yongqiang Zhu and Yujia Wu and Yuxuan Cai and Zenan Liu},
      year={2025},
      eprint={2508.02324},
      archivePrefix={arXiv},
      primaryClass={cs.CV},
      url={https://arxiv.org/abs/2508.02324}, 
}

@article{jolliffe2016PCA,
  author  = {Jolliffe, Ian T. and Cadima, Jorge},
  title   = {Principal component analysis: a review and recent developments},
  journal = {Philosophical Transactions of the Royal Society A: Mathematical, Physical and Engineering Sciences},
  year    = {2016},
  volume  = {374},
  number  = {2065},
  pages   = {20150202},
  doi     = {10.1098/rsta.2015.0202},
  pmid    = {26953178},
  pmcid   = {PMC4792409}
}

@article{schonemann1966generalized,
  author  = {Sch{\"o}nemann, Peter H.},
  title   = {A generalized solution of the orthogonal {Procrustes} problem},
  journal = {Psychometrika},
  year    = {1966},
  volume  = {31},
  number  = {1},
  pages   = {1--10},
  doi     = {10.1007/BF02289451},
  issn    = {1860-0980}
}

@article{nichols2002nonparametric,
  author  = {Nichols, Thomas E. and Holmes, Andrew P.},
  title   = {Nonparametric permutation tests for functional neuroimaging: a primer with examples},
  journal = {Human Brain Mapping},
  year    = {2002},
  volume  = {15},
  number  = {1},
  pages   = {1--25},
  doi     = {10.1002/hbm.1058},
  pmid    = {11747097},
  pmcid   = {PMC6871862},
  issn    = {1097-0193}
}

@inproceedings{zhang2026interpreting,
title={Interpreting and Enhancing Emotional Circuits in Large Vision-Language Models via Cross-Modal Information Flow},
author={Chengsheng Zhang and Chenghao Sun and Zhining Xie and Xinmei Tian},
booktitle={Forty-third International Conference on Machine Learning},
year={2026},
url={https://openreview.net/forum?id=wnKgWnx8mH}
}

@InProceedings{lian2025affectgpt,
  title = 	 {{A}ffect{GPT}: A New Dataset, Model, and Benchmark for Emotion Understanding with Multimodal Large Language Models},
  author =       {Lian, Zheng and Chen, Haoyu and Chen, Lan and Sun, Haiyang and Sun, Licai and Ren, Yong and Cheng, Zebang and Liu, Bin and Liu, Rui and Peng, Xiaojiang and Yi, Jiangyan and Tao, Jianhua},
  booktitle = 	 {Proceedings of the 42nd International Conference on Machine Learning},
  pages = 	 {36993--37014},
  year = 	 {2025},
  editor = 	 {Singh, Aarti and Fazel, Maryam and Hsu, Daniel and Lacoste-Julien, Simon and Berkenkamp, Felix and Maharaj, Tegan and Wagstaff, Kiri and Zhu, Jerry},
  volume = 	 {267},
  series = 	 {Proceedings of Machine Learning Research},
  month = 	 {13--19 Jul},
  publisher =    {PMLR},
  url = 	 {https://proceedings.mlr.press/v267/lian25a.html}
}

@inproceedings{cheng2024emotion,
author = {Cheng, Zebang and Cheng, Zhi-Qi and He, Jun-Yan and Sun, Jingdong and Wang, Kai and Lin, Yuxiang and Lian, Zheng and Peng, Xiaojiang and Hauptmann, Alexander G.},
title = {{Emotion-LLaMA}: multimodal emotion recognition and reasoning with instruction tuning},
year = {2024},
isbn = {9798331314385},
publisher = {Curran Associates Inc.},
address = {Red Hook, NY, USA},
booktitle = {Proceedings of the 38th International Conference on Neural Information Processing Systems},
articleno = {3518},
numpages = {49},
location = {Vancouver, BC, Canada},
series = {NIPS '24}
}

@article{xing2026emo,
  author  = {Xing, Bohao and Yu, Zitong and Liu, Xin and Yuan, Kaishen and Ye, Qilang and Xie, Weicheng and Yue, Huanjing and Yang, Jingyu and K{\"a}lvi{\"a}inen, Heikki},
  title   = {{EMO-LLaMA}: Enhancing Facial Emotion Understanding with Instruction Tuning},
  journal = {International Journal of Computer Vision},
  year    = {2026},
  volume  = {134},
  number  = {7},
  pages   = {350},
  doi     = {10.1007/s11263-026-02918-9},
  url     = {https://doi.org/10.1007/s11263-026-02918-9}
}

@article{mo2021confusion,
  author  = {Mo, Fan and Gu, Jingjin and Zhao, Ke and Fu, Xiaolan},
  title   = {Confusion Effects of Facial Expression Recognition in Patients With Major Depressive Disorder and Healthy Controls},
  journal = {Frontiers in Psychology},
  volume  = {12},
  pages   = {703888},
  year    = {2021},
  doi     = {10.3389/fpsyg.2021.703888},
  url     = {https://doi.org/10.3389/fpsyg.2021.703888}
}
%%%%%%%%%%%%%%%%%%%%%%%%%%%%%%%%%%%%%%%%%%%%%%%%%%%%%%%%%%%%
\appendix
\newpage

\section*{\textbf{Appendix}}

The appendix is organized as follows.
\begin{itemize}
\item \textbf{Appendix~\ref{app:related-work}}. Related work. We review emotion representations, activation steering, and cross-model transfer, and explain how our study differs from prior work.
\item \textbf{Appendix~\ref{sec:setup}}. Experimental setup and reproducibility. This section documents the models, stimulus sources and splits, generation prompts, emotion-vector extraction, steering and alignment protocols, similarity measures, and research materials planned for release.
\item \textbf{Appendix~\ref{app:exist}}. Existence and characterization across modalities, sources, and models. We present lexical and visual evidence for emotion vectors and explore their structure across depth, agreement with human affective ratings, relational geometry, and sensitivity to stimulus source and filtering.
\item \textbf{Appendix~\ref{app:sim}}. Representational similarity. We compare matched emotion vectors within modalities, across modalities, and across architectures using cosine similarity before alignment and held-out LOEO alignment with permutation controls. We also test mappings learned from paired image activations.
\item \textbf{Appendix~\ref{app:causal}}. Causal steering. We test steering within models and transfer across modalities and architectures. Further experiments explore specificity, continuation coherence, injection position, removal of the emotion-vector span, effects beyond the emotion readout, and depth dependence.
\item \textbf{Appendix~\ref{app:demix}}. Shared emotion subspace. We construct and evaluate the shared subspace through variance decomposition, geometric reconstruction, causal interventions, lexical readouts, and generated continuations. We also test held-out generalization to Gemma-3-12B and include Gemma-3-27B as a model-size ablation.
\item \textbf{Appendix~\ref{app:lim}}. Limitations, future work, and broader impact. We discuss limitations in model and emotion coverage, source dependence, and the need to evaluate practical applications. We also outline future research directions and potential risks of emotion inference and steering.
\item \textbf{Appendix~\ref{app:ethics}}. Ethics statement. This statement addresses data access and privacy, restrictions on redistributing real face images, limits of emotion inference, and potential misuse of emotion steering.
\item \textbf{Appendix~\ref{app:reproducibility}}. Reproducibility statement. This statement identifies where experimental protocols, results, and controls are documented and summarizes the planned release of research materials.
\item \textbf{Appendix~\ref{app:LLM}}. LLM usage statement. This statement describes the use of generative models for synthetic stories and images and of LLMs for language polishing.
\end{itemize}

\section{Related Work}
\label{app:related-work}

\paragraph{Emotion representations in language models.}
Recent work suggests that LLMs contain structured emotion representations. \citet{reichman2026emotions} report a low-dimensional emotional manifold whose main dimensions resemble human valence and arousal~\citep{russell1980circumplex}. More recently, \citet{ben2026where} recover similar valence-related geometry in two open-weight LLMs, while finding that arousal alignment is more sensitive to the story corpus used for extraction. Other studies identify emotion directions that generalize across contexts and their associated circuits~\citep{wang2025circuits}, as well as hierarchical organization similar to human emotion taxonomies~\citep{zhao2025hierarchical}. However, these studies mainly characterize emotion structure from text, leaving its consistency across text and image sources unclear. We extend prior work to image-derived emotion vectors and compare the geometry of the six emotions across stimulus sources, modalities, and architectures.

\paragraph{Activation steering and causal intervention.}
Activation steering changes model behavior by intervening along internal directions~\citep{zou2023repe}. For emotion, prior work shows that interventions along emotion directions can alter emotional judgments, downstream decisions, and generated behavior~\citep{anthropic2026emotion,wang2025circuits,zheng2026neurons,reichman2026emotions}. These studies mainly consider interventions within individual language models using text inputs, providing limited evidence for the transfer of emotion steering across modalities and architectures. More recently, \citet{zhang2026interpreting} trace image-to-text information flow and identify emotion-related attention heads and neurons within individual VLMs. In contrast, we study whether independently extracted text- and image-derived emotion vectors preserve relational geometry and causal effects across sources, modalities, and architectures.

\paragraph{Cross-model representational and steering transfer.}
Prior work has shown that representations learned by different neural networks can often be related despite differences in their native coordinate systems~\citep{bansal2021stitching,moschella2023relative,huh2024platonic}. More recent work directly aligns concept representations across text LLMs and transfers probes or steering vectors between models using linear or affine transformations~\citep{huang2025cross,chen2026transferring,agarwal2026cross}. However, successful transfer of individual probes or steering vectors does not by itself establish that a group of related concepts retains its internal geometry and causal function across modalities and architectures. We study six related emotion concepts as a group and compare their relational geometry across stimulus sources, modalities, and architectures. We also evaluate the transfer of their causal effects across models after alignment, linking the organization of the emotion group to the function of its individual vectors.

\section{Experimental Setup and Reproducibility}
\label{sec:setup}

This appendix specifies the models, stimulus sources, activation-extraction procedure, intervention protocol, cross-architecture alignment, and similarity measures used throughout the paper. Unless depth is varied explicitly, analyses use layers at approximately two-thirds of each model's depth. Representational analyses use layer~24 in all three core models. For causal experiments, emotion vectors are extracted and injected at layer~24 in Qwen3-VL and InternVL3.5 and layer~23 in Ministral-3. We study the six Ekman emotions---anger, disgust, fear, happiness, sadness, and surprise. Neutral stimuli are used to estimate nuisance directions and are not treated as a seventh emotion vector.

\subsection{Models and Inference}
\label{app:models}

We conduct the core analyses on three instruction-tuned VLMs: Qwen3-VL-8B-Instruct\footnote{\url{https://huggingface.co/Qwen/Qwen3-VL-8B-Instruct}} (36 layers, $D=4096$), InternVL3.5-8B-Instruct\footnote{\url{https://huggingface.co/OpenGVLab/InternVL3_5-8B-Instruct}} (36 layers, $D=4096$), and Ministral-3-8B-Instruct-2512\footnote{\url{https://huggingface.co/mistralai/Ministral-3-8B-Instruct-2512}} (34 layers, $D=4096$). For held-out architecture generalization (Appendix~\ref{app:gemma}), we additionally evaluate Gemma-3-12B-IT\footnote{\url{https://huggingface.co/google/gemma-3-12b-it}} (48 layers, $D=3840$) and Gemma-3-27B-IT\footnote{\url{https://huggingface.co/google/gemma-3-27b-it}} (62 layers, $D=5376$) as a within-family model-size ablation. The Gemma-3 analyses and interventions use layers~32 and~41 for 12B and 27B, respectively. Both variants differ in residual width and depth from the core models and are excluded from the three-model consensus construction. All experiments use bfloat16 inference on NVIDIA GH200 GPUs with each model's native multimodal processor.

\subsection{Stimulus Sources}
\label{app:stimuli}

CMES contains three text-story sources and four image sources. The text sources are a public story corpus generated by Gemini~3.1~Pro Preview (T1; \url{https://huggingface.co/datasets/ryancodrai/emotion-probes}) and two story sets generated for this work with Qwen3-Max (T2) and Qwen3-VL-8B (T3). T3 serves as a self-generation control for Qwen3-VL and as a third matched story source for the other architectures. The image sources comprise real facial-expression images from AffectNet and RAF-DB, synthetic facial portraits generated with Qwen-Image (Synth-P), and synthetic emotion-evoking scenes generated with Qwen-Image (Synth-S).

Each source contains the six Ekman emotions together with a neutral class used for nuisance-component estimation. AffectNet emotion vectors are estimated exclusively from the training split and evaluated on the validation split; RAF-DB vectors are likewise estimated from training and evaluated on test. No evaluation image is used to construct an emotion vector. Success-filtered variants are used only in Appendix~\ref{app:filter}.

\begin{table}[tbp]
\centering
\small
\setlength{\tabcolsep}{4pt}
\caption{Stimulus sources and sample counts. Unfiltered text counts follow the generation protocol. Every set contains six target emotion classes and a neutral class. Per-class counts cover all seven classes; the filtered text block lists counts per target emotion and for neutral separately. QF means Qwen-filtered, retaining stimuli whose labels are correctly predicted by Qwen3-VL.}
\label{tab:stimuli}
\begin{tabular}{@{}lllrr@{}}
\toprule
Label & Source & Stimulus type & Per class & Total \\
\midrule
\multicolumn{5}{l}{\emph{Text sources}} \\
T1 & Gemini 3.1 Pro Preview & Generated text & $1{,}200$ & $8{,}400$ \\
T2 & Qwen3-Max stories & Generated text & $1{,}200$ & $8{,}400$ \\
T3 & Qwen3-VL stories & Generated text & $1{,}200$ & $8{,}400$ \\
\midrule
\multicolumn{5}{l}{\emph{Image sources}} \\
\multicolumn{2}{l}{AffectNet} & Real faces & $3{,}000$ & $21{,}000$ \\
\multicolumn{2}{l}{RAF-DB} & Real faces & $281$--$500$ & $3{,}281$ \\
Synth-P & synthetic portraits & Generated images & $3{,}000$ & $21{,}000$ \\
Synth-S & synthetic scenes & Generated images & $3{,}000$ & $21{,}000$ \\
\midrule
\multicolumn{5}{l}{\emph{Success-filtered image controls}} \\
\multicolumn{2}{l}{AffectNet (filtered)} & Real faces & $377$ & $2{,}639$ \\
\multicolumn{2}{l}{RAF-DB (filtered)} & Real faces & $106$ & $742$ \\
\midrule
\multicolumn{5}{l}{\emph{Success-filtered text controls}} \\
Label & Source & \multicolumn{1}{r}{Per emotion} & Neutral & Total \\
\cmidrule(lr){1-5}
T1-\mbox{QF} & T1, filtered & \multicolumn{1}{r}{$622$} & $395$ & $4{,}127$ \\
T2-\mbox{QF} & T2, filtered & \multicolumn{1}{r}{$188$} & $343$ & $1{,}471$ \\
T3-\mbox{QF} & T3, filtered & \multicolumn{1}{r}{$82$} & $506$ & $998$ \\
\bottomrule
\end{tabular}
\end{table}

\subsection{Stimulus Generation and Prompts}
\label{app:prompts}

Figure~\ref{fig:generation-prompts} shows the prompt templates for generating stories, synthetic portraits, and synthetic scenes.

\paragraph{Stories.}
T2 and T3 follow the story-generation protocol of \citet{anthropic2026emotion}. The full generation protocol crosses a $171$-word emotion vocabulary with $100$ topics and requests $12$ independent stories per emotion--topic pair. For the six target emotion classes used here, we request $1{,}200$ stories per class and $7{,}200$ in total. Neutral stories are generated separately with the same generator, using a prompt that asks for factual descriptions without emotional content.
The conditioning instruction is never supplied to the probed model; activation extraction receives only the generated story text. Among the inspected stories from the six emotion-conditioned classes, the conditioning word or its corresponding noun/adjective form appears in only $0.14\%$ of T2 stories and $0.85\%$ of T3 stories, remaining below $1.5\%$ in every class. We retain these rare occurrences rather than filtering examples based on surface text.

\paragraph{Synthetic images.}
Synthetic images are generated with \texttt{Qwen/Qwen-Image-2512} at $1024\times1024$ resolution using $50$ diffusion steps and one recorded random seed per image. Portrait prompts cross emotion with demographic, style, composition, and lighting. Scene prompts instead use environment-centric topics. No generated image is manually selected, filtered, or regenerated after generation.

\begin{figure}[tbp]
\centering
\begingroup
\tcbset{colback=gray!5!white, colframe=black, colbacktitle=gray!65!black,
coltitle=white, fonttitle=\bfseries, boxrule=0.5pt, arc=2pt,
left=8pt, right=8pt, top=6pt, bottom=6pt, before skip=0pt, after skip=8pt}
\begin{tcolorbox}[title={Stories (T2 and T3)}]
\small\ttfamily
Write \{n\_stories\} different stories based on the following premise.\\[2pt]
Topic: \{topic\}\\[2pt]
The story should follow a character who is feeling \{emotion\}.\\[2pt]
Format the stories like so: \textrm{[\ldots]} The paragraphs should each be a fresh start, with no continuity. Try to make them diverse and not use the same turns of phrase.
\end{tcolorbox}
\begin{tcolorbox}[title={Synthetic portraits (Synth-P)}]
\small\ttfamily
A \{style\} \{composition\} of a \{demographic\} feeling \{emotion\}. \{lighting\}. Photorealistic, expressive face, no text, no watermark.
\end{tcolorbox}
\begin{tcolorbox}[title={Synthetic scenes (Synth-S)}, after skip=0pt]
\small\ttfamily
A \{style\} photograph of \{topic\}. Atmosphere: \{emotion\}. \{lighting\}. Photorealistic, scene-focused composition, no text, no watermark.
\end{tcolorbox}
\endgroup
\caption{Prompt templates for CMES stimulus generation. Braces mark placeholders. The story-format example is omitted.}
\label{fig:generation-prompts}
\end{figure}

\subsection{Activation Extraction and Emotion-Vector Construction}
\label{app:extract}

\paragraph{Text pooling.}
For each story, we forward only the generated story text without a chat template, hook the residual stream at every transformer layer, and average activations from token position~50 to the final story token. Stories shorter than $60$ tokens are discarded.

\paragraph{Image pooling.}
Images are processed with each model's native multimodal processor. We identify the model-specific visual-token span in the processed sequence and average residual activations over that span. The same model-specific span definition is used for subsequent image-token steering.

\paragraph{Emotion vectors.}
For model $m$, source $q$, layer $\ell$, and emotion $e\in\mathcal E$, let $\bm\mu_e$ denote the mean pooled activation of that class and
\begin{equation}
\bar{\bm\mu}
=
\frac{1}{|\mathcal E|}\sum_{e\in\mathcal E}\bm\mu_e.
\end{equation}
We form the centered class contrast
\begin{equation}
\bm d_e
=
\bm\mu_e-\bar{\bm\mu}.
\label{eq:diffmeans}
\end{equation}
Neutral stimuli are used separately to estimate high-variance nuisance directions. Let $\widetilde{\mathbf X}_0$ collect centered neutral activations and let $\mathbf U$ contain the leading principal components whose cumulative explained variance first reaches $50\%$. The final emotion vector is
\begin{equation}
\bm v_e
=
\frac{
\bm d_e(\mathbf I-\mathbf U\mathbf U^\top)
}{
\left\|
\bm d_e(\mathbf I-\mathbf U\mathbf U^\top)
\right\|_2
}.
\label{eq:setup-ev}
\end{equation}

\subsection{Steering Protocol}
\label{app:steerproto}

At intervention layer $\ell$, we add a scaled emotion vector to each residual-stream position in the image-token span:
\begin{equation}
\bm h_t
\leftarrow
\bm h_t
+
s\,\bar{\rho}_{\ell}\,\bm v_e,
\qquad
t\in\mathcal I_{\mathrm{img}},
\label{eq:setup-steer}
\end{equation}
where $\bar{\rho}_{\ell}$ is the mean residual norm at that layer estimated from $40$ held-out images. Unless stated otherwise, we use unit-length vectors at one layer with $s=+0.5$, or $s=-0.5$ for negative interventions. Appendix~\ref{app:demix-causal} additionally tests the shared projection and remainder at their original lengths, so the two injected components sum to the original vector before the model response is measured.

The model answers \emph{How does this person feel?} after the prime \emph{They feel}. We score the label words \emph{angry, disgusted, afraid, happy, sad, surprised}, and \emph{neutral} from the next-token distribution. Only the six emotion directions are injected. For an injected emotion $e$, \emph{diag} is the mean change in $\log p(e)$ relative to the unsteered model, while \emph{off} is the mean change over the other five emotion labels. Neutral remains a candidate readout and an input class, but is excluded from the off-diagonal average and is not a seventh emotion vector.

Within-model steering uses held-out images selected as correctly classified in a separate screening run, capped at $50$ per class including neutral. This yields $274/283$ RAF-DB/AffectNet images for Qwen3-VL, $249/243$ for InternVL3.5, and $253/227$ for Ministral-3. The Gemma-3 AffectNet subsets contain $283$ images for 12B and $332$ for 27B, also including neutral inputs. Every intervention run recomputes the unsteered distribution on its evaluation images. Accuracy under additional readout conditions is measured with the relevant candidate labels. Where uncertainty is reported, confidence intervals use $2{,}000$ paired percentile-bootstrap resamples over images, keeping all six injections for each sampled image together.

\subsection{Cross-Architecture Alignment}
\label{app:align-proto}

For source model $m$ and target model $m'$ among the three equal-width core architectures, we estimate an orthogonal transformation from paired activations without using the six emotion vectors or their labels. Given centered paired activation matrices $\widetilde{\mathbf X}_m$ and $\widetilde{\mathbf X}_{m'}$, we fit
\begin{equation}
\mathbf R_{m\rightarrow m'}
=
\arg\min_{\mathbf R^\top\mathbf R=\mathbf I}
\left\|
\widetilde{\mathbf X}_m\mathbf R
-
\widetilde{\mathbf X}_{m'}
\right\|_F.
\label{eq:setup-procrustes}
\end{equation}
An emotion vector from stimulus source $q$ is transferred by
\begin{equation}
\bm v^{\,m\rightarrow m',q}_e
=
\bm v^{m,q}_e\mathbf R_{m\rightarrow m'}.
\end{equation}

The default transformation is estimated from $3{,}000$ AffectNet neutral training faces processed by both models. As an out-of-domain control, we fit the same transformation from $3{,}000$ generic ImageNet validation images. For the three-model analysis, generalized Procrustes alignment places all architectures in a symmetric consensus frame before the shared emotion subspace is constructed. For the unequal-width Gemma-3 transfers, we compute the thin SVD $\widetilde{\mathbf X}_m^\top\widetilde{\mathbf X}_{m'}=\mathbf U\boldsymbol\Sigma\mathbf V^\top$ and use the rectangular map $\mathbf U\mathbf V^\top$. This map need not preserve every source direction's norm, so transferred vectors are normalized in the target space before standard steering. Gemma-3 does not contribute emotion vectors to consensus construction; its paired activations are used only to estimate a map from the fitted consensus frame to Gemma-3's coordinates.

\subsection{Similarity Measures}
\label{app:metrics}

Given two six-emotion sets $\{\bm v_e\}_{e\in\mathcal E}$ and $\{\bm w_e\}_{e\in\mathcal E}$, we distinguish axis alignment, relational geometry, and one-to-one vector correspondence.

\paragraph{Axis alignment.}
For each set, the six vectors are stacked as rows into $\mathbf V\in\mathbb R^{|\mathcal E|\times D}$, centered across emotions, and analyzed with PCA. PC1 and PC2 are oriented using the Russell--Mehrabian reference: PC1 is sign-aligned to pleasure and PC2 to arousal. Corresponding axes are compared using Pearson correlation over the six emotions.

For display in Figure~\ref{fig:main-pca}, each two-dimensional configuration is centered, normalized to unit Frobenius norm, and Procrustes-aligned to R\&M. This display alignment allows rotation and reflection, while the axis correlations above use only sign alignment. Faint circles mark the normalized R\&M reference positions.

\paragraph{Relational geometry.}
Linear CKA compares centered inner-product geometry:
\begin{equation}
\operatorname{CKA}(\mathbf V,\mathbf W)
=
\frac{
\left\|
\mathbf W^\top\mathbf V
\right\|_F^2
}{
\left\|
\mathbf V^\top\mathbf V
\right\|_F
\left\|
\mathbf W^\top\mathbf W
\right\|_F
}.
\label{eq:setup-cka}
\end{equation}
RSA instead forms each set's $6\times6$ cosine-distance matrix and computes the Spearman correlation between the $15$ unique pairwise distances. CKA and RSA therefore compare relational structure rather than native vector coordinates.

\paragraph{Vector correspondence.}
Mean cosine similarity measures matched-emotion correspondence in the original coordinates as follows.
\begin{equation}
\operatorname{CosSim}(\mathbf V,\mathbf W)
=
\frac{1}{|\mathcal E|}
\sum_{e\in\mathcal E}
\cos(\bm v_e,\bm w_e).
\label{eq:setup-rawcos}
\end{equation}
Full-fit cosine fits an orthogonal Procrustes transformation using all six matched emotions and evaluates those same pairs, so it is treated only as an optimistic description of alignability. LOEO cosine instead fits on five matched emotions and evaluates the held-out sixth, averaging over all six folds.

For each fit, the training rows are centered separately in the two spaces. Because five emotion vectors do not determine a full residual-space rotation, we select the orthogonal least-squares solution closest to the identity among solutions with the same fitting error. This specifies the otherwise undetermined action outside the fitted directions and preserves vector norms in the common fitting space. For unequal widths, we first pad the shorter space with zeros, apply the same square orthogonal fit, and retain the target coordinates for cosine evaluation. The padding is a computational convention and does not define native cosine across unequal widths. These emotion-based fits are separate from the activation-based transformations in Appendix~\ref{app:align-proto}.

The held-out emotion pair is excluded from alignment fitting, while its vectors follow the common six-class construction in Appendix~\ref{app:extract}. LOEO therefore tests generalization of the relation among the six emotion vectors, rather than extraction of a new emotion concept without any examples from that class.

\paragraph{Permutation null.}
To test whether LOEO correspondence is specific to emotion identity, we repeat the procedure after permuting the labels of one set~\citep{nichols2002nonparametric} and standardize the observed score as
\begin{equation}
z
=
\frac{
\operatorname{LOEO}_{\mathrm{obs}}
-
\mu_{\mathrm{perm}}
}{
\sigma_{\mathrm{perm}}
}.
\label{eq:setup-perm}
\end{equation}
The null uses $500$ non-identity label shuffles drawn with a fixed random seed. Each shuffle reruns all six LOEO folds, including fitting each transformation on the other five emotion pairs. Thus, the null evaluates the same held-out statistic as the observed score. We use $z>2$ as evidence that correspondence exceeds that expected from the unlabeled six-point geometry and report cosine alongside this standardized score to show its magnitude.

\subsection{Reproducibility Artifacts}
\label{app:compute}

We will release the analysis code, complete generation prompts, stimulus sample lists, synthetic-generation seeds, experiment configurations, and extracted emotion vectors. Exact model revisions, software environments, and job scripts are provided with the artifact. RAF-DB and AffectNet images themselves are not redistributed; instead, we release the identifiers required to reconstruct the experimental subsets under their original licences.

\section{Existence and Characterization Across Modalities, Sources, and Models}
\label{app:exist}

This appendix explores what the emotion vectors represent and how the six emotions are related across stimulus sources and models. We first explore the words, images, and image regions associated with individual vectors in Appendices~\ref{app:logitlens} and~\ref{app:qualitative}. We then compare the relationships among the six vectors across layers in Appendix~\ref{app:depth}, with human valence and arousal in Appendix~\ref{app:axes}, and across sources and models in Appendix~\ref{app:setlevel}. Finally, Appendix~\ref{app:filter} explores how stimulus source and selection affect the leading axes. These analyses provide the basis for the correspondence, causal steering, and shared-subspace experiments in Appendices~\ref{app:sim}--\ref{app:demix}.

\subsection{Logit-Lens of Emotion Vectors}
\label{app:logitlens}

\begin{figure}[htp]
\centering
\includegraphics[width=\linewidth]{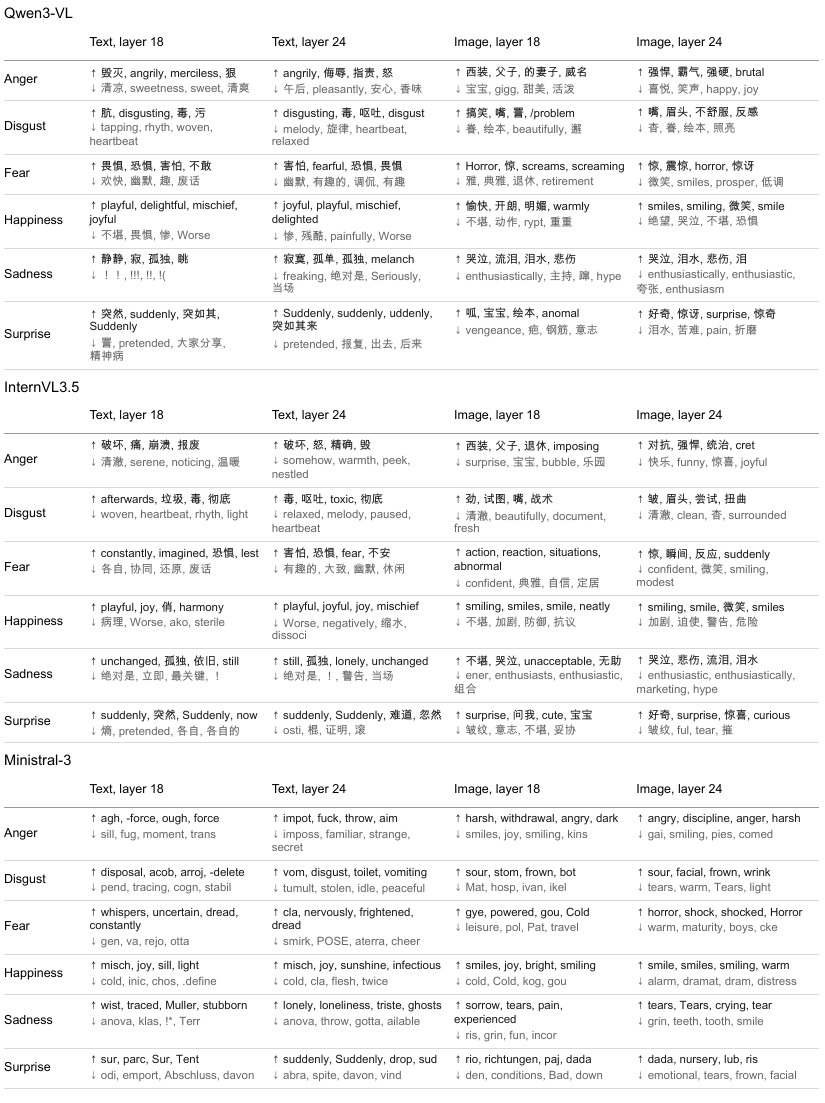}
\caption{Top-4 promoted ($\uparrow$) and suppressed ($\downarrow$) tokens for story- and AffectNet-derived emotion vectors at layers~18 and~24. Scores use each model's final normalization and unembedding.}
\label{fig:logitlens}
\end{figure}

We first explore whether the emotion vectors promote words related to their target emotions. Following \citet{anthropic2026emotion}, we pass each vector through the model's final normalization and unembedding and inspect its highest- and lowest-scoring tokens. Figure~\ref{fig:logitlens} compares text-story and AffectNet image vectors at layers~18 and~24. These are at $0.50$ and $0.67$ of the depth of Qwen3-VL and InternVL3.5, and at approximately $0.53$ and $0.71$ of the depth of Ministral-3. At layer~18, text vectors already tend to promote emotion-related words, although the pattern is less clear for Ministral-3. Image vectors at the same layer also promote words describing visual content. For example, Qwen3-VL's anger vector promotes tokens referring to a suit and family roles, which do not directly name the emotion.

At layer~24, the words promoted by image vectors are more closely related to emotion. The lowest-scoring tokens also often describe contrasting emotional states. For example, Qwen3-VL's AffectNet anger vector promotes \emph{brutal} while suppressing \emph{happy} and \emph{joy}. These examples suggest that image vectors become easier to interpret in emotional terms at the later layer. However, the readouts are not exact emotion labels, and some tokens still describe facial appearance or related emotions. The comparison supports a change between the two sampled layers, similar to the depth-related pattern reported for text representations by \citet{anthropic2026emotion}.

\subsection{Qualitative Evidence from Retrieved Images and Patch-Level Readouts}
\label{app:qualitative}

\begin{figure}[t]
\centering
\includegraphics[width=0.85\linewidth]{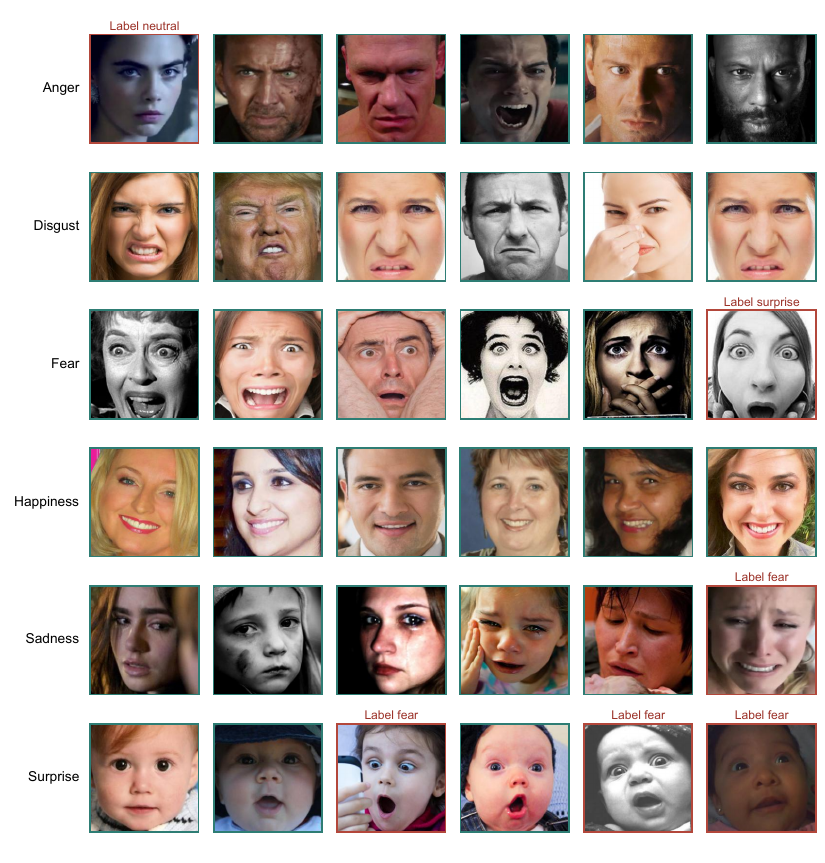}
\caption{Six random images from each Qwen3-VL emotion vector's retrievals. Green frames mark label matches; red frames mark mismatches and show the ground-truth label.}
\label{fig:topk}
\end{figure}

\paragraph{Images retrieved by each vector provide visual evidence of its meaning.}
We next test whether an emotion vector retrieves images of the corresponding emotion. Figure~\ref{fig:topk} shows Qwen3-VL results on AffectNet validation images. We rank all $3{,}999$ images by cosine similarity between their pooled activations and each vector, then randomly sample six examples from its top-$50$. The displayed disgust and happiness examples mostly match their target labels. Some mismatches involve related expressions, including fear and surprise in both directions. This is consistent with the overlap between these expressions reported in human facial perception \citep{ekman1971constants,smith2005transmitting}.

\paragraph{The full top-$50$ sets support the retrieval pattern in the examples.}
To check the pattern beyond the six displayed images, Table~\ref{tab:topk-purity} reports precision at $50$ (P@50), the fraction of target-class images among the top-$50$ retrievals for each emotion and model. The AffectNet pool contains eight approximately balanced classes, giving a reference rate of about $12.5\%$. Every emotion vector retrieves its target class well above this rate. Across models, happiness ($84$--$92\%$) and sadness ($82$--$90\%$) have the highest P@50. Disgust ($66$--$80\%$) and fear ($70$--$72\%$) are intermediate, while anger ($38$--$60\%$) and surprise ($54$--$58\%$) are generally lower. These results show that emotion vectors retrieve clear examples of their target classes. To explore whether the vectors also distinguish emotions beyond these top-ranked examples, we next evaluate six-way classification on all validation images from the six target classes.

\begin{table}[tbp]
\centering
\small
\setlength{\tabcolsep}{4pt}
\caption{Retrieval P@50 and six-way classification recall for AffectNet-trained vectors. P@50 evaluates the top-$50$ retrievals; recall evaluates all validation images within each of the six target classes. All values are percentages.}
\label{tab:topk-purity}
\begin{tabular}{@{}lrrrrrr@{}}
\toprule
Emotion & \multicolumn{3}{c}{Retrieval P@50 (\%)} & \multicolumn{3}{c}{Classification Recall (\%)} \\
\cmidrule(lr){2-4}\cmidrule(lr){5-7}
 & Qwen3-VL & InternVL3.5 & Ministral-3 & Qwen3-VL & InternVL3.5 & Ministral-3 \\
\midrule
anger & 58 & 60 & 38 & 41 & 52 & 29 \\
disgust & 80 & 66 & 68 & 34 & 37 & 30 \\
fear & 72 & 70 & 72 & 52 & 49 & 45 \\
happiness & 92 & 84 & 88 & 93 & 89 & 93 \\
sadness & 82 & 88 & 90 & 54 & 47 & 50 \\
surprise & 58 & 54 & 56 & 34 & 36 & 31 \\
\bottomrule
\end{tabular}
\end{table}

For this six-way classification, we subtract the equally weighted mean of the six training-class activations from each pooled image activation, score it by its dot product with each unit emotion vector, and predict the highest-scoring class. We evaluate all validation images from the six target classes, excluding neutral and contempt where present. For each emotion, recall is the fraction of all images with that ground-truth label that the classifier predicts correctly. Across the $18$ model--emotion combinations, P@50 and recall have a Spearman correlation of $\rho=+0.72$. Happiness performs well under both measures, but a clean top-ranked set does not always imply high class-wide recall. For example, Qwen3-VL's disgust vector has P@50 of $80\%$ but recall of only $34\%$. Retrieval ranks images separately for each vector using uncentered cosine, whereas classification compares the six scores after centering. An emotion vector can therefore retrieve clear examples of its target class even when classification across the full dataset remains limited. For Qwen3-VL on AffectNet, vector-based classification performs at a broadly similar level to direct questioning. On the same validation images, accuracy is $51.4\%$ for six-way vector classification and $47.6\%$ for direct questioning.

\paragraph{RAF-DB disgust vectors have low retrieval P@50 but retain classification ability.}
We repeat the analysis with RAF-DB-trained vectors and the RAF-DB test set in Table~\ref{tab:topk-purity-rafdb}. Because this pool is imbalanced, each P@50 value should be compared with the corresponding class's share of the pool. Five emotions show clear enrichment over their reference rates. Disgust differs, with P@50 of $8/16/6\%$ for Qwen3-VL, InternVL3.5, and Ministral-3, compared with a pool share of $5.2\%$. The increase is small for Qwen3-VL and Ministral-3 and larger for InternVL3.5, but all three have much lower P@50 for disgust than on AffectNet. Anger has a similar pool share of $5.3\%$ yet reaches P@50 of $72/66/64\%$, so class rarity alone does not explain the weak disgust retrievals.

Despite this low P@50, the centered six-way classifier has disgust recall of $60/48/53\%$, compared with $34/37/30\%$ on AffectNet. These values come from different evaluation datasets, so they do not provide a controlled comparison of classification difficulty. They do show that weak disgust retrieval does not imply that the RAF-DB vectors cannot distinguish disgust from the other five classes. To understand what the vector retrieves instead, we next explore the labels within its top-$50$ set and compare this pattern with the vector geometry.

\begin{table}[t]
\centering
\small
\setlength{\tabcolsep}{4pt}
\caption{Retrieval P@50 and six-way classification recall for RAF-DB-trained vectors. P@50 evaluates the top-$50$ retrievals; recall uses all test images within each target class. Pool share gives each class's reference rate in the $3{,}068$-image retrieval pool. All values are percentages.}
\label{tab:topk-purity-rafdb}
\begin{tabular}{@{}lrrrrrrr@{}}
\toprule
Emotion & \shortstack{Pool share\\(\%)} & \multicolumn{3}{c}{Retrieval P@50 (\%)} & \multicolumn{3}{c}{Classification Recall (\%)} \\
\cmidrule(lr){3-5}\cmidrule(lr){6-8}
 & & Qwen3-VL & InternVL3.5 & Ministral-3 & Qwen3-VL & InternVL3.5 & Ministral-3 \\
\midrule
anger & 5.3 & 72 & 66 & 64 & 69 & 51 & 54 \\
disgust & 5.2 & 8 & 16 & 6 & 60 & 48 & 53 \\
fear & 2.4 & 28 & 64 & 48 & 62 & 62 & 64 \\
happiness & 38.6 & 100 & 94 & 98 & 60 & 63 & 65 \\
sadness & 15.6 & 86 & 80 & 96 & 66 & 61 & 61 \\
surprise & 10.7 & 48 & 40 & 34 & 40 & 45 & 39 \\
\bottomrule
\end{tabular}
\end{table}

\paragraph{Retrieval labels help explain the source-specific differences.}
Table~\ref{tab:topk-composition} breaks down the top-$50$ retrievals for disgust and fear by their ground-truth labels. On AffectNet, most mismatches for disgust are anger images, while most mismatches for fear are surprise images. On RAF-DB, anger accounts for at most $2\%$ of the disgust retrievals. Happiness ($46/32/40\%$) and neutral ($44/34/48\%$) instead make up most of the set. Anger also accounts for a larger share of fear retrievals on RAF-DB ($24/22/20\%$) than on AffectNet ($8/6/6\%$), although fear and surprise remain prominent in these sets.

The vector geometry shows a related source difference, as shown in Figure~\ref{fig:raf-distance} in Appendix~\ref{app:sim-within}. Across the three models, RAF-DB places disgust closer to happiness and fear closer to anger, while the other sources more often associate disgust with anger and fear with surprise. Retrieval and vector geometry therefore provide complementary evidence. The first explores which labeled images score highly along each vector, while the second compares the six vectors directly. Together, they suggest that RAF-DB changes the relationships among some emotions, rather than simply weakening all emotion information.

\begin{table}[t]
\centering
\small
\setlength{\tabcolsep}{3pt}
\caption{Ground-truth composition (\%) of the top-$50$ images retrieved by the disgust and fear vectors on AffectNet and RAF-DB. Bold values mark the vector's target class and equal its retrieval P@50. A dash denotes $<0.5\%$.}
\label{tab:topk-composition}
\begin{tabular}{@{}llrrrrrrr@{}}
\toprule
Emotion vector & Model & Disgust & Fear & Anger & Happiness & Neutral & Surprise & Sadness \\
\midrule
\multicolumn{9}{l}{\emph{AffectNet validation pool}} \\
disgust & Qwen3-VL & \textbf{80} & 2 & 18 & -- & -- & -- & -- \\
 & InternVL3.5 & \textbf{66} & 4 & 26 & -- & -- & -- & 4 \\
 & Ministral-3 & \textbf{68} & 6 & 18 & -- & -- & -- & 8 \\
\addlinespace
fear & Qwen3-VL & 2 & \textbf{72} & 8 & -- & -- & 18 & -- \\
 & InternVL3.5 & 2 & \textbf{70} & 6 & -- & -- & 22 & -- \\
 & Ministral-3 & -- & \textbf{72} & 6 & -- & -- & 22 & -- \\
\midrule
\multicolumn{9}{l}{\emph{RAF-DB test pool}} \\
disgust & Qwen3-VL & \textbf{8} & -- & 2 & 46 & 44 & -- & -- \\
 & InternVL3.5 & \textbf{16} & -- & 2 & 32 & 34 & 4 & 12 \\
 & Ministral-3 & \textbf{6} & -- & -- & 40 & 48 & -- & 6 \\
\addlinespace
fear & Qwen3-VL & -- & \textbf{28} & 24 & 10 & -- & 32 & 6 \\
 & InternVL3.5 & 2 & \textbf{64} & 22 & 2 & -- & 8 & 2 \\
 & Ministral-3 & 4 & \textbf{48} & 20 & 4 & -- & 22 & 2 \\
\bottomrule
\end{tabular}
\end{table}

\begin{figure}[t]
\centering
\includegraphics[width=\linewidth]{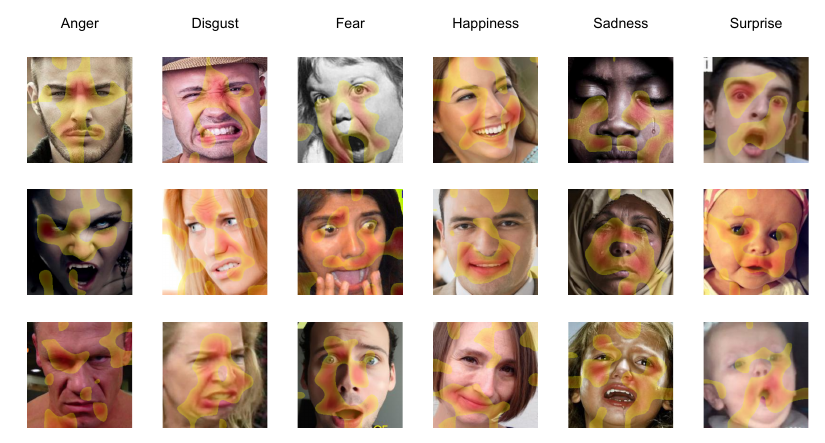}
\caption{Patch cosine similarity to Qwen3-VL emotion vectors. Each emotion shows three label-matching AffectNet validation images sampled from its top-$50$ retrievals. Overlays show the top $30\%$ of image tokens ranked by $\cos(\bm h_t,\bm v_e)$, with colors normalized for each image.}
\label{fig:patch-heatmaps}
\end{figure}

\paragraph{Patch-level readouts often highlight expressive facial regions.}
We then explore where high-similarity image tokens occur. Figure~\ref{fig:patch-heatmaps} compares each image-token activation $\bm h_t$ directly with the corresponding emotion vector $\bm v_e$, before pooling the tokens. In the displayed examples, high-cosine regions often overlap facial areas involved in the expression. These include the brow and central face for anger, the nose and mouth for disgust, and the eyes and mouth for fear and surprise. Happiness examples often highlight the mouth area, while sadness examples include regions around the eyes and brow. Hair, clothing, and image borders are generally less prominent in these examples.

These maps provide a qualitative view of token similarity, not a test of which image regions cause the model's prediction. The images are tightly cropped around faces, so they also do not establish a preference for faces over natural backgrounds. The signal is spatially broad. A token's position indicates where its visual input came from, but its later-layer activation can also contain information from other positions. The image-level retrieval result should therefore be understood as a property of the pooled visual representation, without assuming that one local patch carries the emotion signal.

\subsection{Emotion-Vector Structure Across Depth}
\label{app:depth}

\begin{figure}[h]
\centering
\includegraphics[width=\linewidth]{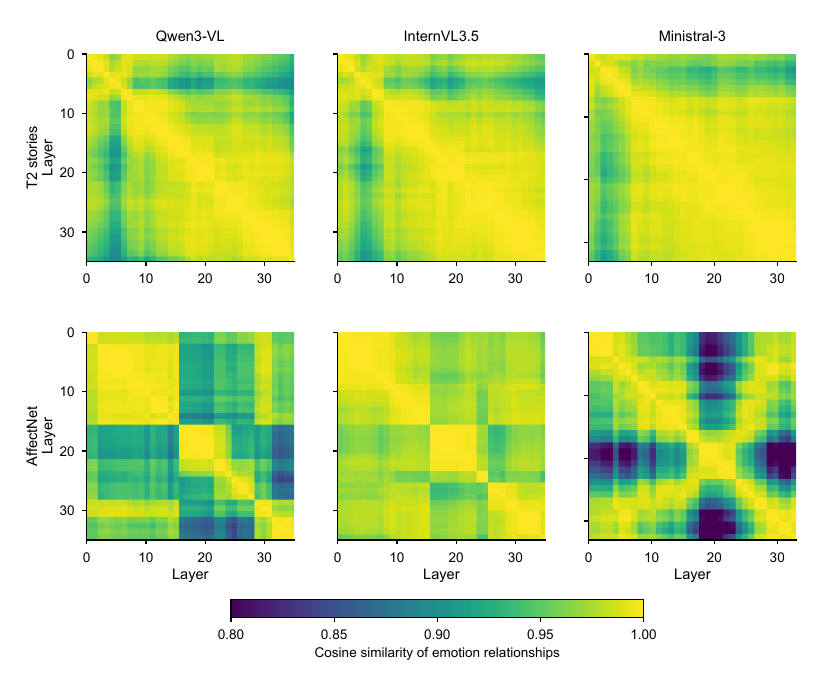}
\caption{Cross-layer geometry of T2 story and AffectNet emotion vectors. Each cell compares two layers by the cosine similarity between the $15$ unique off-diagonal entries of their emotion--emotion cosine matrices. }
\label{fig:layer-consist}
\end{figure}

We next explore whether the relationships among the six emotion vectors change across layers. At each layer, we compute the $6\times6$ emotion--emotion cosine matrix and keep its $15$ unique off-diagonal entries. We then compare two layers by the cosine similarity between these $15$-value lists. This compares the relationships among emotions, rather than the directions of individual vectors. Figure~\ref{fig:layer-consist} shows the resulting layer-by-layer comparisons for T2 stories and AffectNet images.

\paragraph{T2 geometry is stable across depth, while AffectNet geometry changes between layer ranges.}
For T2 stories, the similarity between the emotion relationships is at least $0.89$ for every pair of sampled layers in all three models. This is consistent with the text-based depth analysis of \citet{anthropic2026emotion}. AffectNet instead shows blocks of high similarity separated by changes across layers. Thus, its six-emotion geometry is relatively stable within some layer ranges but less stable across those ranges. Together with the lexical readouts in Appendix~\ref{app:logitlens}, this suggests that the image-derived representation changes as the model processes the image.

\subsection{Principal Axes Across Models and Stimulus Sources}
\label{app:axes}

We next use PCA to identify the main dimensions of variation among the six vectors. Table~\ref{tab:axes-variance} reports how much variance each principal component explains. It contains the $21$ original model--source sets, six filtered image sets, and three native-frame copies of the same consensus set. PC1 explains $30$--$46\%$ of the variance, and PC1$+$PC2 explain $55$--$75\%$, compared with $40\%$ if variance were evenly spread across five dimensions. The first two components therefore capture more than half of the variance in every reported set and at least $60\%$ in $25$ of the $30$ rows. 

\begin{figure}[h]
\centering
\includegraphics[width=\linewidth]{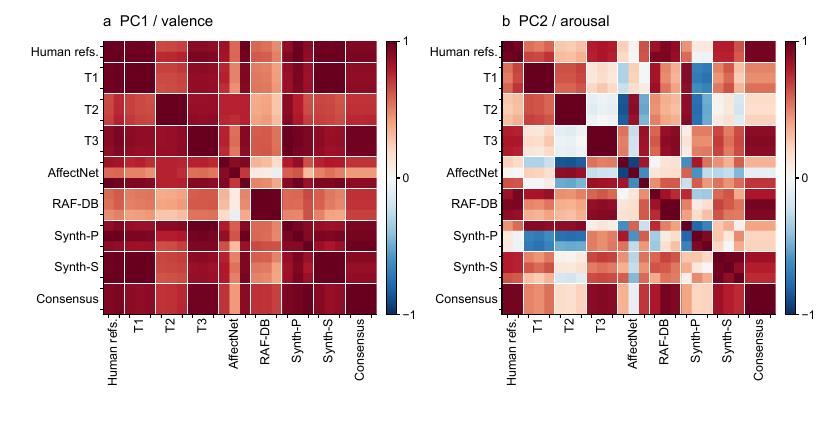}
\caption{Pearson correlations across six-emotion PC1 (a, valence) and PC2 (b, arousal) coordinates, with signs aligned to R\&M. Models are ordered Qwen3-VL, InternVL3.5, and Ministral-3 within each source and consensus group. Human references are ordered R\&M, then NRC-VAD. Consensus vectors are mapped into each model's native frame. }
\label{fig:pairwise-axes}
\end{figure}

\paragraph{The leading axes agree more consistently with valence than with arousal.}
We compare the six emotions' PC1 coordinates with human valence and their PC2 coordinates with human arousal using the R\&M \citep{russell1977evidence} and NRC-VAD \citep{mohammad2018nrc} references. Figure~\ref{fig:pairwise-axes} shows the correlations with human references and between emotion-vector sets. Among the $21$ original model--source sets, PC1 has a median correlation of $+0.90$ with R\&M valence, ranging from $+0.47$ to $+0.98$. The corresponding NRC-VAD median is $+0.93$, with a range of $+0.51$ to $+0.99$. Pairwise PC1 agreement among these sets also remains high, with a median of $+0.78$ over $210$ pairs.

PC2 agreement with human arousal varies more. Its median correlation is $+0.58$ with R\&M and $+0.66$ with NRC-VAD, while median pairwise PC2 agreement is $+0.41$. The lowest pairwise correlation is $-0.92$, between the Qwen3-VL and InternVL3.5 AffectNet sets. However, their correlations with R\&M arousal are only $0.23$ and $0.03$, respectively. The negative pairwise value therefore describes their sign-oriented PC2 coordinates and should not be read as evidence that the models encode opposite human arousal orderings. More generally, each correlation uses only six emotions. As a descriptive comparison, $14$ of the $21$ sets exceed $0.8$ for valence against at least one human reference, compared with four for arousal. We use this cutoff to summarize the strength of agreement, not as a statistical significance threshold.

\begin{table}[htbp]
\centering
\small
\setlength{\tabcolsep}{5pt}
\caption{PCA variance explained (\%) for six centered emotion vectors. PC1$+$PC2 gives the cumulative share of the leading plane. Consensus rows share the same spectrum because orthogonal mapping into each model preserves variance shares (Appendix~\ref{app:demix}).}
\label{tab:axes-variance}
\begin{tabular}{@{}lrrrrrr@{}}
\toprule
Stimulus source & PC1 & PC2 & PC3 & PC4 & PC5 & PC1$+$PC2 \\
\midrule
\multicolumn{7}{l}{\emph{Qwen3-VL}} \\
T1 & 30.7 & 25.2 & 19.2 & 13.9 & 11.1 & 55.9 \\
T2 & 38.2 & 27.6 & 20.3 & 8.4 & 5.5 & 65.8 \\
T3 & 40.2 & 25.9 & 19.9 & 8.0 & 6.1 & 66.0 \\
AffectNet & 34.0 & 29.8 & 18.8 & 11.6 & 5.8 & 63.8 \\
AffectNet (filtered) & 32.8 & 23.0 & 21.4 & 15.2 & 7.5 & 55.9 \\
RAF-DB & 43.0 & 19.9 & 17.5 & 12.6 & 6.9 & 63.0 \\
RAF-DB (filtered) & 38.5 & 24.5 & 17.1 & 11.9 & 8.1 & 62.9 \\
Synth-P & 38.7 & 29.9 & 22.5 & 6.4 & 2.5 & 68.6 \\
Synth-S & 42.6 & 24.3 & 16.6 & 10.7 & 5.8 & 66.9 \\
\addlinespace
Consensus & 41.4 & 25.6 & 20.2 & 7.5 & 5.3 & 67.0 \\
\midrule
\multicolumn{7}{l}{\emph{InternVL3.5}} \\
T1 & 30.4 & 25.1 & 18.8 & 14.4 & 11.3 & 55.5 \\
T2 & 37.7 & 27.4 & 20.9 & 8.1 & 5.9 & 65.1 \\
T3 & 40.6 & 25.9 & 19.1 & 7.7 & 6.7 & 66.5 \\
AffectNet & 37.0 & 29.2 & 18.9 & 10.3 & 4.7 & 66.2 \\
AffectNet (filtered) & 31.9 & 26.1 & 21.3 & 14.5 & 6.2 & 58.0 \\
RAF-DB & 43.2 & 19.9 & 16.0 & 12.3 & 8.6 & 63.1 \\
RAF-DB (filtered) & 42.5 & 22.2 & 15.1 & 12.7 & 7.5 & 64.7 \\
Synth-P & 37.9 & 29.6 & 21.4 & 7.4 & 3.8 & 67.5 \\
Synth-S & 45.9 & 23.8 & 14.6 & 9.6 & 6.2 & 69.7 \\
\addlinespace
Consensus & 41.4 & 25.6 & 20.2 & 7.5 & 5.3 & 67.0 \\
\midrule
\multicolumn{7}{l}{\emph{Ministral-3}} \\
T1 & 30.0 & 24.8 & 20.3 & 14.2 & 10.7 & 54.8 \\
T2 & 36.9 & 26.6 & 21.3 & 8.8 & 6.4 & 63.5 \\
T3 & 41.6 & 26.3 & 18.6 & 7.6 & 5.9 & 67.9 \\
AffectNet & 39.5 & 27.4 & 14.3 & 13.2 & 5.6 & 66.9 \\
AffectNet (filtered) & 37.4 & 23.1 & 17.2 & 14.5 & 7.8 & 60.5 \\
RAF-DB & 43.9 & 22.5 & 13.8 & 10.9 & 8.9 & 66.4 \\
RAF-DB (filtered) & 38.9 & 21.7 & 17.4 & 13.3 & 8.6 & 60.7 \\
Synth-P & 46.3 & 23.6 & 19.7 & 6.0 & 4.4 & 69.9 \\
Synth-S & 45.2 & 29.6 & 12.4 & 7.6 & 5.2 & 74.8 \\
\addlinespace
Consensus & 41.4 & 25.6 & 20.2 & 7.5 & 5.3 & 67.0 \\
\bottomrule
\end{tabular}
\end{table}

\subsection{Set-Level Similarity Across Stimulus Sources}
\label{app:setlevel}

The first two components leave about $25$--$45\%$ of the variance outside the displayed plane. We therefore compare all six vectors in the full representation space. Figure~\ref{fig:pairwise-full} reports three measures, defined in Appendix~\ref{app:metrics}. Linear CKA compares the centered inner-product structure and is unchanged by an orthogonal rotation or a uniform change in scale. RSA compares the rank order of the $15$ pairwise cosine distances between emotions. Cosine similarity compares each emotion vector directly with its counterpart in the native model coordinates. These measures answer different questions, so high agreement under one does not require high agreement under the others.

\begin{figure}[h]
\centering
\includegraphics[width=\linewidth]{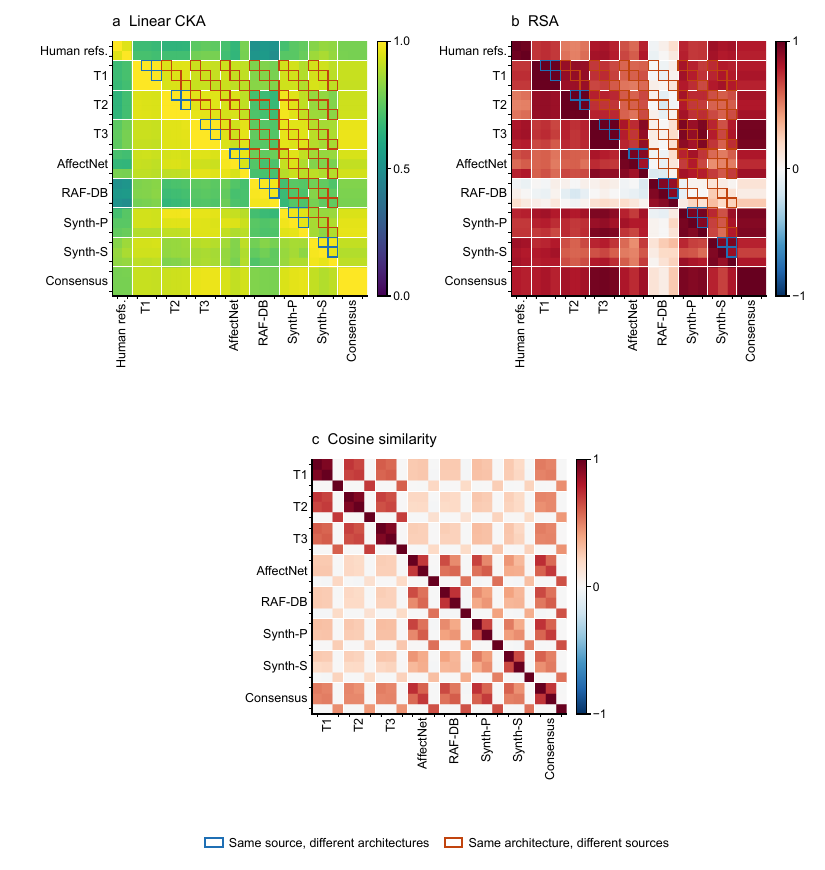}
\caption{Pairwise linear CKA (a), RSA (b), and mean matched-emotion cosine similarity (c). Models are ordered Qwen3-VL, InternVL3.5, and Ministral-3 within each source and consensus group. Human references are ordered R\&M, then NRC-VAD, and are omitted from (c) because they use different coordinates. Consensus vectors are mapped into each model's native frame. Blue outlines mark pairs with the same source and different architectures; orange outlines mark pairs with the same architecture and different sources.}
\label{fig:pairwise-full}
\end{figure}

\paragraph{Changing stimulus source produces larger geometry differences than changing architecture.}
To separate source and architecture effects, we compare pairs that differ in only one factor. We first hold the source fixed and vary the architecture. Across these $21$ pairs, linear CKA ranges from $0.94$ to $1.00$, with a median of $0.99$. The high agreement also holds for pairs involving Ministral-3, whose native coordinates differ strongly from those of Qwen3-VL and InternVL3.5. Thus, similar centered geometry does not depend on the models sharing native vector directions.

\begin{figure}[h]
\centering
\includegraphics[width=\linewidth]{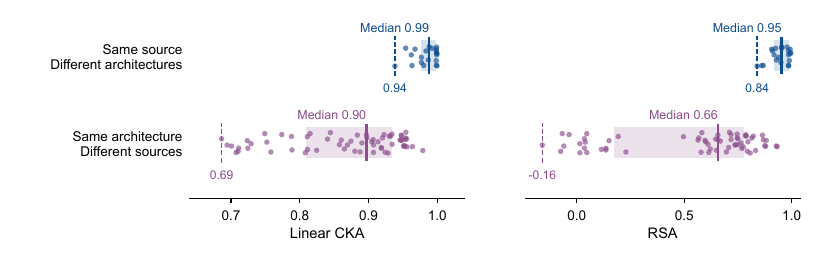}
\caption{CKA and RSA for the $21$ original model--source sets. Blue varies architecture with source fixed ($21$ pairs); purple varies source with architecture fixed ($63$ pairs). Points show all pairs and boxes show the interquartile range. Solid lines mark medians and dashed lines mark minima.}
\label{fig:factor-split}
\end{figure}

We then hold the architecture fixed and vary the stimulus source. Across these $63$ pairs, CKA ranges from $0.69$ to $0.98$, with a median of $0.90$ (Figure~\ref{fig:factor-split}). Some sources remain similar, so the two distributions overlap. However, source changes can reduce CKA to $0.69$, while architecture changes within a source keep it at or above $0.94$. Within the models and sources studied here, source differences therefore have a larger effect on the centered six-emotion geometry.

\paragraph{RSA makes the RAF-DB source difference more visible.}
RSA shows the same source-versus-architecture pattern. Its median is $0.95$ when the source is fixed and $0.66$ when the architecture is fixed, compared with CKA medians of $0.99$ and $0.90$. RAF-DB accounts for much of the lower source agreement. Within each model, its median RSA with the other sources is $0.04$, compared with $0.75$ among the remaining sources. All $18$ source pairs below $0.4$ involve RAF-DB. In contrast, RAF-DB vectors from different architectures have RSA of $0.87$--$0.93$. The RAF-DB pattern therefore repeats across all three models, although these comparisons alone do not rule out effects of the shared stimulus sample.

CKA also places RAF-DB apart from the other sources, but the difference is smaller. This suggests that RAF-DB changes the ranking of emotion relationships more strongly than their overall centered inner-product structure. The pattern is not a general text--image difference, because RAF-DB differs from both text sources and other image sources. Appendix~\ref{app:sim-within} explores the emotion pairs in more detail and identifies disgust and fear as the main contributors.

\paragraph{Cosine similarity shows differences in native vector directions.}
An orthogonal change of coordinates leaves CKA and RSA unchanged, but it can change cosine similarity between models. For Ministral-3 versus Qwen3-VL or InternVL3.5, matched-emotion cosine is near zero, as explored further in Appendix~\ref{app:sim-arch}. Qwen3-VL and InternVL3.5 instead have substantially higher native correspondence. Same-source text vectors reach approximately $0.93$--$0.94$, while image vectors reach $0.66$--$0.72$. InternVL3.5 uses a Qwen3-based language backbone, which may help explain this pattern.

\paragraph{Full-set similarity and axis agreement provide related but different views.}
We next compare these measures across the $210$ pairs of original model--source sets. CKA and RSA are strongly associated with pairwise PC1 agreement, with Spearman correlations of $0.75$ and $0.79$. Their correlations with PC2 agreement are smaller, at $0.13$ and $0.20$. Cosine similarity has little association with either axis, with correlations of $0.03$ for PC1 and $-0.03$ for PC2. These are descriptive comparisons among overlapping pairs, not independent observations. They support the distinction between the geometry of the six-emotion set and the native directions of its vectors.

The full-space measures also distinguish source pairs involving RAF-DB from the other source pairs. Both CKA and RSA give AUC values that round to $1.00$, compared with $0.97$ for PC1 agreement, while PC2 agreement provides little separation. Here, AUC summarizes how well each similarity score separates the two groups of source pairs. The small difference from PC1 alone does not establish that the full-space measures capture information independent of valence. Rather, axis agreement describes one dimension at a time, while CKA and RSA compare all relationships among the six vectors.

\subsection{Source Dependence of the Arousal Axis}
\label{app:filter}

The preceding sections show that PC1 generally agrees with human valence, while PC2 agreement with arousal varies more. We now explore whether this variation mainly follows stimulus source or model architecture. We then test whether selecting stimuli that a model recognizes correctly reduces the variation, and whether changes in individual axes also change the consensus vectors.

\begin{table}[tbp]
\centering
\small
\setlength{\tabcolsep}{5pt}
\caption{Absolute Pearson correlations with R\&M ratings across six emotions. PC1 is compared with valence and PC2 with arousal. Mean averages the three models; Spread is their maximum minus minimum. Sources are ordered by mean arousal agreement.}
\label{tab:arousal-source}
\begin{tabular}{@{}lrrrrr@{}}
\toprule
Stimulus source & Qwen3-VL & InternVL3.5 & Ministral-3 & Mean & Spread \\
\midrule
\multicolumn{6}{l}{\emph{Valence agreement $|r(\mathrm{PC1},\mathrm{valence})|$}} \\
RAF-DB & $0.57$ & $0.55$ & $0.47$ & $0.53$ & $0.10$ \\
T3 & $0.90$ & $0.91$ & $0.88$ & $0.90$ & $0.02$ \\
Synth-S & $0.97$ & $0.97$ & $0.98$ & $0.97$ & $0.01$ \\
T1 & $0.96$ & $0.97$ & $0.96$ & $0.97$ & $0.01$ \\
AffectNet & $0.85$ & $0.56$ & $0.98$ & $0.80$ & $0.42$ \\
T2 & $0.66$ & $0.67$ & $0.68$ & $0.67$ & $0.02$ \\
Synth-P & $0.88$ & $0.98$ & $0.90$ & $0.92$ & $0.10$ \\
\midrule
\multicolumn{6}{l}{\emph{Arousal agreement $|r(\mathrm{PC2},\mathrm{arousal})|$}} \\
RAF-DB & $0.85$ & $0.94$ & $0.89$ & $0.89$ & $0.09$ \\
T3 & $0.78$ & $0.81$ & $0.79$ & $0.80$ & $0.03$ \\
Synth-S & $0.78$ & $0.79$ & $0.64$ & $0.73$ & $0.15$ \\
T1 & $0.55$ & $0.51$ & $0.58$ & $0.55$ & $0.07$ \\
AffectNet & $0.23$ & $0.03$ & $0.64$ & $0.30$ & $0.61$ \\
T2 & $0.27$ & $0.25$ & $0.30$ & $0.28$ & $0.05$ \\
Synth-P & $0.45$ & $0.10$ & $0.11$ & $0.22$ & $0.35$ \\
\bottomrule
\end{tabular}
\end{table}

\paragraph{Arousal agreement varies widely across stimulus sources.}
Table~\ref{tab:arousal-source} reports each model's agreement with human ratings and summarizes the mean and spread within each source. Averaged across the three models, arousal agreement ranges from $0.22$ for synthetic portraits to $0.89$ for RAF-DB. This between-source range is $0.67$, while the average within-source range across models is $0.19$. Valence agreement varies less, with source means from $0.53$ to $0.97$ and an average within-source model range of $0.10$. These summaries describe the observed variation, rather than a formal decomposition of source and model effects.

Both modalities contain sources with weak and strong arousal agreement. Among text sources, the means range from $0.28$ for T2 to $0.80$ for T3. Among image sources, they range from $0.22$ for synthetic portraits to $0.89$ for RAF-DB. The variation therefore does not follow a simple text--image division. Some sources also vary substantially across models. For example, AffectNet has arousal agreement from $0.03$ to $0.64$, so the larger overall source range does not mean that model differences are absent.

\paragraph{Selecting correctly recognized stimuli can raise or lower axis agreement.}
To test whether hard-to-recognize stimuli account for the source variation, we re-estimate emotion vectors using only stimuli that Qwen3-VL classifies correctly. Within each source, we balance the six emotion classes to the smallest retained class. All three models use the same selected examples. Thus, selection depends on Qwen3-VL's predictions, not on the predictions of InternVL3.5 or Ministral-3. We refer to this procedure as success filtering.

Table~\ref{tab:filter-effect} shows that filtering changes arousal agreement in both directions. AffectNet rises from $0.03$--$0.64$ to $0.41$--$0.88$, while RAF-DB falls from $0.85$--$0.94$ to $0.65$--$0.85$. Text sources show the same contrast. T2 rises from $0.25$--$0.30$ to $0.77$--$0.83$, whereas T3 falls from $0.78$--$0.81$ to $0.59$--$0.66$. The correlation between the initial agreement and its change is $r=-0.91$ across these model--source settings. However, the change already includes subtraction of the initial value, so this correlation is not separate evidence that filtering corrects noise. The direct comparisons show that selecting correctly recognized stimuli changes the estimated axes without consistently moving them closer to the human reference.

\begin{table}[tbp]
\centering
\small
\setlength{\tabcolsep}{5pt}
\caption{Effect of success filtering on the two leading emotion axes, measured as sign-aligned $|r|$ against the R\&M norms. The same subset selected by Qwen3-VL is used for all three architectures. Retained sample sizes are reported in Table~\ref{tab:stimuli}.}
\label{tab:filter-effect}
\begin{tabular}{@{}llrrrr@{}}
\toprule
 &  & \multicolumn{2}{c}{Arousal $|r|$ (PC2)} & \multicolumn{2}{c}{Valence $|r|$ (PC1)} \\
\cmidrule(lr){3-4}\cmidrule(lr){5-6}
Source & Model & Unfiltered & Filtered & Unfiltered & Filtered \\
\midrule
AffectNet & Qwen3-VL & $0.23$ & $0.41$ & $0.85$ & $0.97$ \\
 & InternVL3.5 & $0.03$ & $0.63$ & $0.56$ & $0.98$ \\
 & Ministral-3 & $0.64$ & $0.88$ & $0.98$ & $0.97$ \\
\addlinespace
RAF-DB & Qwen3-VL & $0.85$ & $0.65$ & $0.57$ & $0.67$ \\
 & InternVL3.5 & $0.94$ & $0.79$ & $0.55$ & $0.65$ \\
 & Ministral-3 & $0.89$ & $0.85$ & $0.47$ & $0.56$ \\
\addlinespace
T1 & Qwen3-VL & $0.55$ & $0.71$ & $0.96$ & $0.93$ \\
 & InternVL3.5 & $0.51$ & $0.70$ & $0.97$ & $0.90$ \\
 & Ministral-3 & $0.58$ & $0.70$ & $0.96$ & $0.91$ \\
\addlinespace
T2 & Qwen3-VL & $0.27$ & $0.77$ & $0.66$ & $0.95$ \\
 & InternVL3.5 & $0.25$ & $0.83$ & $0.67$ & $0.93$ \\
 & Ministral-3 & $0.30$ & $0.81$ & $0.68$ & $0.97$ \\
\addlinespace
T3 & Qwen3-VL & $0.78$ & $0.59$ & $0.90$ & $0.75$ \\
 & InternVL3.5 & $0.81$ & $0.62$ & $0.91$ & $0.78$ \\
 & Ministral-3 & $0.79$ & $0.66$ & $0.88$ & $0.85$ \\
\bottomrule
\end{tabular}
\end{table}

\paragraph{Size-matched subsets help separate sample-count and selection effects.}
Filtering also reduces the sample count. As shown in Table~\ref{tab:filter-control}, we therefore compare the filtered AffectNet set with random subsets of the same size, using $377$ examples per class and five random seeds. The clearest selection effect occurs for InternVL3.5. Its arousal agreement is $0.03\pm0.03$ under random subsampling but $0.63$ after filtering. Ministral-3 shows a smaller difference, from $0.69\pm0.07$ to $0.88$. Qwen3-VL is less clear, with $0.37\pm0.23$ for random subsets and $0.41$ after filtering. 

\begin{table}[tbp]
\centering
\small
\setlength{\tabcolsep}{4pt}
\caption{AffectNet sample-size control. Random and filtered subsets contain $377$ images per class. Random results are mean $\pm$ standard deviation over five seeds. Valence and arousal are correlations of PC1 and PC2 with human ratings. CKA and RSA compare image vectors with the same model's T2 text vectors.}
\label{tab:filter-control}
\begin{tabular}{@{}lrrrrrr@{}}
\toprule
 & \multicolumn{2}{c}{R\&M} & \multicolumn{2}{c}{NRC-VAD} & \multicolumn{2}{c}{vs.\ T2 text} \\
\cmidrule(lr){2-3}\cmidrule(lr){4-5}\cmidrule(lr){6-7}
AffectNet subset & Valence & Arousal & Valence & Arousal & CKA & RSA \\
\midrule
\multicolumn{7}{l}{\emph{Qwen3-VL}} \\
Full set & 0.85 & 0.23 & 0.83 & 0.20 & 0.93 & 0.61 \\
Random subset & $0.81\pm0.22$ & $0.37\pm0.23$ & $0.79\pm0.22$ & $0.34\pm0.23$ & $0.92\pm0.02$ & $0.64\pm0.05$ \\
Success-filtered & 0.97 & 0.41 & 0.99 & 0.28 & 0.95 & 0.86 \\
\midrule
\multicolumn{7}{l}{\emph{InternVL3.5}} \\
Full set & 0.56 & 0.03 & 0.57 & 0.00 & 0.95 & 0.60 \\
Random subset & $0.56\pm0.16$ & $0.03\pm0.03$ & $0.57\pm0.15$ & $0.02\pm0.06$ & $0.94\pm0.01$ & $0.66\pm0.03$ \\
Success-filtered & 0.98 & 0.63 & 0.98 & 0.55 & 0.93 & 0.77 \\
\midrule
\multicolumn{7}{l}{\emph{Ministral-3}} \\
Full set & 0.98 & 0.64 & 0.97 & 0.63 & 0.91 & 0.66 \\
Random subset & $0.98\pm0.01$ & $0.69\pm0.07$ & $0.98\pm0.02$ & $0.68\pm0.08$ & $0.90\pm0.01$ & $0.63\pm0.05$ \\
Success-filtered & 0.97 & 0.88 & 0.98 & 0.85 & 0.90 & 0.65 \\
\bottomrule
\end{tabular}
\end{table}

\paragraph{Arousal agreement and shared emotion structure can change differently.}
RAF-DB has the highest mean arousal agreement in Table~\ref{tab:arousal-source}, at $0.89$, yet the weakest cross-source RSA in Appendix~\ref{app:setlevel}. AffectNet shows the reverse pattern, with lower arousal agreement but stronger relational agreement with the other non-RAF-DB sources. Agreement with one human axis therefore does not summarize how well a source preserves the relationships among all six emotions.

We also check whether filtering changes the consensus vectors constructed in Appendix~\ref{app:demix}. In this comparison, we replace only the AffectNet and RAF-DB vectors with their filtered versions, keeping the other five sources and the activation-based alignment setting unchanged. The two versions have mean matched-emotion cosine of $0.98$ and a minimum of about $0.95$. The planes formed by their first two principal components also remain close, with principal angles of $6^\circ$--$18^\circ$. These angles compare the leading two-dimensional planes, not the full shared subspaces. The result supports stability of the consensus vectors under this specific filtering comparison, even though individual source axes can change substantially. Our main analyses use the full labeled stimulus sets.

\section{Representational Similarity}
\label{app:sim}

Appendix~\ref{app:exist} shows that emotion vectors can share relationships even when their individual directions differ. We now explore whether vectors for the same emotion correspond across stimulus sources, modalities, and architectures. These comparisons complement the set-level geometry in Appendix~\ref{app:setlevel} and provide the basis for the causal-transfer experiments in Appendix~\ref{app:causal}.

We first vary the stimulus source within the same model and modality in Appendix~\ref{app:sim-within}. We then compare text and image vectors within each model in Appendix~\ref{app:sim-modal}, followed by comparisons across architectures in Appendix~\ref{app:sim-arch}. This order helps separate source differences from changes in modality and model coordinates.

We report four measures with different roles. Cosine similarity before alignment compares matched emotions in their original coordinates. Full-fit cosine uses all six emotion pairs to fit and evaluate an orthogonal transformation, so it measures fit to known pairs. LOEO cosine fits on five emotions and evaluates the sixth, testing whether the relationships learned from the other emotions predict a held-out pair. The permutation $z$ score compares mean LOEO cosine with the result of shuffling emotion labels and refitting. We use $z>2$ as the reporting threshold. All cosine values are averaged over the six emotions unless stated otherwise. Appendix~\ref{app:metrics} gives the definitions and controls.

\subsection{Within-Modality Consistency Across Stimulus Sources}
\label{app:sim-within}

We first explore how stimulus source affects matched-emotion correspondence while keeping model and modality fixed. Table~\ref{tab:within-modal} compares vectors from different story generators and different image datasets under this protocol.

\begin{table}[tbp]
\centering
\small
\setlength{\tabcolsep}{5pt}
\caption{Within-modality correspondence. Each row changes the stimulus source within one model. Cosines are means over six emotions. Full-fit uses all six pairs. LOEO fits five and evaluates the sixth. The $z$ score compares LOEO with the refitted label-permutation null. Source labels follow Table~\ref{tab:stimuli}.}
\label{tab:within-modal}
\begin{tabular}{@{}llrrrr@{}}
\toprule
 &  & \multicolumn{3}{c}{Mean cosine} & \multicolumn{1}{c}{Permutation} \\
\cmidrule(lr){3-5}
Model & Source pair & \shortstack{Before\\alignment} & Full-fit & LOEO & $z$ \\
\midrule
\multicolumn{6}{l}{\emph{Text sources}} \\
Qwen3-VL & T1--T2 & $0.71$ & $0.98$ & $0.72$ & $4.09$ \\
 & T1--T3 & $0.61$ & $0.96$ & $0.60$ & $4.38$ \\
 & T2--T3 & $0.69$ & $0.98$ & $0.72$ & $4.52$ \\
\addlinespace
InternVL3.5 & T1--T2 & $0.70$ & $0.98$ & $0.71$ & $4.26$ \\
 & T1--T3 & $0.60$ & $0.96$ & $0.60$ & $4.42$ \\
 & T2--T3 & $0.67$ & $0.98$ & $0.70$ & $4.57$ \\
\addlinespace
Ministral-3 & T1--T2 & $0.71$ & $0.98$ & $0.73$ & $4.16$ \\
 & T1--T3 & $0.60$ & $0.95$ & $0.59$ & $4.32$ \\
 & T2--T3 & $0.69$ & $0.98$ & $0.72$ & $4.64$ \\
\midrule
\multicolumn{6}{l}{\emph{Image sources}} \\
Qwen3-VL & AffectNet--RAF-DB & $0.65$ & $0.92$ & $0.43$ & $3.19$ \\
 & AffectNet--Synth-P & $0.68$ & $0.96$ & $0.65$ & $4.18$ \\
 & AffectNet--Synth-S & $0.44$ & $0.95$ & $0.42$ & $3.83$ \\
 & RAF-DB--Synth-P & $0.47$ & $0.86$ & $0.27$ & $2.08$ \\
 & RAF-DB--Synth-S & $0.40$ & $0.92$ & $0.35$ & $2.59$ \\
 & Synth-P--Synth-S & $0.52$ & $0.93$ & $0.50$ & $3.75$ \\
\addlinespace
InternVL3.5 & AffectNet--RAF-DB & $0.57$ & $0.90$ & $0.36$ & $3.02$ \\
 & AffectNet--Synth-P & $0.61$ & $0.97$ & $0.66$ & $4.15$ \\
 & AffectNet--Synth-S & $0.36$ & $0.94$ & $0.39$ & $3.37$ \\
 & RAF-DB--Synth-P & $0.44$ & $0.87$ & $0.28$ & $2.33$ \\
 & RAF-DB--Synth-S & $0.34$ & $0.93$ & $0.33$ & $2.72$ \\
 & Synth-P--Synth-S & $0.46$ & $0.94$ & $0.47$ & $3.84$ \\
\addlinespace
Ministral-3 & AffectNet--RAF-DB & $0.55$ & $0.89$ & $0.37$ & $2.77$ \\
 & AffectNet--Synth-P & $0.55$ & $0.94$ & $0.59$ & $3.95$ \\
 & AffectNet--Synth-S & $0.30$ & $0.95$ & $0.46$ & $3.65$ \\
 & RAF-DB--Synth-P & $0.40$ & $0.85$ & $0.30$ & $2.88$ \\
 & RAF-DB--Synth-S & $0.36$ & $0.89$ & $0.33$ & $2.75$ \\
 & Synth-P--Synth-S & $0.51$ & $0.95$ & $0.59$ & $4.28$ \\
\bottomrule
\end{tabular}
\end{table}

\paragraph{Text vectors show more consistent correspondence across sources than image vectors.}
Table~\ref{tab:within-modal} shows consistent correspondence across the three story sets. Cosine similarity before alignment ranges from $0.60$ to $0.71$, and LOEO cosine from $0.59$ to $0.73$. All comparisons exceed the permutation threshold ($z=4.09$--$4.64$). Thus, the relationship learned from five emotions also predicts the sixth when the story generator changes.

Image-source comparisons vary more, with cosine similarity before alignment ranging from $0.30$ to $0.68$. However, the three models rank the six image-source pairs almost identically by cosine similarity (Spearman $\rho=0.94$--$1.00$). AffectNet and synthetic portraits show strong held-out correspondence in every model ($z=3.95$--$4.18$), while several pairs involving RAF-DB are weaker. The repeated ordering across models suggests that these differences depend strongly on the stimulus sources, consistent with the set-level comparison in Appendix~\ref{app:setlevel}.

\paragraph{Held-out alignment distinguishes generalization from fit to known pairs.}
To determine whether high fitted similarity also predicts an unseen pair, we compare full-fit and LOEO cosine in Table~\ref{tab:within-modal}. Full-fit cosine is high for all image-source pairs ($0.85$--$0.97$). Yet RAF-DB pairs give lower LOEO cosine ($0.27$--$0.43$) than the strongest comparisons without RAF-DB. All nine RAF-DB comparisons still exceed the permutation threshold ($z=2.08$--$3.19$). Their correspondence is therefore weaker, but the other five emotions still predict the held-out pair better than shuffled labels do.

\begin{figure}[t]
\centering
\includegraphics[width=\linewidth]{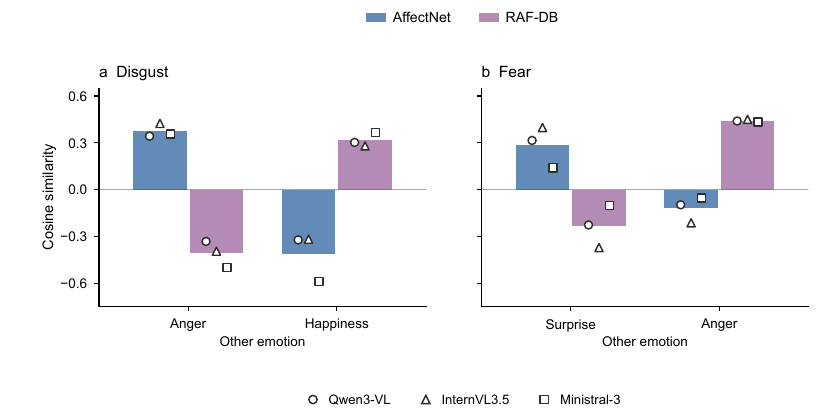}
\caption{Source differences in the neighbors of disgust (a) and fear (b). Cosine similarities are computed between emotion vectors from AffectNet or RAF-DB. Bars show the mean over three models, and marker shapes identify individual models.}
\label{fig:raf-distance}
\end{figure}

\paragraph{RAF-DB changes the neighbors of disgust and fear.}
To explore which emotion relationships differ, Figure~\ref{fig:raf-distance} compares the neighbors of disgust and fear. Across the three models, AffectNet places disgust closer to anger and fear closer to surprise, whereas RAF-DB places disgust closer to happiness and fear closer to anger. The retrieval-label patterns in Table~\ref{tab:topk-composition} provide related evidence. These results suggest that the two datasets emphasize different visual forms of the same labeled emotions, although they do not identify the visual features responsible for the shift.

Changes in a few emotion relationships can reduce RSA and held-out correspondence even when cosine similarity before alignment remains moderate. In a separate classification check, RAF-DB-derived probes classify $3{,}000$ AffectNet images balanced across the six target classes. Accuracy is $0.38$--$0.42$, above the uniform-chance reference of $0.167$, while disgust and fear recall remain below each model's overall accuracy. These observations support a source-specific change in emotion relationships rather than a uniform loss of emotion information.

\subsection{Correspondence Between Text- and Image-Derived Vectors}
\label{app:sim-modal}

We next explore whether text- and image-derived vectors correspond within the same model. For each architecture, Table~\ref{tab:cross-modal} compares each of the three story sources with each of the four image sources. This changes both modality and source, so the within-modality results above provide a reference for interpreting the differences.

\begin{table}[htbp]
\centering
\small
\setlength{\tabcolsep}{5pt}
\caption{Text--image correspondence within each model. Cosines are means over six emotions. Full-fit uses all six pairs. LOEO fits five and evaluates the sixth. The $z$ score compares LOEO with the refitted label-permutation null. Null controls replace image vectors with matched-norm directions. Source labels follow Table~\ref{tab:stimuli}.}
\label{tab:cross-modal}
\begin{tabular}{@{}lrrrr@{}}
\toprule
 & \multicolumn{3}{c}{Mean cosine} & \multicolumn{1}{c}{Permutation} \\
\cmidrule(lr){2-4}
Text $\to$ image source & \shortstack{Before\\alignment} & Full-fit & LOEO & $z$ \\
\midrule
\multicolumn{5}{l}{\emph{Qwen3-VL}} \\
T1 $\to$ AffectNet & $0.26$ & $0.97$ & $0.33$ & $4.34$ \\
T1 $\to$ RAF-DB & $0.26$ & $0.94$ & $0.21$ & $2.65$ \\
T1 $\to$ Synth-P & $0.30$ & $0.94$ & $0.38$ & $4.61$ \\
T1 $\to$ Synth-S & $0.26$ & $0.97$ & $0.34$ & $4.29$ \\
\addlinespace[2pt]
T2 $\to$ AffectNet & $0.21$ & $0.98$ & $0.40$ & $3.77$ \\
T2 $\to$ RAF-DB & $0.20$ & $0.90$ & $0.12$ & $1.07$ \\
T2 $\to$ Synth-P & $0.27$ & $0.98$ & $0.53$ & $4.21$ \\
T2 $\to$ Synth-S & $0.24$ & $0.94$ & $0.29$ & $2.62$ \\
\addlinespace[2pt]
T3 $\to$ AffectNet & $0.25$ & $0.98$ & $0.43$ & $4.34$ \\
T3 $\to$ RAF-DB & $0.24$ & $0.89$ & $0.20$ & $2.01$ \\
T3 $\to$ Synth-P & $0.31$ & $0.98$ & $0.53$ & $4.18$ \\
T3 $\to$ Synth-S & $0.28$ & $0.95$ & $0.36$ & $3.34$ \\
\midrule
\multicolumn{5}{l}{\emph{InternVL3.5}} \\
T1 $\to$ AffectNet & $0.27$ & $0.97$ & $0.33$ & $4.27$ \\
T1 $\to$ RAF-DB & $0.25$ & $0.94$ & $0.20$ & $2.77$ \\
T1 $\to$ Synth-P & $0.28$ & $0.95$ & $0.33$ & $4.36$ \\
T1 $\to$ Synth-S & $0.21$ & $0.97$ & $0.33$ & $4.14$ \\
\addlinespace[2pt]
T2 $\to$ AffectNet & $0.20$ & $0.98$ & $0.43$ & $3.70$ \\
T2 $\to$ RAF-DB & $0.20$ & $0.89$ & $0.08$ & $0.89$ \\
T2 $\to$ Synth-P & $0.25$ & $0.97$ & $0.45$ & $3.83$ \\
T2 $\to$ Synth-S & $0.20$ & $0.95$ & $0.31$ & $2.78$ \\
\addlinespace[2pt]
T3 $\to$ AffectNet & $0.24$ & $0.97$ & $0.40$ & $3.85$ \\
T3 $\to$ RAF-DB & $0.23$ & $0.89$ & $0.16$ & $1.83$ \\
T3 $\to$ Synth-P & $0.29$ & $0.98$ & $0.44$ & $3.75$ \\
T3 $\to$ Synth-S & $0.22$ & $0.95$ & $0.34$ & $3.15$ \\
\midrule
\multicolumn{5}{l}{\emph{Ministral-3}} \\
T1 $\to$ AffectNet & $0.21$ & $0.96$ & $0.29$ & $3.77$ \\
T1 $\to$ RAF-DB & $0.19$ & $0.94$ & $0.20$ & $2.73$ \\
T1 $\to$ Synth-P & $0.25$ & $0.91$ & $0.34$ & $4.48$ \\
T1 $\to$ Synth-S & $0.23$ & $0.95$ & $0.33$ & $4.22$ \\
\addlinespace[2pt]
T2 $\to$ AffectNet & $0.16$ & $0.97$ & $0.36$ & $3.49$ \\
T2 $\to$ RAF-DB & $0.15$ & $0.90$ & $0.08$ & $0.80$ \\
T2 $\to$ Synth-P & $0.22$ & $0.94$ & $0.43$ & $4.26$ \\
T2 $\to$ Synth-S & $0.19$ & $0.93$ & $0.29$ & $2.64$ \\
\addlinespace[2pt]
T3 $\to$ AffectNet & $0.20$ & $0.98$ & $0.42$ & $3.98$ \\
T3 $\to$ RAF-DB & $0.19$ & $0.89$ & $0.17$ & $1.81$ \\
T3 $\to$ Synth-P & $0.27$ & $0.98$ & $0.50$ & $4.06$ \\
T3 $\to$ Synth-S & $0.23$ & $0.95$ & $0.36$ & $3.04$ \\
\midrule
\multicolumn{5}{l}{\emph{Null controls for Qwen3-VL, T2 $\to$ AffectNet}} \\
Random directions & $-0.01$ & $0.86$ & $0.00$ & $-0.58$ \\
Orthogonal directions & $-0.01$ & $0.86$ & $0.00$ & $-0.45$ \\
\bottomrule
\end{tabular}
\end{table}

\paragraph{Most source pairs retain held-out correspondence despite low cosine similarity.}
Table~\ref{tab:cross-modal} shows low cosine similarity before alignment across all $36$ text--image comparisons ($0.15$--$0.31$). However, for AffectNet and the two synthetic image sources, LOEO cosine reaches $0.29$--$0.53$, and all $27$ comparisons exceed $z=2$. Thus, the relationships among five emotions can predict the sixth across modalities even when matched text and image vectors have low similarity in their original coordinates.

The controls show why fitting and evaluating the same emotion pairs is insufficient. Full-fit cosine reaches $0.89$--$0.98$ for the text--image pairs, but also reaches $0.86$ after replacing image vectors with matched-norm random or orthogonal directions. These replacement controls instead give near-zero LOEO cosine. The held-out result and its label-permutation control therefore provide the evidence for emotion correspondence.

\paragraph{Correspondence and causal transfer test different properties.}
The LOEO result concerns relationships among emotion vectors. It does not establish whether a text-derived vector can act on image representations without alignment. Appendix~\ref{app:causal-modal} tests that question directly by injecting text-derived vectors at image-token positions (Tables~\ref{tab:steer-modal} and~\ref{tab:steer-modal-permodel}).

\paragraph{RAF-DB correspondence is weaker within and across modalities.}
To explore whether weaker text--image correspondence is specific to the modality change, we compare RAF-DB across both settings. Its text--image LOEO cosine is $0.08$--$0.21$, with only four of nine comparisons above $z=2$ ($z=0.80$--$2.77$). RAF-DB also gives weaker image--image correspondence in Table~\ref{tab:within-modal}, although all those comparisons exceed the threshold. The reduction therefore appears within as well as across modalities, consistent with the source differences in Appendix~\ref{app:sim-within}.

\paragraph{The same source ordering appears across architectures.}
Averaging LOEO cosine over the three story sources gives the same broad ordering in every model. Synthetic portraits rank highest and RAF-DB lowest, with AffectNet and synthetic scenes between them. Together with the within-modality results, this suggests that stimulus source helps explain which emotion relationships transfer across text and images.

\subsection{Cross-Architecture Correspondence and Alignment}
\label{app:sim-arch}

We further explore whether held-out emotion correspondence survives changes in model architecture. Table~\ref{tab:xmodel} compares Ministral-3 with Qwen3-VL, a pair with nearly zero native cosine, across all seven stimulus sources. We then test whether mappings learned from paired image activations can recover correspondence without fitting to the six emotion vectors.

\begin{table}[tbp]
\centering
\small
\setlength{\tabcolsep}{5pt}
\caption{Ministral-3$\rightarrow$Qwen3-VL correspondence. Cosines are means over six emotions. Before alignment, cosine similarity is measured in native coordinates. Full-fit and LOEO use six and five emotion pairs, respectively. RAF-DB controls replace source or target vectors with matched-norm directions. The $z$ score compares LOEO with the refitted label-permutation null. Source labels follow Table~\ref{tab:stimuli}.}
\label{tab:xmodel}
\begin{tabular}{@{}lrrrr@{}}
\toprule
 & \multicolumn{3}{c}{Mean cosine} & \multicolumn{1}{c}{Permutation} \\
\cmidrule(lr){2-4}
Source / control & \shortstack{Before\\alignment} & Full-fit & LOEO & $z$ \\
\midrule
T1 & $0.00$ & $1.00$ & $0.13$ & $3.32$ \\
T2 & $0.00$ & $1.00$ & $0.34$ & $3.39$ \\
T3 & $0.00$ & $1.00$ & $0.38$ & $3.73$ \\
\addlinespace
AffectNet & $0.00$ & $0.99$ & $0.30$ & $3.16$ \\
RAF-DB & $-0.01$ & $0.99$ & $0.23$ & $2.60$ \\
Synth-P & $0.00$ & $0.98$ & $0.45$ & $3.63$ \\
Synth-S & $0.00$ & $0.99$ & $0.36$ & $3.28$ \\
\midrule
\multicolumn{5}{l}{\emph{RAF-DB null controls}} \\
Random (target) & $-0.01$ & $0.87$ & $0.00$ & $-0.58$ \\
Random (source) & $0.01$ & $0.87$ & $0.02$ & $1.92$ \\
Orthogonal (target) & $-0.01$ & $0.87$ & $0.00$ & $-0.58$ \\
Orthogonal (source) & $0.01$ & $0.87$ & $0.02$ & $1.92$ \\
\bottomrule
\end{tabular}
\end{table}

\paragraph{Cross-architecture correspondence survives near-zero cosine similarity.}
Table~\ref{tab:xmodel} shows near-zero cosine similarity before alignment for every source. Nevertheless, LOEO cosine is positive for all seven sources ($0.13$--$0.45$), and all exceed the permutation threshold ($z=2.60$--$3.73$). Random and orthogonal replacement controls give near-zero held-out cosine. Thus, emotion relationships learned from five pairs can generalize to the sixth despite differences in model coordinates.

Full-fit cosine is again less informative because replacement controls already reach $0.87$. We therefore use LOEO and its permutation control to assess correspondence beyond the emotion pairs used to fit the map.

\begin{figure}[tbp]
\centering
\includegraphics[width=\linewidth]{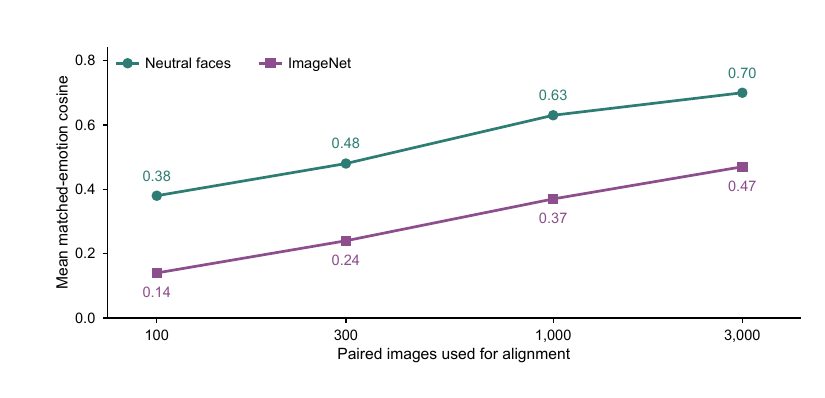}
\caption{Alignment sample size and Ministral-3$\rightarrow$Qwen3-VL correspondence. Maps use paired neutral-face or ImageNet activations without emotion-vector fitting. Points show mean cosine across six AffectNet emotion pairs. Sample size is on a logarithmic axis. Reference cosine similarities are $0.00$ before alignment, $0.30$ with LOEO, and $0.99$ with a full fit.}
\label{fig:alignment-sample-size}
\end{figure}

\paragraph{Learning a mapping without emotion-vector pairs.}
The LOEO results in Tables~\ref{tab:cross-modal} and~\ref{tab:xmodel} show that mappings learned from five matched emotion pairs can generalize to the sixth across modalities and architectures. This motivates us to explore whether we can learn a mapping without using emotion vectors or their labels and then apply it to all six emotion vectors. Following Section~\ref{sec:crossarch}, we use activations from the same neutral-face or ImageNet images in both models to fit an orthogonal transformation (Equation~\ref{eq:neutralmap}). We vary the number of paired images used for fitting and apply each resulting mapping unchanged to all six AffectNet emotion vectors (Figure~\ref{fig:alignment-sample-size}).

\paragraph{Emotion correspondence improves with more paired images.}
As shown in Figure~\ref{fig:alignment-sample-size}, neutral-face mapping raises mean cosine from $0.38$ with $100$ paired images to $0.70$ with $3{,}000$. To test whether this recovery depends on images from an emotion dataset, we also fit the mapping with ImageNet. Mean cosine increases from $0.14$ with $100$ paired images to $0.47$ with $3{,}000$. Both mappings improve correspondence at every tested sample size, although neutral faces give higher cosine.

Together, these results suggest that a mapping learned independently of the six emotion vectors can recover their correspondence, even when the fitting images come from outside the emotion datasets. This supports the view that differences in model coordinates account for part of the low cosine similarity before alignment. We next test whether this recovered correspondence also supports causal transfer by injecting the mapped vectors into the target model (Table~\ref{tab:steer-arch} in Appendix~\ref{app:causal-arch}).

\begin{table}[tbp]
\centering
\small
\setlength{\tabcolsep}{4pt}
\caption{Per-emotion LOEO cosine for Ministral-3$\rightarrow$Qwen3-VL. Each cell evaluates the named emotion after fitting the other five. Source labels follow Table~\ref{tab:stimuli}.}
\label{tab:xmodel-peremo}
\begin{tabular}{@{}lrrrrrr@{}}
\toprule
Stimulus source & Anger & Disgust & Fear & Happiness & Sadness & Surprise \\
\midrule
T1 & $0.10$ & $0.11$ & $0.21$ & $0.15$ & $0.12$ & $0.07$ \\
T2 & $0.50$ & $0.54$ & $0.37$ & $0.25$ & $0.17$ & $0.20$ \\
T3 & $0.53$ & $0.55$ & $0.44$ & $0.31$ & $0.21$ & $0.24$ \\
\addlinespace
AffectNet & $0.26$ & $0.28$ & $0.46$ & $0.31$ & $0.21$ & $0.31$ \\
RAF-DB & $0.36$ & $0.26$ & $0.41$ & $0.35$ & $0.01$ & $0.00$ \\
Synth-P & $0.64$ & $0.66$ & $0.56$ & $0.24$ & $0.34$ & $0.23$ \\
Synth-S & $0.33$ & $0.32$ & $0.64$ & $0.46$ & $0.18$ & $0.23$ \\
\bottomrule
\end{tabular}
\end{table}

\paragraph{Held-out correspondence is distributed across emotions.}
To check whether a single emotion drives the mean, Table~\ref{tab:xmodel-peremo} reports each held-out cosine. Several emotions contribute positive correspondence in every source. However, strength varies by emotion, and RAF-DB sadness and surprise are close to zero. The permutation result therefore supports correspondence for the six-emotion set as a whole, without implying equally strong correspondence for every emotion. 

\paragraph{Activation-based mapping and five-emotion alignment provide complementary information.}
We next explore whether combining activation-based mapping with emotion-based LOEO improves held-out correspondence. We first apply a fixed neutral-face or ImageNet mapping. Within each LOEO fold, we then fit an additional orthogonal transformation on five emotion pairs and evaluate the sixth. The minimum-change completion in Appendix~\ref{app:metrics} keeps this additional transformation as close to the identity as the fit allows. Both vector sets use the same stimulus source, either AffectNet or T2. For each of $500$ label permutations, we refit all six LOEO folds while keeping the initial activation-based mapping fixed.

\begin{table}[tbp]
\centering
\small
\setlength{\tabcolsep}{4pt}
\caption{LOEO after activation-based mapping from Ministral-3 to Qwen3-VL. Direct and After LOEO give mean cosine over six emotions. Held-out range gives the six LOEO scores. The $z$ score uses refitted label permutations. Each activation-based map uses $3{,}000$ paired images. No mapping gives ordinary LOEO. A random map rotates the real source vectors, whereas the final control replaces them with target-orthogonal random directions.}
\label{tab:map-loeo}
\begin{tabular}{@{}lrrrr@{}}
\toprule
 & \multicolumn{2}{c}{Mean cosine} &  &  \\
\cmidrule(lr){2-3}
Initial condition & Direct & After LOEO & Held-out range & Permutation $z$ \\
\midrule
\multicolumn{5}{l}{\emph{Image vectors (AffectNet)}} \\
No mapping & $0.000$ & $0.305$ & $0.206$--$0.456$ & $3.16$ \\
Neutral-face map & $0.697$ & $0.755$ & $0.694$--$0.827$ & $4.37$ \\
ImageNet map & $0.466$ & $0.584$ & $0.503$--$0.675$ & $4.71$ \\
\addlinespace
Random orthogonal map & $0.020$ & $0.316$ & $0.190$--$0.463$ & $3.25$ \\
Orthogonal random directions & $0.000$ & $-0.005$ & $-0.014$--$0.008$ & $-0.57$ \\
\midrule
\multicolumn{5}{l}{\emph{Text vectors (T2)}} \\
No mapping & $0.001$ & $0.338$ & $0.167$--$0.539$ & $3.39$ \\
Neutral-face map & $0.209$ & $0.464$ & $0.297$--$0.634$ & $4.45$ \\
ImageNet map & $0.287$ & $0.533$ & $0.374$--$0.682$ & $4.82$ \\
\bottomrule
\end{tabular}
\end{table}

Table~\ref{tab:map-loeo} shows higher held-out cosine after combining the two fits. For AffectNet, additional LOEO raises cosine from $0.697$ to $0.755$ after neutral-face mapping and from $0.466$ to $0.584$ after ImageNet mapping, compared with $0.305$ for LOEO alone. T2 shows the same pattern, increasing from $0.209$ to $0.464$ and from $0.287$ to $0.533$, compared with $0.338$ for LOEO alone. Paired image activations therefore provide useful information that the five-emotion fit can build on. The combined procedure uses five emotion pairs and tests the excluded sixth. Only the initial activation-based map is fitted without emotion pairs.

The controls separate the role of the initial map from the relationships among the emotion vectors. A random orthogonal map followed by LOEO gives $0.316$ on AffectNet, close to $0.305$ for LOEO alone. This rotation preserves the source vectors' relationships. Replacing those vectors with independently sampled, matched-norm directions orthogonal to their corresponding targets instead gives near-zero LOEO cosine ($-0.005$). The combined fit improves held-out correspondence under this procedure. However, it does not change the source vectors' internal geometry, because orthogonal maps preserve within-source angles.

\paragraph{The need for mapping depends on the model pair.}
Ministral-3 and Qwen3-VL provide a clear test of mapping because their native emotion vectors have nearly zero cosine. InternVL3.5 uses a Qwen3-based language backbone and already has higher native similarity with Qwen3-VL. Appendix~\ref{app:causal-arch} includes this pair to test whether mapping also helps when the native vectors already correspond.

\section{Causal Steering}
\label{app:causal}

Appendix~\ref{app:sim} compares emotion vectors across sources, modalities, and architectures. We now explore whether their causal effects also transfer. We first test image-derived vectors within each model, then substitute text-derived vectors and vectors from other models. Further controls test effects on other readouts, generated continuations, injection positions, removal of the emotion-vector span, and intervention depth.

The main steering comparisons use a separate screening run to select held-out images that the target model classifies correctly, capped at $50$ per class across the six emotions and neutral. We use the same selected images when comparing interventions within a target model. This gives $274$ RAF-DB and $283$ AffectNet images for Qwen3-VL, $249$ and $243$ for InternVL3.5, and $253$ and $227$ for Ministral-3. Each intervention run recomputes its unsteered baseline, so measured baseline accuracy can differ from $100\%$. The controls below state any changes to the evaluation subset or readout.

\subsection{Within-Model Steering with Image-Derived Vectors}
\label{app:causal-within}

We first explore whether image-derived emotion vectors selectively change the same model's emotion predictions. We inject at all image-token positions. We report the mean change in the matched emotion's log-probability as diag and the mean change over the other five emotions as off. Neutral is a readout label but is excluded from both averages.

\paragraph{Image emotion vectors selectively steer the readout in both directions.}
To test whether the response follows intervention strength and sign, Table~\ref{tab:steer-within} reports the Qwen3-VL dose response. At $s=+0.5$, the matched emotion's log-probability increases by $5.34$ on RAF-DB and $4.96$ on AffectNet, while competing emotions decrease on average. Negative injection reverses this pattern, and the smaller interventions at $s=\pm0.1$ give smaller effects with the same signs. This response supports a selective effect on the injected emotion.

We also inject matched-norm random and emotion-orthogonal directions to test whether an arbitrary perturbation produces the same response. Their mean effects have absolute values of at most $0.31$ and do not show the same separation between matched and competing emotions. The comparison therefore concerns selective steering, rather than requiring every control effect to be exactly zero. Figure~\ref{fig:steering-dose} summarizes these effects alongside the text-vector results in Appendix~\ref{app:causal-modal}.

\begin{figure}[tbp]
\centering
\includegraphics[width=\linewidth]{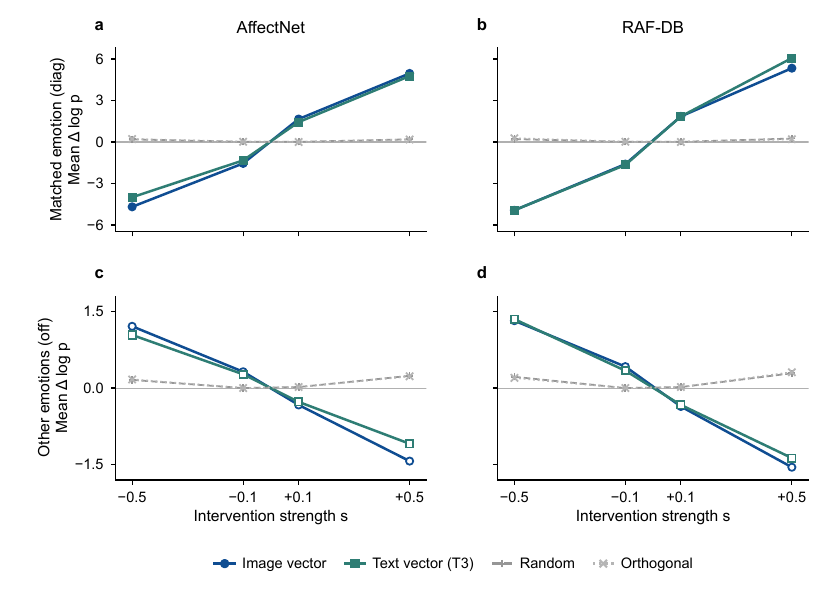}
\caption{Image- and text-vector dose responses in Qwen3-VL. Columns compare AffectNet and RAF-DB. The upper row shows mean changes in the injected emotion's log-probability (diag), and the lower row shows the mean over the other five emotions (off). The two rows use different vertical scales. Points show the means in Tables~\ref{tab:steer-within} and~\ref{tab:steer-modal}; lines connect tested strengths. Image and T3 text vectors are injected at image-token positions. Random and orthogonal controls have matched norms.}
\label{fig:steering-dose}
\end{figure}

\begin{table}[tbp]
\centering
\small
\setlength{\tabcolsep}{4pt}
\caption{Within-model steering for Qwen3-VL at image-token positions. diag and off are mean changes in log-probability for the injected emotion and the other five emotions, respectively. Random and orthogonal controls have matched norms.}
\label{tab:steer-within}
\begin{tabular}{@{}lrrrrrrrr@{}}
\toprule
 & \multicolumn{2}{c}{$s=-0.5$} & \multicolumn{2}{c}{$s=-0.1$} & \multicolumn{2}{c}{$s=+0.1$} & \multicolumn{2}{c}{$s=+0.5$} \\
\cmidrule(lr){2-3}\cmidrule(lr){4-5}\cmidrule(lr){6-7}\cmidrule(lr){8-9}
Injected vector & diag & off & diag & off & diag & off & diag & off \\
\midrule
\multicolumn{9}{l}{\emph{RAF-DB}} \\
Image vector & $-4.94$ & $+1.32$ & $-1.60$ & $+0.42$ & $+1.85$ & $-0.36$ & $+5.34$ & $-1.55$ \\
Random & $+0.21$ & $+0.22$ & $+0.01$ & $+0.00$ & $+0.01$ & $+0.02$ & $+0.25$ & $+0.29$ \\
Orthogonal & $+0.27$ & $+0.20$ & $+0.02$ & $+0.00$ & $+0.00$ & $+0.02$ & $+0.20$ & $+0.31$ \\
\midrule
\multicolumn{9}{l}{\emph{AffectNet}} \\
Image vector & $-4.69$ & $+1.21$ & $-1.55$ & $+0.32$ & $+1.66$ & $-0.33$ & $+4.96$ & $-1.43$ \\
Random & $+0.19$ & $+0.16$ & $+0.01$ & $+0.00$ & $+0.01$ & $+0.02$ & $+0.19$ & $+0.24$ \\
Orthogonal & $+0.21$ & $+0.17$ & $+0.02$ & $+0.00$ & $+0.00$ & $+0.02$ & $+0.18$ & $+0.23$ \\
\bottomrule
\end{tabular}
\end{table}

\begin{table}[htbp]
\centering
\small
\setlength{\tabcolsep}{4pt}
\caption{Within-model steering across the three core models. diag and off are mean changes in log-probability for the injected emotion and the other five emotions. Controls have matched norms.}
\label{tab:steer-permodel}
\begin{tabular}{@{}lrrrrrrrr@{}}
\toprule
 & \multicolumn{4}{c}{RAF-DB} & \multicolumn{4}{c}{AffectNet} \\
\cmidrule(lr){2-5}\cmidrule(lr){6-9}
 & \multicolumn{2}{c}{$s=-0.5$} & \multicolumn{2}{c}{$s=+0.5$} & \multicolumn{2}{c}{$s=-0.5$} & \multicolumn{2}{c}{$s=+0.5$} \\
\cmidrule(lr){2-3}\cmidrule(lr){4-5}\cmidrule(lr){6-7}\cmidrule(lr){8-9}
Injected vector & diag & off & diag & off & diag & off & diag & off \\
\midrule
\multicolumn{9}{l}{\emph{Qwen3-VL}} \\
Image vector & $-4.94$ & $+1.32$ & $+5.34$ & $-1.55$ & $-4.69$ & $+1.21$ & $+4.96$ & $-1.43$ \\
Random & $+0.21$ & $+0.22$ & $+0.25$ & $+0.29$ & $+0.19$ & $+0.16$ & $+0.19$ & $+0.24$ \\
Orthogonal & $+0.27$ & $+0.20$ & $+0.20$ & $+0.31$ & $+0.21$ & $+0.17$ & $+0.18$ & $+0.23$ \\
\midrule
\multicolumn{9}{l}{\emph{InternVL3.5}} \\
Image vector & $-1.93$ & $+0.51$ & $+1.88$ & $-0.40$ & $-2.07$ & $+0.45$ & $+2.19$ & $-0.52$ \\
Random & $+0.02$ & $+0.05$ & $+0.07$ & $+0.07$ & $+0.01$ & $+0.06$ & $+0.11$ & $+0.11$ \\
Orthogonal & $+0.04$ & $+0.05$ & $+0.05$ & $+0.07$ & $+0.03$ & $+0.06$ & $+0.09$ & $+0.11$ \\
\midrule
\multicolumn{9}{l}{\emph{Ministral-3}} \\
Image vector & $-0.88$ & $+0.24$ & $+1.25$ & $-0.21$ & $-0.97$ & $+0.22$ & $+1.29$ & $-0.19$ \\
Random & $-0.01$ & $+0.04$ & $+0.05$ & $-0.02$ & $-0.01$ & $+0.03$ & $+0.02$ & $-0.02$ \\
Orthogonal & $0.00$ & $+0.04$ & $+0.04$ & $-0.02$ & $0.00$ & $+0.03$ & $+0.02$ & $-0.02$ \\
\bottomrule
\end{tabular}
\end{table}

\begin{figure}[t]
\centering
\includegraphics[width=\linewidth]{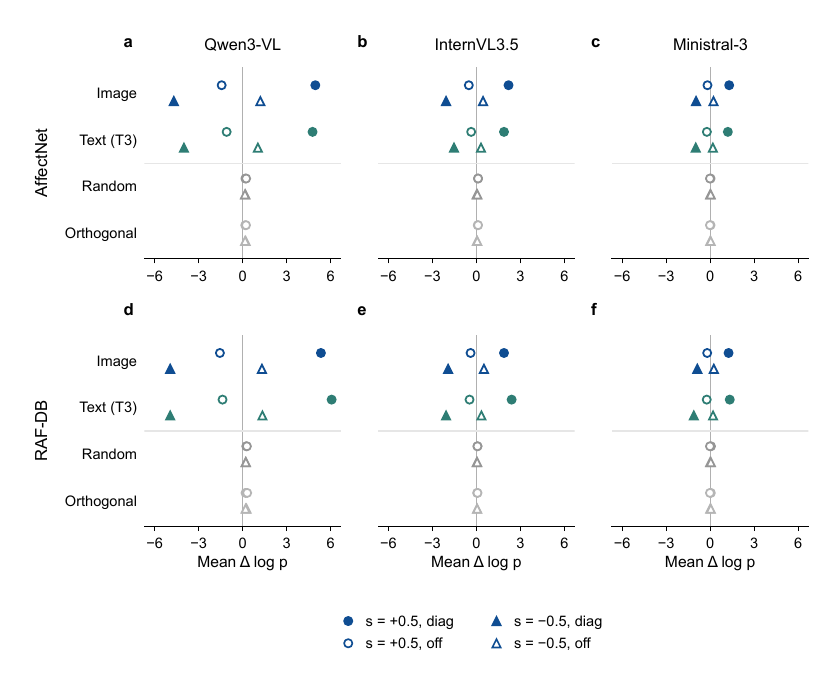}
\caption{Within-model steering across three models and two image datasets. Image and T3 text vectors are compared with matched-norm random and orthogonal controls. The same horizontal scale is used throughout. Filled and open markers show diag and off; circles and triangles show $s=+0.5$ and $s=-0.5$. Points show the means in Tables~\ref{tab:steer-permodel} and~\ref{tab:steer-modal-permodel}.}
\label{fig:steering-models}
\end{figure}

\paragraph{The selective effect replicates across architectures.}
To test whether the pattern depends on Qwen3-VL, we repeat the intervention in InternVL3.5 and Ministral-3. Figure~\ref{fig:steering-models} shows the same selective sign reversal in both models and datasets, with exact values in Table~\ref{tab:steer-permodel}. Absolute effects are smaller than in Qwen3-VL, but positive injection still raises the matched emotion and suppresses competing emotions on average. The random and orthogonal controls do not reproduce this pattern.

\subsection{Cross-Modal Steering with Text-Derived Vectors}
\label{app:causal-modal}

\paragraph{Text-derived emotion vectors retain causal function in the image stream.}
To test transfer across modalities, we replace the image-derived vector with the same model's T3 text-derived vector and keep the rest of the protocol unchanged. No cross-modal alignment is applied. Qwen3-VL's T3 vectors have cosine similarity of only $0.24$ with RAF-DB vectors and $0.25$ with AffectNet vectors (Table~\ref{tab:cross-modal}). Nevertheless, Figure~\ref{fig:steering-dose} shows selective steering in both directions, with exact values in Table~\ref{tab:steer-modal}. At $s=+0.5$, text-derived vectors raise the matched emotion's log-probability by $6.07$ on RAF-DB and $4.77$ on AffectNet, close to the corresponding image-vector effects.

\begin{table}[htbp]
\centering
\small
\setlength{\tabcolsep}{4pt}
\caption{Steering Qwen3-VL with its T3 text-derived vectors at image-token positions, without cross-modal alignment. diag and off are mean changes in log-probability for the injected emotion and the other five emotions.}
\label{tab:steer-modal}
\begin{tabular}{@{}lrrrrrrrr@{}}
\toprule
 & \multicolumn{2}{c}{$s=-0.5$} & \multicolumn{2}{c}{$s=-0.1$} & \multicolumn{2}{c}{$s=+0.1$} & \multicolumn{2}{c}{$s=+0.5$} \\
\cmidrule(lr){2-3}\cmidrule(lr){4-5}\cmidrule(lr){6-7}\cmidrule(lr){8-9}
Evaluation set & diag & off & diag & off & diag & off & diag & off \\
\midrule
RAF-DB & $-4.94$ & $+1.35$ & $-1.66$ & $+0.34$ & $+1.85$ & $-0.33$ & $+6.07$ & $-1.37$ \\
AffectNet & $-4.00$ & $+1.04$ & $-1.33$ & $+0.27$ & $+1.44$ & $-0.27$ & $+4.77$ & $-1.09$ \\
\bottomrule
\end{tabular}
\end{table}

\paragraph{Text-derived vectors also steer images in the other two models.}
We repeat the text-vector substitution in InternVL3.5 and Ministral-3. Figure~\ref{fig:steering-models} shows the same bidirectional selective pattern across both datasets, with effects comparable to those of each model's image-derived vectors (Table~\ref{tab:steer-modal-permodel}). Together with the LOEO results in Appendix~\ref{app:sim-modal}, this shows that low cosine similarity can coexist with both cross-modal emotion relationships and similar causal effects.

\begin{table}[htbp]
\centering
\small
\setlength{\tabcolsep}{4pt}
\caption{Text- and image-vector steering within each model. T3 text vectors are injected at image-token positions without alignment. Image vectors provide the within-model reference. diag and off are mean changes in log-probability for the injected emotion and the other five emotions.}
\label{tab:steer-modal-permodel}
\begin{tabular}{@{}lrrrrrrrr@{}}
\toprule
 & \multicolumn{4}{c}{RAF-DB} & \multicolumn{4}{c}{AffectNet} \\
\cmidrule(lr){2-5}\cmidrule(lr){6-9}
 & \multicolumn{2}{c}{$s=-0.5$} & \multicolumn{2}{c}{$s=+0.5$} & \multicolumn{2}{c}{$s=-0.5$} & \multicolumn{2}{c}{$s=+0.5$} \\
\cmidrule(lr){2-3}\cmidrule(lr){4-5}\cmidrule(lr){6-7}\cmidrule(lr){8-9}
Injected vector & diag & off & diag & off & diag & off & diag & off \\
\midrule
\multicolumn{9}{l}{\emph{Qwen3-VL}} \\
Text vector (T3) & $-4.94$ & $+1.35$ & $+6.07$ & $-1.37$ & $-4.00$ & $+1.04$ & $+4.77$ & $-1.09$ \\
Image vector & $-4.94$ & $+1.32$ & $+5.34$ & $-1.55$ & $-4.69$ & $+1.21$ & $+4.96$ & $-1.43$ \\
\midrule
\multicolumn{9}{l}{\emph{InternVL3.5}} \\
Text vector (T3) & $-2.07$ & $+0.34$ & $+2.40$ & $-0.47$ & $-1.53$ & $+0.31$ & $+1.88$ & $-0.36$ \\
Image vector & $-1.93$ & $+0.51$ & $+1.88$ & $-0.40$ & $-2.07$ & $+0.45$ & $+2.19$ & $-0.52$ \\
\midrule
\multicolumn{9}{l}{\emph{Ministral-3}} \\
Text vector (T3) & $-1.12$ & $+0.19$ & $+1.33$ & $-0.24$ & $-0.99$ & $+0.18$ & $+1.20$ & $-0.23$ \\
Image vector & $-0.88$ & $+0.24$ & $+1.25$ & $-0.21$ & $-0.97$ & $+0.22$ & $+1.29$ & $-0.19$ \\
\bottomrule
\end{tabular}
\end{table}

\subsection{Cross-Architecture Steering with Mapped Vectors}
\label{app:causal-arch}

\begin{figure}[p]
\centering
\includegraphics[width=\linewidth]{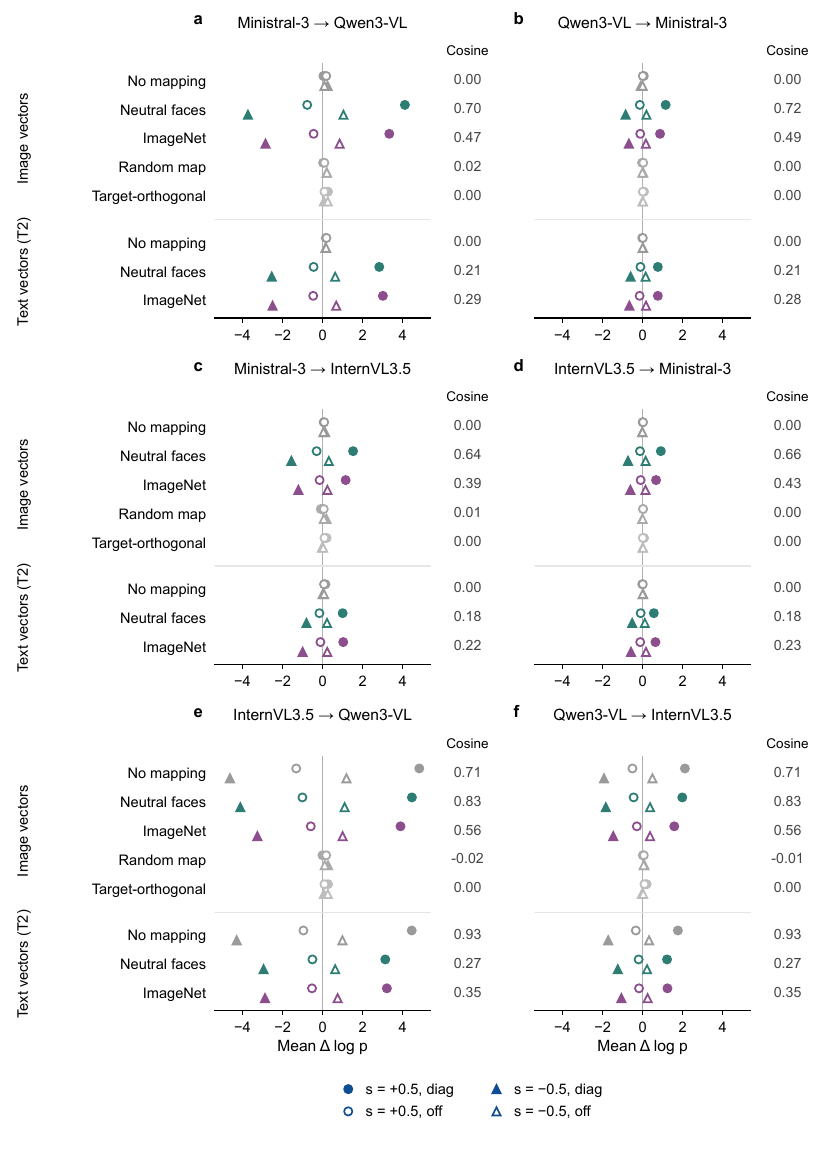}
\caption{Cross-architecture steering on AffectNet for all six directed model pairs. Image and T2 text vectors are compared under no mapping, neutral-face mapping, and ImageNet mapping, with random orthogonal maps and target-orthogonal directions as image-vector controls. Filled/open markers show diag/off, and circles/triangles show positive/negative injection at $|s|=0.5$. All comparisons use the same effect scale. The right-hand columns give cosine similarity to the target vectors of the same modality and source. Values are the means reported in Table~\ref{tab:steer-arch}.}
\label{fig:steering-architecture}
\end{figure}

We next explore whether mappings that recover vector correspondence also transfer causal effects across architectures. We use the activation-based mappings in Appendix~\ref{app:sim-arch}, learned without the six emotion vectors or their labels. Figure~\ref{fig:steering-architecture} compares direct injection with neutral-face and ImageNet mapping for all six ordered model pairs, with exact values in Table~\ref{tab:steer-arch}. We test both image-derived and T2 text-derived vectors. 

\begin{table}[htbp]
\centering
\small
\setlength{\tabcolsep}{4pt}
\setlength{\belowrulesep}{2pt}
\caption{Cross-architecture steering on AffectNet. Source image or T2 text vectors are injected at the target's image-token positions. Cosine compares mapped vectors with target vectors of the same modality and source. diag and off are mean $\Delta\log p$ for the injected emotion and the other five emotions. Neutral-face and ImageNet maps use neither the six emotion vectors nor their labels.}
\label{tab:steer-arch}
\begin{tabular}{@{}llrrrrr@{}}
\toprule
 &  &  & \multicolumn{2}{c}{$s=-0.5$} & \multicolumn{2}{c}{$s=+0.5$} \\
\cmidrule(lr){4-5}\cmidrule(lr){6-7}
Vector source & Mapping / control & Cosine & diag & off & diag & off \\
\midrule
\multicolumn{7}{l}{\emph{Ministral-3 $\to$ Qwen3-VL}} \\
Image & No mapping & $0.00$ & $+0.25$ & $+0.09$ & $+0.04$ & $+0.18$ \\
 & Random orthogonal map & $0.02$ & $+0.21$ & $+0.21$ & $+0.02$ & $+0.09$ \\
 & Target-orthogonal directions & $0.00$ & $+0.06$ & $+0.26$ & $+0.27$ & $+0.09$ \\
 & Neutral-face map & $0.70$ & $-3.73$ & $+1.04$ & $+4.11$ & $-0.77$ \\
 & ImageNet map & $0.47$ & $-2.85$ & $+0.85$ & $+3.33$ & $-0.45$ \\
Text (T2) & No mapping & $0.00$ & $+0.14$ & $+0.17$ & $+0.16$ & $+0.19$ \\
 & Neutral-face map & $0.21$ & $-2.54$ & $+0.63$ & $+2.83$ & $-0.45$ \\
 & ImageNet map & $0.29$ & $-2.50$ & $+0.68$ & $+3.01$ & $-0.47$ \\
\midrule
\multicolumn{7}{l}{\emph{Qwen3-VL $\to$ Ministral-3}} \\
Image & No mapping & $0.00$ & $-0.09$ & $+0.00$ & $+0.06$ & $+0.01$ \\
 & Random orthogonal map & $0.00$ & $+0.03$ & $-0.01$ & $-0.03$ & $+0.03$ \\
 & Target-orthogonal directions & $0.00$ & $-0.01$ & $+0.02$ & $+0.07$ & $+0.01$ \\
 & Neutral-face map & $0.72$ & $-0.85$ & $+0.19$ & $+1.15$ & $-0.14$ \\
 & ImageNet map & $0.49$ & $-0.69$ & $+0.16$ & $+0.87$ & $-0.12$ \\
Text (T2) & No mapping & $0.00$ & $+0.02$ & $+0.00$ & $-0.02$ & $+0.01$ \\
 & Neutral-face map & $0.21$ & $-0.60$ & $+0.15$ & $+0.76$ & $-0.11$ \\
 & ImageNet map & $0.28$ & $-0.66$ & $+0.17$ & $+0.76$ & $-0.15$ \\
\midrule
\multicolumn{7}{l}{\emph{Ministral-3 $\to$ InternVL3.5}} \\
Image & No mapping & $0.00$ & $+0.12$ & $+0.04$ & $+0.05$ & $+0.08$ \\
 & Random orthogonal map & $0.01$ & $+0.19$ & $+0.06$ & $-0.09$ & $+0.06$ \\
 & Target-orthogonal directions & $0.00$ & $-0.04$ & $+0.03$ & $+0.20$ & $+0.09$ \\
 & Neutral-face map & $0.64$ & $-1.56$ & $+0.31$ & $+1.52$ & $-0.30$ \\
 & ImageNet map & $0.39$ & $-1.21$ & $+0.24$ & $+1.15$ & $-0.15$ \\
Text (T2) & No mapping & $0.00$ & $0.00$ & $+0.07$ & $+0.14$ & $+0.07$ \\
 & Neutral-face map & $0.18$ & $-0.81$ & $+0.22$ & $+1.00$ & $-0.16$ \\
 & ImageNet map & $0.22$ & $-1.00$ & $+0.23$ & $+1.03$ & $-0.11$ \\
\midrule
\multicolumn{7}{l}{\emph{InternVL3.5 $\to$ Ministral-3}} \\
Image & No mapping & $0.00$ & $-0.02$ & $-0.01$ & $+0.00$ & $+0.02$ \\
 & Random orthogonal map & $0.00$ & $-0.02$ & $0.00$ & $+0.02$ & $+0.02$ \\
 & Target-orthogonal directions & $0.00$ & $-0.01$ & $+0.02$ & $+0.07$ & $+0.01$ \\
 & Neutral-face map & $0.66$ & $-0.73$ & $+0.15$ & $+0.91$ & $-0.13$ \\
 & ImageNet map & $0.43$ & $-0.61$ & $+0.14$ & $+0.67$ & $-0.10$ \\
Text (T2) & No mapping & $0.00$ & $+0.03$ & $0.00$ & $-0.02$ & $+0.01$ \\
 & Neutral-face map & $0.18$ & $-0.52$ & $+0.11$ & $+0.56$ & $-0.10$ \\
 & ImageNet map & $0.23$ & $-0.58$ & $+0.17$ & $+0.64$ & $-0.12$ \\
\midrule
\multicolumn{7}{l}{\emph{InternVL3.5 $\to$ Qwen3-VL}} \\
Image & No mapping & $0.71$ & $-4.63$ & $+1.19$ & $+4.83$ & $-1.32$ \\
 & Random orthogonal map & $-0.02$ & $+0.27$ & $+0.11$ & $-0.01$ & $+0.18$ \\
 & Target-orthogonal directions & $0.00$ & $+0.06$ & $+0.26$ & $+0.27$ & $+0.09$ \\
 & Neutral-face map & $0.83$ & $-4.11$ & $+1.10$ & $+4.46$ & $-1.01$ \\
 & ImageNet map & $0.56$ & $-3.26$ & $+1.00$ & $+3.89$ & $-0.59$ \\
Text (T2) & No mapping & $0.93$ & $-4.29$ & $+0.99$ & $+4.45$ & $-0.96$ \\
 & Neutral-face map & $0.27$ & $-2.95$ & $+0.63$ & $+3.13$ & $-0.51$ \\
 & ImageNet map & $0.35$ & $-2.88$ & $+0.75$ & $+3.21$ & $-0.53$ \\
\midrule
\multicolumn{7}{l}{\emph{Qwen3-VL $\to$ InternVL3.5}} \\
Image & No mapping & $0.71$ & $-1.93$ & $+0.49$ & $+2.11$ & $-0.51$ \\
 & Random orthogonal map & $-0.01$ & $+0.09$ & $+0.05$ & $+0.01$ & $+0.06$ \\
 & Target-orthogonal directions & $0.00$ & $-0.04$ & $+0.03$ & $+0.20$ & $+0.09$ \\
 & Neutral-face map & $0.83$ & $-1.84$ & $+0.37$ & $+1.98$ & $-0.45$ \\
 & ImageNet map & $0.56$ & $-1.47$ & $+0.37$ & $+1.58$ & $-0.29$ \\
Text (T2) & No mapping & $0.93$ & $-1.72$ & $+0.32$ & $+1.76$ & $-0.34$ \\
 & Neutral-face map & $0.27$ & $-1.24$ & $+0.22$ & $+1.22$ & $-0.20$ \\
 & ImageNet map & $0.35$ & $-1.06$ & $+0.25$ & $+1.24$ & $-0.18$ \\
\bottomrule
\end{tabular}
\end{table}

\paragraph{Neutral-face mapping restores selective transfer for Ministral-3 pairs.}
The four ordered pairs involving Ministral-3 test transfer when native cosine is nearly zero. As shown in Table~\ref{tab:steer-arch}, direct injection gives little selective effect. Neutral-face mapping raises image-vector cosine to $0.64$--$0.72$ and restores bidirectional steering. For Ministral-3$\rightarrow$Qwen3-VL at $s=+0.5$, the matched-emotion effect rises from $0.04$ without mapping to $4.11$, while the competing-emotion mean changes from $+0.18$ to $-0.77$. Negative injection reverses the selective response.

\paragraph{ImageNet mapping also restores selective transfer.}
To test whether transfer requires alignment images from an emotion dataset, we fit the maps to $3{,}000$ generic ImageNet images. Across the four Ministral-3 pairings, mapped image-vector cosine is lower than with neutral faces ($0.39$--$0.49$ versus $0.64$--$0.72$), but the selective sign reversal remains. Thus, paired images outside the emotion datasets also provide a useful map for causal transfer.

\paragraph{Mappings learned from images also transfer text-derived vectors.}
We further test transfer across both architectures and modalities by applying the image-based maps to text-derived vectors. Across the four Ministral-3 pairings, mapped text-vector cosine is only $0.18$--$0.29$, yet Table~\ref{tab:steer-arch} shows selective steering under both injection signs. The causal effect therefore transfers even when correspondence between individual mapped and target vectors remains modest.

\paragraph{Mapping does not improve every model pair.}
Qwen3-VL and InternVL3.5 test whether mapping helps when native vectors already correspond. Before alignment, their image vectors have cosine similarity $0.71$ and their T2 text vectors have cosine similarity $0.93$. Direct injection already gives selective steering in both directions. Neutral-face mapping increases image-vector cosine but does not strengthen steering. Both maps reduce the cosine and steering strength of the text vectors. These results show that higher mapped cosine does not always imply a larger causal effect, and that mapping can be unnecessary for an already aligned pair.

Across the six ordered pairs, mapping is most useful when native vector similarity and direct transfer are weak. The recovered effects support shared causal function across models, while the differences between mapping conditions show that geometric similarity and steering strength remain distinct measures.

\subsection{Specificity, Coherence, and Injection Position}
\label{app:causal-specificity}

We next test whether steering selectively changes emotion-related predictions while preserving other model behavior. We compare emotion and gender readouts, check generated continuations for repetition, and vary the injection positions. These controls address different possible explanations for the observed change in emotion log-probabilities.

\paragraph{Emotion steering largely preserves the gender readout.}
To test effects on an unrelated visual attribute, we keep the $s=+0.5$ image-token intervention and change the readout question from emotion to gender. Both tasks use the RAF-DB images. We measure prediction flips relative to each task's unsteered baseline and accuracy against its dataset labels.

Table~\ref{tab:steer-specificity} shows a much larger change in emotion predictions than in gender predictions. Averaged over injections, emotion predictions change on $47\%$ of images and accuracy falls from $0.982$ to $0.527$. Gender predictions change on only $2\%$, with accuracy remaining close to baseline ($0.961$ to $0.950$). This supports selectivity relative to the gender readout, although the two reported averages use slightly different evaluation subsets.

\begin{table}[tbp]
\centering
\small
\setlength{\tabcolsep}{5pt}
\caption{Emotion and gender readouts under Qwen3-VL steering at $s=+0.5$. Flip rate (\%) compares predictions with the unsteered baseline. Accuracy is a fraction, and Mean averages the six injections.}
\label{tab:steer-specificity}
\begin{tabular}{@{}lrrrr@{}}
\toprule
 & \multicolumn{2}{c}{Emotion readout} & \multicolumn{2}{c}{Gender readout} \\
\cmidrule(lr){2-3}\cmidrule(lr){4-5}
Injected emotion & Flip rate & Accuracy & Flip rate & Accuracy \\
\midrule
No intervention & --- & $0.982$ & --- & $0.961$ \\
\addlinespace
Anger & $79\%$ & $0.204$ & $2\%$ & $0.969$ \\
Disgust & $13\%$ & $0.883$ & $3\%$ & $0.937$ \\
Fear & $54\%$ & $0.460$ & $0\%$ & $0.961$ \\
Happiness & $38\%$ & $0.628$ & $4\%$ & $0.925$ \\
Sadness & $62\%$ & $0.365$ & $0\%$ & $0.957$ \\
Surprise & $38\%$ & $0.620$ & $1\%$ & $0.949$ \\
\midrule
Mean & $\mathbf{47\%}$ & $\mathbf{0.527}$ & $\mathbf{2\%}$ & $\mathbf{0.950}$ \\
\bottomrule
\end{tabular}
\end{table}

\begin{figure}[htbp]
\centering
\footnotesize
% Emphasis colors and commands are local to this figure.
\definecolor{EPUnsteered}{HTML}{216E39}
\definecolor{EPSteered}{HTML}{AD3333}
\newcommand{\epbaseemotion}[1]{\textcolor{EPUnsteered}{\textbf{#1}}}
\newcommand{\epsteeremotion}[1]{\textcolor{EPSteered}{\textbf{#1}}}
\newcommand{\epbasecue}[1]{\textcolor{EPUnsteered}{\textit{#1}}}
\newcommand{\epsteercue}[1]{\textcolor{EPSteered}{\textit{#1}}}
\setlength{\tabcolsep}{5pt}
\renewcommand{\arraystretch}{1.15}
\begin{tabular}{@{}m{0.115\linewidth} m{0.75\linewidth}@{}}
\toprule
Image & Continuation, unsteered above and steered below \\
\midrule
\includegraphics[width=\linewidth]{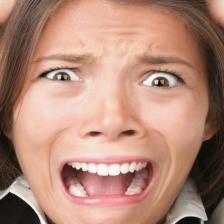} &
\epbaseemotion{none}: extreme \epbaseemotion{fear}, \epbaseemotion{shock}, or \epbaseemotion{terror}. Here's what in the image shows it: \epbasecue{Wide, bulging eyes}: The eyes are opened very wide, with\ldots \newline \epsteeremotion{anger}: extreme \epsteeremotion{anger} or \epsteeremotion{rage}. Here's what in the image shows it: Eyes: They are wide open and staring intensely, which can indicate \epsteeremotion{shock}, \epsteeremotion{fear}, or extreme \epsteeremotion{anger}. Eyebrows: The \epsteercue{eyebrows are drawn together and furrowed}, creating deep lines on the\ldots \\
\midrule
\includegraphics[width=\linewidth]{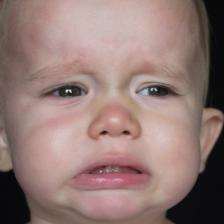} &
\epbaseemotion{none}: \epbaseemotion{sad} or \epbaseemotion{upset}, and the image shows this through their \epbasecue{tear-streaked face, downturned mouth, and furrowed brow}. \newline \epsteeremotion{fear}: \epsteeremotion{scared} or \epsteeremotion{frightened}. The image shows this through several classic signs of \epsteeremotion{fear} in a child: - \epsteercue{Wide, dilated pupils} (visible in the eyes) --- this is a physiological response to \epsteeremotion{fear} or \epsteeremotion{surprise}, as the body prepares to react. - \epsteercue{Tense facial muscles}\ldots \\
\midrule
\includegraphics[width=\linewidth]{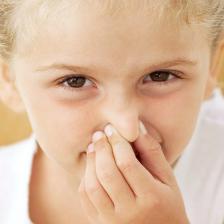} &
\epbaseemotion{none}: \epbaseemotion{uncomfortable}, \epbaseemotion{irritated}, or possibly \epbaseemotion{sick}. Here's what in the image shows this: 1. The Action: The child is actively\ldots \newline \epsteeremotion{happiness}: \epsteeremotion{happy} or \epsteeremotion{content}. Here's what in the image shows it: Smiling Mouth: The person's \epsteercue{lips are curved upwards in a clear, gentle smile}. Glowing Eyes: Their \epsteercue{eyes are bright and crinkled at the corners}, which is often associated with genuine\ldots \\
\midrule
\includegraphics[width=\linewidth]{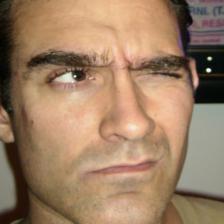} &
\epbaseemotion{none}: \epbaseemotion{annoyed} or \epbaseemotion{irritated}. Here's what in the image shows it: - Eyebrows: Their \epbasecue{eyebrows are drawn together and slightly raised}, creating\ldots \newline \epsteeremotion{sadness}: \epsteeremotion{sad} or \epsteeremotion{disappointed}. The image shows this through their facial expression: their \epsteercue{eyebrows are slightly furrowed}, their \epsteercue{eyes are downcast}, and their \epsteercue{mouth is turned downward in a frown}. These are all classic signs of \epsteeremotion{sadness} or \epsteeremotion{disappointment}. \\
\midrule
\includegraphics[width=\linewidth]{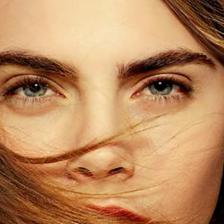} &
\epbaseemotion{none}: \epbaseemotion{intense} and \epbaseemotion{focused}, with a hint of \epbaseemotion{vulnerability} or \epbaseemotion{intensity}. Here's what in the image shows it: - Eyes: Their \epbasecue{eyes are wide open}\ldots \newline \epsteeremotion{surprise}: \epsteeremotion{surprised} or \epsteeremotion{startled}. Here's what in the image shows it: - Wide, Alert Eyes: The most telling feature is the person's eyes. \epsteercue{They are wide open, and the pupils are dilated}, which is a classic physiological response to being \epsteeremotion{startled} or \epsteeremotion{surprised}. The gaze\ldots \\
\midrule
\includegraphics[width=\linewidth]{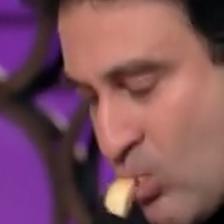} &
\epbaseemotion{none}: \epbaseemotion{content} and \epbaseemotion{satisfied}. The image shows this through the person's \epbasecue{relaxed facial expression} --- their \epbasecue{eyes are gently closed}, and their\ldots \newline \epsteeremotion{disgust}: \epsteeremotion{disgusted} or \epsteeremotion{repulsed}. Here's what in the image shows it: Facial Expression: The person's \epsteercue{eyes are squeezed shut}, and their \epsteercue{nose is wrinkled}. This is a classic, involuntary reaction to something unpleasant or foul-tasting. Action: They are\ldots \\
\bottomrule
\end{tabular}
\caption{Selected Qwen3-VL continuations under its own AffectNet emotion vectors at image-token positions and $s=+0.5$. Each block pairs a different validation image with one injected emotion and shows unsteered text above steered text. Green and red highlights mark unsteered and steered text, respectively. Bold marks labels and emotion terms, and italics mark selected visual descriptions. Ellipses mark truncation.}
\label{fig:gen-examples}
\end{figure}

\paragraph{Decoded continuations remain coherent and follow the induced interpretation.}
To test whether steering changes more than the first emotion word, we decode full continuations using Qwen3-VL's image-derived vectors. Across $630$ continuations per dataset, distinct-1, the fraction of unique generated tokens, averages $0.92$ on RAF-DB and $0.93$ on AffectNet and never falls below $0.73$. The longest repeated-token run is one token even at $s=\pm0.5$. These diagnostics show no repetition collapse in the tested outputs.

Figure~\ref{fig:gen-examples} shows selected continuations that name the injected emotion and explain the answer using visual cues. However, the disgust example is a rare success. Across $3{,}500$ tested AffectNet validation images, disgust steering changes the readout to the target on only $32$ images, under $1\%$. The other five emotions succeed on $14$--$60\%$ of the images tested. Thus, a mean log-probability shift does not guarantee frequent changes in generated emotion labels. Disgust also has low classification recall in Table~\ref{tab:topk-purity}, although classification and induced generation measure different abilities.

\paragraph{Steering strength and selectivity depend on injection position.}
To explore how the intervention location affects steering, we keep the vector and strength fixed and vary the token positions. Table~\ref{tab:steer-position} shows that question-token injection gives a weaker matched-emotion effect and does not suppress competing emotions on average. Answer-prefix injection gives a matched effect close to image-token injection, while all-position injection suppresses competing emotions most strongly. The vectors therefore remain effective at several locations, with different strengths and effects on competing labels.

\begin{table}[tbp]
\centering
\small
\setlength{\tabcolsep}{5pt}
\caption{Injection-position comparison for Qwen3-VL at $s=+0.5$, with vectors held fixed. diag and off are mean changes in log-probability for the injected emotion and the other five emotions.}
\label{tab:steer-position}
\begin{tabular}{@{}lrrrr@{}}
\toprule
 & \multicolumn{2}{c}{RAF-DB} & \multicolumn{2}{c}{AffectNet} \\
\cmidrule(lr){2-3}\cmidrule(lr){4-5}
Injection positions & diag & off & diag & off \\
\midrule
Image tokens (default) & $+5.34$ & $-1.55$ & $+4.96$ & $-1.43$ \\
Question tokens & $+2.50$ & $+0.70$ & $+2.27$ & $+0.13$ \\
Answer prefix & $+4.71$ & $-2.31$ & $+4.73$ & $-1.95$ \\
All positions & $+5.23$ & $-5.51$ & $+5.55$ & $-5.57$ \\
\bottomrule
\end{tabular}
\end{table}

We use image-token injection as the default because it produces selective steering and places the intervention at the visual-input positions. This specifies where we change the residual stream, without implying that later effects remain limited to those positions.

\subsection{Removing the Emotion-Vector Span}
\label{app:causal-ablate}

The steering experiments show that adding emotion vectors can change emotion predictions. We next explore whether removing their components from hidden states impairs normal emotion recognition. We compare individual emotion directions with the centered span of the six source-specific vectors. We obtain an orthonormal basis for the span by re-centering the vectors and applying SVD. Each condition is compared with a random control of the same dimension.

Let $\mathbf B$ contain an orthonormal basis for the selected direction or subspace. At the same layer and image-token positions as in the steering experiments, we remove a fraction $f$ of each hidden state's projection onto it,
\begin{equation}
\bm h
\leftarrow
\bm h - f\,\bm h\mathbf B\mathbf B^\top,
\qquad
f\in[0,1].
\label{eq:ablate}
\end{equation}
Here, $f=0.5$ removes half of the projected component, and $f=1$ removes it completely. For single-vector removal, we remove each of the six emotion directions separately and average the resulting accuracies.

\begin{table}[tbp]
\centering
\small
\setlength{\tabcolsep}{5pt}
\caption{Emotion recognition accuracy after partial ($f=0.5$) or full ($f=1$) removal in Qwen3-VL at image-token positions. Classification includes the six emotions and neutral. Each random control matches the dimension of the emotion direction or span directly above it.}
\label{tab:steer-ablate}
\begin{tabular}{@{}lrrrr@{}}
\toprule
 & \multicolumn{2}{c}{RAF-DB} & \multicolumn{2}{c}{AffectNet} \\
\cmidrule(lr){2-3}\cmidrule(lr){4-5}
Removed direction or span & $f=0.5$ & $f=1$ & $f=0.5$ & $f=1$ \\
\midrule
No removal ($f=0$) & \multicolumn{2}{c}{$0.993$} & \multicolumn{2}{c}{$0.986$} \\
\midrule
Single emotion vector (rank 1) & $0.969$ & $0.946$ & $0.950$ & $0.902$ \\
Random direction & $0.991$ & $0.991$ & $0.988$ & $0.987$ \\
\addlinespace
Centered emotion-vector span (rank 5) & $0.938$ & $\mathbf{0.814}$ & $0.901$ & $\mathbf{0.781}$ \\
Random subspace & $0.990$ & $0.988$ & $0.988$ & $0.990$ \\
\bottomrule
\end{tabular}
\end{table}

\paragraph{Removing the emotion-vector span reduces recognition accuracy.}
As shown in Table~\ref{tab:steer-ablate}, full removal of the centered emotion-vector span lowers accuracy from $0.993$ to $0.814$ on RAF-DB and from $0.986$ to $0.781$ on AffectNet. Partial removal and removal of a single emotion direction cause smaller decreases. In contrast, the random controls leave accuracy close to baseline under both removal strengths. This comparison supports a contribution of the emotion-vector span beyond the effect of removing arbitrary directions. However, removing the span does not eliminate emotion recognition. The extracted directions therefore contribute to normal recognition but do not capture all of the information that supports it.

\subsection{Beyond the Emotion Readout}
\label{app:causal-beyond}

We further explore whether emotion steering affects a related prediction that uses no emotion label. We also test whether the induced emotion is associated specifically with the person in the image by changing who the question refers to.

\begin{figure}[t]
\centering
\includegraphics[width=\linewidth]{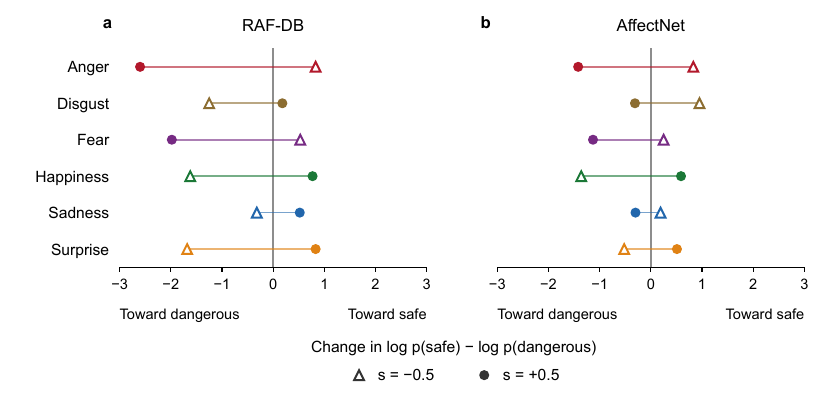}
\caption{Safe--dangerous readout under Qwen3-VL emotion steering. Points show mean changes from the unsteered model, with paired markers for $s=-0.5$ (open triangles) and $s=+0.5$ (filled circles). Negative values shift the prediction toward dangerous and positive values toward safe.}
\label{fig:steering-safe-dangerous}
\end{figure}

\paragraph{Emotion steering also changes the safe--dangerous prediction.}
We replace the emotion question with a binary \emph{safe}--\emph{dangerous} prediction while keeping the image and intervention unchanged. Figure~\ref{fig:steering-safe-dangerous} shows the change in $\log p(\text{safe})-\log p(\text{dangerous})$ relative to the unsteered model, with exact values in Table~\ref{tab:steer-judgment}.

\begin{table}[t]
\centering
\small
\setlength{\tabcolsep}{5pt}
\caption{Change in the safe--dangerous log-probability contrast under Qwen3-VL emotion steering at image-token positions. Positive values favor \emph{safe}, and negative values favor \emph{dangerous}.}
\label{tab:steer-judgment}
\begin{tabular}{@{}lrrrr@{}}
\toprule
 & \multicolumn{2}{c}{RAF-DB} & \multicolumn{2}{c}{AffectNet} \\
\cmidrule(lr){2-3}\cmidrule(lr){4-5}
Injected emotion & $s=-0.5$ & $s=+0.5$ & $s=-0.5$ & $s=+0.5$ \\
\midrule
Anger & $+0.83$ & $-2.60$ & $+0.83$ & $-1.42$ \\
Disgust & $-1.25$ & $+0.18$ & $+0.95$ & $-0.31$ \\
Fear & $+0.53$ & $-1.98$ & $+0.25$ & $-1.13$ \\
Happiness & $-1.62$ & $+0.77$ & $-1.36$ & $+0.59$ \\
Sadness & $-0.32$ & $+0.52$ & $+0.19$ & $-0.30$ \\
Surprise & $-1.68$ & $+0.83$ & $-0.52$ & $+0.51$ \\
\bottomrule
\end{tabular}
\end{table}

Positive anger and fear injections move the prediction toward \emph{dangerous} on both datasets, while happiness moves it toward \emph{safe}. Negative injection reverses these effects. Disgust and sadness show less consistent changes across datasets. Together with the largely unchanged gender readout, this suggests that steering can affect a related prediction beyond the explicit emotion label.

\paragraph{Steering is strongest for the depicted person but also affects other subjects.}
To test who the induced emotion is attributed to, we keep the image and intervention fixed and ask about the depicted person, the assistant, or a third party absent from the image.

\begin{table}[t]
\centering
\small
\setlength{\tabcolsep}{5pt}
\caption{Effect of the queried subject on Qwen3-VL emotion steering at image-token positions. diag and off are mean changes in log-probability for the injected emotion and the other five emotions.}
\label{tab:steer-speaker}
\begin{tabular}{@{}lrrrrrrrr@{}}
\toprule
 & \multicolumn{4}{c}{RAF-DB} & \multicolumn{4}{c}{AffectNet} \\
\cmidrule(lr){2-5}\cmidrule(lr){6-9}
 & \multicolumn{2}{c}{$s=-0.5$} & \multicolumn{2}{c}{$s=+0.5$} & \multicolumn{2}{c}{$s=-0.5$} & \multicolumn{2}{c}{$s=+0.5$} \\
\cmidrule(lr){2-3}\cmidrule(lr){4-5}\cmidrule(lr){6-7}\cmidrule(lr){8-9}
Emotion attributed to & diag & off & diag & off & diag & off & diag & off \\
\midrule
the depicted person & $-3.43$ & $+0.30$ & $\mathbf{+2.63}$ & $-1.10$ & $-2.83$ & $+0.34$ & $\mathbf{+2.37}$ & $-0.83$ \\
the assistant & $-1.94$ & $+0.37$ & $+1.78$ & $-0.53$ & $-1.73$ & $+0.26$ & $+1.61$ & $-0.44$ \\
a third party & $-2.20$ & $+0.31$ & $+1.91$ & $-0.60$ & $-1.68$ & $+0.25$ & $+1.63$ & $-0.38$ \\
\bottomrule
\end{tabular}
\end{table}

Table~\ref{tab:steer-speaker} shows the strongest matched-emotion effects when the question refers to the depicted person. However, steering also changes answers about the assistant and the third party. The effect is therefore partly tied to the visual subject, but can also influence emotion predictions about other subjects in the prompt.

\subsection{Causal Effect Across Depth}
\label{app:causal-depth}

\begin{figure}[t]
\centering
\includegraphics[width=\linewidth]{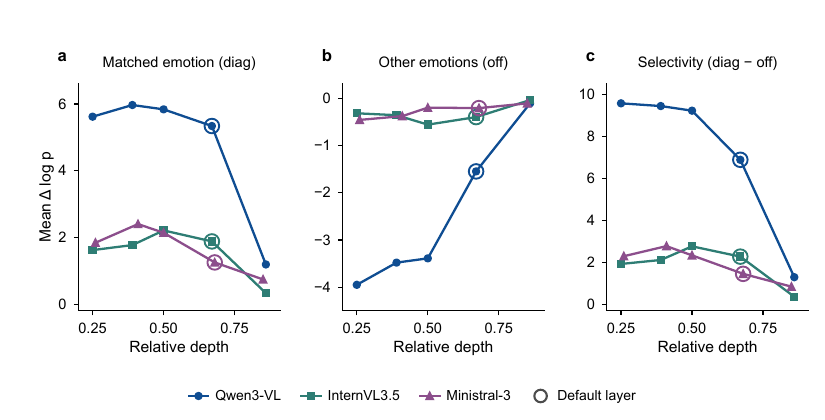}
\caption{Depth dependence of steering on RAF-DB at $s=+0.5$. Mean $\Delta\log p$ is shown for the matched emotion (diag), the other five emotions (off), and their difference. Lines connect measured layers. Rings mark the default layers. All depths use the same residual-norm scaling rule. Table~\ref{tab:steer-depth} gives exact values.}
\label{fig:steering-depth}
\end{figure}

We choose the default intervention layer at approximately the same relative depth as the representational analyses. To explore whether this choice is also where steering is strongest, we repeat the intervention at approximately matched fractional depths in all three models (Figure~\ref{fig:steering-depth}; Table~\ref{tab:steer-depth}).

\begin{table}[tbp]
\centering
\small
\setlength{\tabcolsep}{5pt}
\caption{Steering across depth on RAF-DB at $s=+0.5$. diag and off are mean $\Delta\log p$ for the injected emotion and the other five emotions. Selectivity is diag minus off. Bold marks each model's largest diag and selectivity. Differences use unrounded values.}
\label{tab:steer-depth}
\begin{tabular}{@{}rrrrr@{}}
\toprule
Layer & Relative depth & diag & off & \shortstack{Selectivity\\(diag $-$ off)} \\
\midrule
\multicolumn{5}{l}{\emph{Qwen3-VL}} \\
9 & $0.25$ & $+5.62$ & $-3.95$ & $\mathbf{+9.58}$ \\
14 & $0.39$ & $\mathbf{+5.97}$ & $-3.48$ & $+9.45$ \\
18 & $0.50$ & $+5.84$ & $-3.39$ & $+9.23$ \\
24 & $0.67$ & $+5.34$ & $-1.55$ & $+6.89$ \\
31 & $0.86$ & $+1.19$ & $-0.11$ & $+1.30$ \\
\midrule
\multicolumn{5}{l}{\emph{InternVL3.5}} \\
9 & $0.25$ & $+1.62$ & $-0.32$ & $+1.93$ \\
14 & $0.39$ & $+1.77$ & $-0.36$ & $+2.12$ \\
18 & $0.50$ & $\mathbf{+2.21}$ & $-0.56$ & $\mathbf{+2.77}$ \\
24 & $0.67$ & $+1.88$ & $-0.40$ & $+2.28$ \\
31 & $0.86$ & $+0.33$ & $-0.04$ & $+0.37$ \\
\midrule
\multicolumn{5}{l}{\emph{Ministral-3}} \\
9 & $0.26$ & $+1.84$ & $-0.46$ & $+2.30$ \\
14 & $0.41$ & $\mathbf{+2.40}$ & $-0.38$ & $\mathbf{+2.78}$ \\
17 & $0.50$ & $+2.14$ & $-0.20$ & $+2.34$ \\
23 & $0.68$ & $+1.25$ & $-0.21$ & $+1.46$ \\
29 & $0.85$ & $+0.74$ & $-0.11$ & $+0.85$ \\
\bottomrule
\end{tabular}
\end{table}

\paragraph{Steering is stronger before the default layer and weaker near the output.}
Figure~\ref{fig:steering-depth} shows a broad region of strong effects at relative depths $0.25$--$0.50$ in Qwen3-VL. InternVL3.5 peaks at $0.50$ and Ministral-3 at $0.41$, and all three have weaker effects at the deepest tested layer. At the default layer, selectivity is about $28\%$, $18\%$, and $47\%$ below the respective model maxima. The reported effects therefore do not rely on selecting the strongest intervention layer. Because all depths use residual-norm scaling, these results describe steering under that calibration and do not give a scale-independent ranking of layers.

\section{Shared Emotion Subspace}
\label{app:demix}

The preceding analyses show that emotion vectors can retain similar relationships and causal effects despite differences in source and model coordinates. We now explore whether averaging aligned vectors across models and sources yields a shared emotion subspace. We first construct this subspace, then test how much of the original vectors and their causal effects it retains. Lexical readouts, generated continuations, and held-out Gemma-3 experiments provide further checks.

\subsection{Separating Shared Emotion Structure from Source and Model Variation}
\label{app:decomp}

Let $\bm v_e^{m,q}$ denote the emotion vector for model $m$, stimulus source $q$, and emotion $e\in\mathcal E$, written as a row vector. The matrix $\mathbf V_{m,q}\in\mathbb R^{|\mathcal E|\times D}$ stacks these vectors as rows in the same emotion order for every model and source, where $D$ is the residual width.

We extend the pairwise alignment in Equation~\ref{eq:neutralmap} to the three core models using generalized Procrustes alignment. Let $\widetilde{\mathbf X}_m$ contain centered activations from model $m$, with rows paired across models using the same neutral-face or ImageNet images. For the set of three core models $\mathcal M$, we jointly fit orthogonal transformations $\mathbf R_m$ from each model to a common coordinate frame and a shared activation template $\mathbf G$ by minimizing
\begin{equation}
\min_{\mathbf G,\{\mathbf R_m\}_{m\in\mathcal M}}
\sum_{m\in\mathcal M}
\left\|\widetilde{\mathbf X}_m\mathbf R_m-\mathbf G\right\|_F^2,
\qquad
\mathbf R_m^\top\mathbf R_m=\mathbf I.
\label{eq:gpa}
\end{equation}
We alternate between fitting each transformation to the current template and updating the template to the mean of the aligned activations. This fit uses neither the six emotion vectors nor their labels. We then apply each transformation to the emotion vectors,
\begin{equation}
\widetilde{\mathbf V}_{m,q}
=
\mathbf V_{m,q}\mathbf R_m,
\qquad
\widetilde{\bm v}_e^{m,q}
=
\bm v_e^{m,q}\mathbf R_m,
\quad e\in\mathcal E.
\label{eq:aligned-vectors}
\end{equation}

The consensus matrix $\bar{\mathbf V}=\mathbb E_{m,q}[\widetilde{\mathbf V}_{m,q}]$ is the mean over models and sources. Its row for emotion $e$ is the consensus vector $\bar{\bm v}_e$ from Section~\ref{sec:demix}. We write
\begin{equation}
\widetilde{\mathbf V}_{m,q}
=
\bar{\mathbf V}
+
\boldsymbol{\Delta}_q
+
\boldsymbol{\Delta}_m
+
\mathbf E_{m,q}.
\label{eq:decomp}
\end{equation}
The source term $\boldsymbol{\Delta}_q$ describes each source's difference from this consensus, averaged over models. The model term $\boldsymbol{\Delta}_m$ describes each model's difference, averaged over sources. The remaining term $\mathbf E_{m,q}$ contains source--model interactions and estimation residuals. We define the source and model terms as
\begin{equation}
\boldsymbol{\Delta}_q
=
\mathbb E_m\left[\widetilde{\mathbf V}_{m,q}\right]-\bar{\mathbf V},
\qquad
\boldsymbol{\Delta}_m
=
\mathbb E_q\left[\widetilde{\mathbf V}_{m,q}\right]-\bar{\mathbf V}.
\end{equation}
These definitions give zero mean across sources for $\boldsymbol{\Delta}_q$ and across models for $\boldsymbol{\Delta}_m$. The remaining term then has zero mean over either models or sources, separating the average source and model differences from their interactions.

Source and modality are related because each source contains either text or images. The variance analysis below keeps all model--source--emotion combinations and gives each source equal weight. Total variance is measured before averaging over sources. We obtain the emotion, model, and modality components by averaging over the other factors, and assign the remaining variance to residuals and interactions. This last term includes differences among sources within a modality. These scalar variance components summarize different aspects of the data from the six-emotion source deviations in Equation~\ref{eq:decomp}.

\paragraph{Centered consensus vectors define the shared emotion subspace.}
For each emotion $e\in\mathcal E$, we average its aligned vector over models and stimulus sources,
\begin{equation}
\bar{\bm v}_e
=
\mathbb E_{m,q}
\left[
\widetilde{\bm v}_e^{m,q}
\right],
\qquad
\bar{\bm v}
=
\mathbb E_{e\in\mathcal E}
\left[
\bar{\bm v}_e
\right].
\label{eq:consensus-emotion-vectors}
\end{equation}
We center the rows of $\bar{\mathbf V}$ by subtracting their mean $\bar{\bm v}$ and define the shared emotion subspace as
\begin{equation}
\mathcal{S}_{\mathrm{emo}}
=
\operatorname{span}
\left\{
\bar{\bm v}_e-\bar{\bm v}
\right\}_{e\in\mathcal E}.
\label{eq:semo-app}
\end{equation}
This subspace describes the shared differences among emotions. The six centered vectors sum to zero, which gives $\dim(\mathcal{S}_{\mathrm{emo}})\leq5$. This bound follows from the construction and is not a chosen number of principal components.

When an explicit basis is required for projection, we stack the centered consensus vectors as columns and compute
\begin{equation}
\mathbf W_{\mathrm{emo}}
=
\operatorname{orth}
\left(
\left[
(\bar{\bm v}_e-\bar{\bm v})^\top
\right]_{e\in\mathcal E}
\right).
\label{eq:wemo-basis}
\end{equation}
The columns of $\mathbf W_{\mathrm{emo}}$ are orthonormal and span $\mathcal{S}_{\mathrm{emo}}$, so
\begin{equation}
\operatorname{col}(\mathbf W_{\mathrm{emo}})
=
\mathcal{S}_{\mathrm{emo}}.
\end{equation}
The corresponding orthogonal projector is
\begin{equation}
\mathbf P_{\mathrm{emo}}
=
\mathbf W_{\mathrm{emo}}\mathbf W_{\mathrm{emo}}^\top.
\label{eq:pemo}
\end{equation}

Averaging over models and sources follows the marginalization principle of demixed PCA~\citep{kobak2016demixed}. Here, however, we retain the full span of the centered consensus vectors rather than selecting a number of principal components. We treat this span as a candidate shared emotion subspace and test its geometric, causal, and lexical properties below.

\begin{table}[htbp]
\centering
\small
\setlength{\tabcolsep}{5pt}
\caption{Variance components of emotion vectors across all seven sources, weighted equally. Entries are fractions of total variance before source averaging. Cross-model estimates follow generalized Procrustes alignment. Residual/interactions includes within-modality source variation and interactions among factors. Values may not sum to one after rounding.}
\label{tab:dpca}
\begin{tabular}{@{}lrrrr@{}}
\toprule
Model / alignment & Emotion & Modality & Model & \shortstack{Residual /\\interactions} \\
\midrule
\multicolumn{5}{l}{\emph{Within each model}} \\
Qwen3-VL & $0.474$ & $0.004$ & --- & $0.522$ \\
InternVL3.5 & $0.447$ & $0.004$ & --- & $0.550$ \\
Ministral-3 & $0.423$ & $0.006$ & --- & $0.571$ \\
\midrule
\multicolumn{5}{l}{\emph{Across all three aligned models}} \\
Neutral-face alignment & $0.326$ & $0.003$ & $0.004$ & $0.668$ \\
ImageNet alignment & $0.308$ & $0.002$ & $0.004$ & $0.686$ \\
\bottomrule
\end{tabular}
\end{table}

\paragraph{Shared differences among emotions coexist with source variation.}
To measure the variation shared across conditions, Table~\ref{tab:dpca} separates emotion-related variance from model, modality, and remaining variation. Emotion accounts for $0.42$--$0.47$ of variance within each model. After neutral-face alignment across models, it accounts for $0.326$, compared with $0.004$ for model and $0.003$ for modality. The residual and interaction terms account for $0.668$. Thus, the small variance shares for model and modality describe small average shifts across all emotions. They do not imply that source-specific or emotion-specific differences disappear.

We next measure how much of each aligned source-specific vector lies in the shared span by computing its cosine with its orthogonal projection onto $\mathcal{S}_{\mathrm{emo}}$. Averaged over all seven sources and six emotions, the cosines are $0.64$, $0.59$, and $0.61$ for Qwen3-VL, Ministral-3, and InternVL3.5 under neutral-face alignment. The AffectNet-only values are $0.77$, $0.68$, and $0.76$, respectively. This measures reconstruction of the original vectors by the shared span, not similarity to the consensus vector for each emotion.

We also test whether the shared structure depends on using neutral faces for alignment. ImageNet alignment gives a similar variance pattern, with emotion accounting for $0.308$, model for $0.004$, and residuals and interactions for $0.686$. Mean projection cosine becomes $0.63$, $0.57$, and $0.59$ across all sources, and $0.69$, $0.58$, and $0.66$ on AffectNet, in the same model order. Generic images therefore support the shared subspace, although they give weaker reconstruction, especially for AffectNet vectors.

The consensus vectors also retain agreement with human affective ratings (Figure~\ref{fig:main-pca}e; Appendix~\ref{app:axes}). Against R\&M, PC1--valence $|r|$ is $0.93$ under both alignment conditions. PC2--arousal $|r|$ is $0.96$ with neutral-face alignment and $0.97$ with ImageNet alignment. These results connect the shared emotion relationships to the human affective reference.

\subsection{Causal Localization of the Shared Emotion Subspace}
\label{app:demix-causal}

We next explore whether the shared subspace retains the causal effect of Qwen3-VL's original emotion vectors. We first express its projector in Qwen3-VL's coordinates as $\mathbf P_{\mathrm{emo}}^{\mathrm{Q}}=\mathbf R_{\mathrm{Q}}\mathbf P_{\mathrm{emo}}\mathbf R_{\mathrm{Q}}^\top$. We then split each original vector $\bm v$ into its shared projection and orthogonal remainder,
\begin{equation}
\bm v_{\mathrm{shared}}
=
\bm v \mathbf P_{\mathrm{emo}}^{\mathrm{Q}},
\qquad
\bm v_{\perp}
=
\bm v(\mathbf I-\mathbf P_{\mathrm{emo}}^{\mathrm{Q}}).
\label{eq:emo-residual}
\end{equation}
The shared projection is the part of the original vector inside the mapped shared subspace. The remainder is the part outside it. Neither is the consensus vector itself, which averages aligned vectors across models and sources. The remainder may contain source-dependent and emotion-related information, so we test its effect separately rather than treating it as noise.

\begin{figure}[t]
\centering
\includegraphics[width=\linewidth]{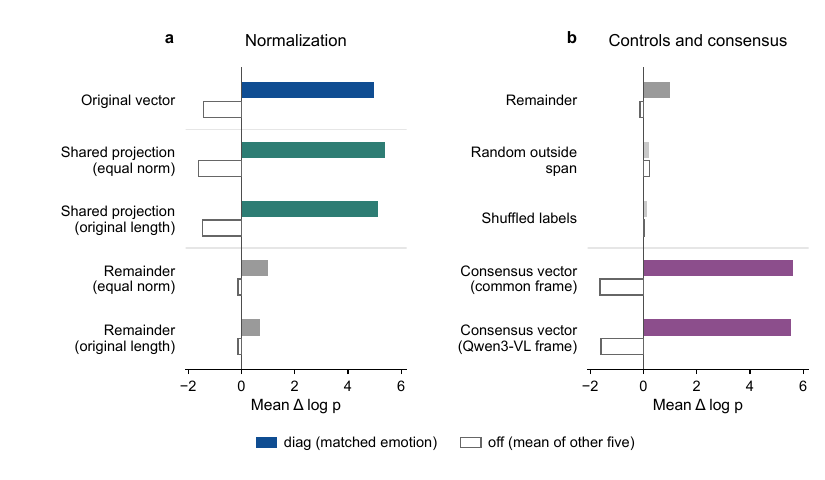}
\caption{Shared-subspace steering in Qwen3-VL on AffectNet at $s=+0.5$. (a) Shared projection and remainder under both norm conditions, compared with the original vector. (b) Equal-norm remainder controls and consensus vectors under two reference frames. The equal-norm remainder is repeated for comparison. Bars show the means in Table~\ref{tab:dpca-causal}; bootstrap intervals for selected comparisons are reported in the text.}
\label{fig:shared-subspace-controls}
\end{figure}

\begin{table}[htbp]
\centering
\small
\setlength{\tabcolsep}{5pt}
\caption{Shared-subspace steering in Qwen3-VL on AffectNet images at $s=+0.5$. The span uses all three core models and seven sources. Equal-norm directions have unit length. The next block retains the original component lengths. Shuffled labels rotate remainder labels. diag and off are mean $\Delta\log p$ for the injected emotion and the other five emotions.}
\label{tab:dpca-causal}
\begin{tabular}{@{}lrr@{}}
\toprule
 & \multicolumn{2}{c}{Mean $\Delta\log p$} \\
\cmidrule(lr){2-3}
Injected vector & diag & off \\
\midrule
\multicolumn{3}{l}{\emph{Equal-norm directions}} \\
Original vector $\bm v$ & $+4.96$ & $-1.43$ \\
Shared projection $\bm v_{\mathrm{shared}}$ & $+5.39$ & $-1.62$ \\
Remainder $\bm v_{\perp}$ & $+1.00$ & $-0.14$ \\
Random outside span $\mathcal{S}_{\mathrm{emo}}$ & $+0.19$ & $+0.22$ \\
Shuffled labels (remainder) & $+0.12$ & $+0.03$ \\
Consensus vector (mapped to Qwen3-VL) & $+5.60$ & $-1.64$ \\
\midrule
\multicolumn{3}{l}{\emph{Original component lengths}} \\
Shared projection $\bm v_{\mathrm{shared}}$ & $+5.14$ & $-1.47$ \\
Remainder $\bm v_{\perp}$ & $+0.68$ & $-0.14$ \\
\midrule
\multicolumn{3}{l}{\emph{Reference-frame control}} \\
Consensus vector (Qwen3-VL reference frame) & $+5.56$ & $-1.60$ \\
\bottomrule
\end{tabular}
\end{table}

\paragraph{The shared projection retains the selective causal effect under both norm conditions.}
To compare the causal effects of the directions themselves, we first normalize each component to unit length. As shown in Figure~\ref{fig:shared-subspace-controls}a and Table~\ref{tab:dpca-causal}, the shared projection gives $(\mathrm{diag},\mathrm{off})=(+5.39,-1.62)$, close to or larger than the original vector's $(+4.96,-1.43)$. The remainder gives a smaller matched effect of $+1.00$ and weaker competing-emotion suppression of $-0.14$. The $95\%$ image-bootstrap intervals for diag are $[5.18,5.61]$, $[4.77,5.16]$, and $[0.94,1.05]$ for the projection, original vector, and remainder. The projection's selectivity exceeds the original vector's by $0.62$, with a paired $95\%$ interval of $[0.59,0.66]$.

Unit normalization changes the lengths of the original components. We therefore repeat the intervention without renormalizing either component. The projection still gives $(\mathrm{diag},\mathrm{off})=(+5.14,-1.47)$, while the remainder gives $(+0.68,-0.14)$. The shared projection therefore retains the selective effect under both norm conditions. Because the model response is nonlinear, the effects of separately injecting the two components need not sum to the original-vector effect.

To test whether the smaller remainder effect depends on emotion identity, we compare it with a random direction in the same orthogonal complement and with remainders assigned rotated emotion labels. At equal norm, these controls give $(\mathrm{diag},\mathrm{off})=(+0.19,+0.22)$ and $(+0.12,+0.03)$, respectively (Figure~\ref{fig:shared-subspace-controls}b). Neither reproduces the selective response of the real remainder. Emotion-related information therefore remains outside the shared span, although its effect is smaller than that of the projection.

\paragraph{Consensus vectors themselves also steer Qwen3-VL.}
We further test the consensus vectors themselves, rather than projections of Qwen3-VL's original vectors. After mapping back to Qwen3-VL, they give $(\mathrm{diag},\mathrm{off})=(+5.60,-1.64)$ in Table~\ref{tab:dpca-causal}. Constructing the consensus with Qwen3-VL as the reference gives a similar result of $(+5.56,-1.60)$, as shown in Figure~\ref{fig:shared-subspace-controls}b. Thus, both constructions produce consensus vectors that retain selective steering.

\subsection{Generated Continuations and Lexical Readouts of Consensus Vectors}
\label{app:demix-logitlens}

\begin{figure}[tbp]
\centering
\footnotesize
% Local emphasis colors match the emotion palette in Figure 2.
\definecolor{EPSharedAnger}{HTML}{B2182B}
\definecolor{EPSharedDisgust}{HTML}{8C6D31}
\definecolor{EPSharedFear}{HTML}{762A83}
\definecolor{EPSharedHappiness}{HTML}{1B7837}
\definecolor{EPSharedSadness}{HTML}{2166AC}
\definecolor{EPSharedSurprise}{HTML}{E08214}
\newcommand{\epsharedemotion}[2]{\textcolor{#1}{\textbf{#2}}}
\newcommand{\epsharedcue}[2]{\textcolor{#1}{\textit{#2}}}
\setlength{\tabcolsep}{5pt}
\renewcommand{\arraystretch}{1.15}
\begin{tabular}{@{}m{0.115\linewidth} l@{}}
\toprule
Image & \begin{tabular}[c]{@{}p{0.10\linewidth} p{0.62\linewidth}@{}}
Injected & Continuation
\end{tabular} \\
\midrule
\includegraphics[width=\linewidth]{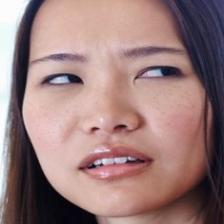} &
\begin{tabular}[c]{@{}p{0.10\linewidth} p{0.62\linewidth}@{}}
\epsharedemotion{black}{none} & \epsharedemotion{black}{annoyed}, \epsharedemotion{black}{skeptical}, or slightly \epsharedemotion{black}{disgusted}. Here's what in the image shows it: 1. Eyebrows: Their \epsharedcue{black}{eyebrows are slightly raised} and drawn\ldots \\[2pt]
\epsharedemotion{EPSharedAnger}{anger} & \epsharedemotion{EPSharedAnger}{angry} or \epsharedemotion{EPSharedAnger}{frustrated}. Here's what in the image shows it: - Eyebrows: Their \epsharedcue{EPSharedAnger}{eyebrows are drawn together and slightly lowered}, a classic sign of \epsharedemotion{EPSharedAnger}{anger} or\ldots \\[2pt]
\epsharedemotion{EPSharedDisgust}{disgust} & \epsharedemotion{EPSharedDisgust}{disgusted} or \epsharedemotion{EPSharedDisgust}{repulsed}. Here's what in the image shows it: - Eyebrow and Eye Expression: Their \epsharedcue{EPSharedDisgust}{eyebrows are slightly furrowed}, and their \epsharedcue{EPSharedDisgust}{eyes are rolled upward} and\ldots \\[2pt]
\epsharedemotion{EPSharedFear}{fear} & \epsharedemotion{EPSharedFear}{scared} or \epsharedemotion{EPSharedFear}{frightened}. Here's what in the image shows it: - \epsharedcue{EPSharedFear}{Wide, darting eyes}: Her \epsharedcue{EPSharedFear}{eyes are wide open and looking sharply to the side} (her right), as if she's\ldots \\[2pt]
\epsharedemotion{EPSharedHappiness}{happiness} & \epsharedemotion{EPSharedHappiness}{amused} or \epsharedemotion{EPSharedHappiness}{amusedly} \epsharedemotion{EPSharedHappiness}{skeptical}, likely at something funny or absurd. Here's what in the image shows it: 1. The Eyes: The most telling feature is the person's eyes.\ldots \\[2pt]
\epsharedemotion{EPSharedSadness}{sadness} & \epsharedemotion{EPSharedSadness}{sad} or \epsharedemotion{EPSharedSadness}{disappointed}. Here's what in the image shows it: - \epsharedcue{EPSharedSadness}{Downward gaze and drooping eyelids}: The person's eyes are looking down and slightly to the side, and\ldots \\[2pt]
\epsharedemotion{EPSharedSurprise}{surprise} & \epsharedemotion{EPSharedSurprise}{surprised} or \epsharedemotion{EPSharedSurprise}{startled}. Here's what in the image shows it: - Eyes: Their \epsharedcue{EPSharedSurprise}{eyes are wide open and looking upward and to the side} (toward the upper right). This is a\ldots \\
\end{tabular} \\
\midrule
\includegraphics[width=\linewidth]{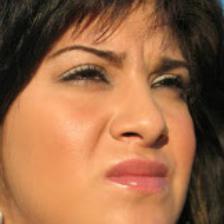} &
\begin{tabular}[c]{@{}p{0.10\linewidth} p{0.62\linewidth}@{}}
\epsharedemotion{black}{none} & \epsharedemotion{black}{annoyed} or \epsharedemotion{black}{frustrated}. Here's what in the image shows it: - Eyebrows: Her \epsharedcue{black}{eyebrows are drawn together and slightly furrowed}, a classic sign of\ldots \\[2pt]
\epsharedemotion{EPSharedAnger}{anger} & \epsharedemotion{EPSharedAnger}{angry} or \epsharedemotion{EPSharedAnger}{frustrated}. Here's what in the image shows it: - Eyebrows: Their \epsharedcue{EPSharedAnger}{eyebrows are drawn together and slightly furrowed}, a classic sign of \epsharedemotion{EPSharedAnger}{anger} or\ldots \\[2pt]
\epsharedemotion{EPSharedDisgust}{disgust} & \epsharedemotion{EPSharedDisgust}{disgusted} or \epsharedemotion{EPSharedDisgust}{repulsed}. Here's what in the image shows it: - Eyebrows: Their \epsharedcue{EPSharedDisgust}{eyebrows are drawn together and slightly raised}, forming a \epsharedcue{EPSharedDisgust}{furrowed brow}. This is a\ldots \\[2pt]
\epsharedemotion{EPSharedFear}{fear} & \epsharedemotion{EPSharedFear}{scared} or \epsharedemotion{EPSharedFear}{frightened}. Here's what in the image shows it: - \epsharedcue{EPSharedFear}{Wide, dilated pupils}: The person's \epsharedcue{EPSharedFear}{eyes are wide open}, and their \epsharedcue{EPSharedFear}{pupils appear very large}. This is a\ldots \\[2pt]
\epsharedemotion{EPSharedHappiness}{happiness} & \epsharedemotion{EPSharedHappiness}{happy} or \epsharedemotion{EPSharedHappiness}{amused}. Here's what in the image shows it: - Smiling mouth: Their \epsharedcue{EPSharedHappiness}{lips are curved upwards in a clear, gentle smile}. - \epsharedcue{EPSharedHappiness}{Crinkled eyes}: The corners\ldots \\[2pt]
\epsharedemotion{EPSharedSadness}{sadness} & \epsharedemotion{EPSharedSadness}{sad} or \epsharedemotion{EPSharedSadness}{upset}. Here's what in the image shows it: - Eyebrows: Their \epsharedcue{EPSharedSadness}{eyebrows are drawn together and slightly furrowed}, a classic sign of \epsharedemotion{EPSharedSadness}{sadness}, \epsharedemotion{EPSharedSadness}{worry}, or\ldots \\[2pt]
\epsharedemotion{EPSharedSurprise}{surprise} & \epsharedemotion{EPSharedSurprise}{surprised} or \epsharedemotion{EPSharedSurprise}{shocked}. Here's what in the image shows it: - Wide, \epsharedemotion{EPSharedSurprise}{startled} eyes: Her \epsharedcue{EPSharedSurprise}{eyes are wide open}, and her \epsharedcue{EPSharedSurprise}{pupils appear dilated}, which is a common\ldots \\
\end{tabular} \\
\bottomrule
\end{tabular}
\caption{Selected continuations for two AffectNet validation images under all six consensus-vector injections in Qwen3-VL at $s=+0.5$. Both images have ground-truth label \emph{disgust}. The first row for each image is unsteered. Highlight colors follow the injected emotions in Figure~\ref{fig:main-pca}, while unsteered highlights remain black. Bold marks labels and emotion terms, and italics mark selected visual descriptions. The leading \emph{They feel} is omitted, and ellipses mark truncation.}
\label{fig:gen-examples-shared}
\end{figure}

\paragraph{Selected continuations follow all six consensus-vector injections.}
To explore whether consensus vectors change generated explanations as well as emotion scores, we apply each of the six injections to the same image. Figure~\ref{fig:gen-examples-shared} shows two images under consensus vectors mapped into Qwen3-VL. For each image, all six continuations follow the injected emotion and describe facial cues consistent with that interpretation. These selected cases show coherent changes in generated interpretation.

\paragraph{Consensus vectors retain emotion-related lexical readouts.}
We also explore whether the consensus vectors retain emotion-related words in each model's vocabulary. We average the $21$ model--source sets in the ImageNet-aligned frame to obtain one consensus vector $\bar{\bm v}_e$ per emotion. We then map each vector back to model $m$ through $\mathbf R_m^\top$ and apply that model's lexical readout, as in Appendix~\ref{app:logitlens}.

Figure~\ref{fig:logitlens-shared} shows emotion-related promoted tokens across all three models, with some opposite-emotion terms among the suppressed tokens. The pattern also appears in Ministral-3 despite its near-zero native cosine with the Qwen-based models before alignment. The coordinate maps use ImageNet activations without emotion labels, while the consensus averages corresponding emotions after mapping. This lexical evidence complements the projection and steering results.

\begin{figure}[t]
\centering
\includegraphics[width=\linewidth]{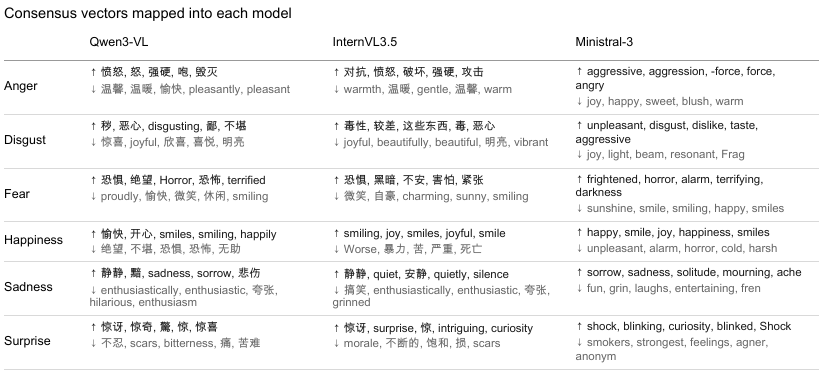}
\caption{Lexical readouts of the three-model consensus vectors. After ImageNet alignment and averaging over all $21$ model--source sets, each vector is mapped back to each model's coordinates. The figure shows the five highest- and lowest-scoring tokens per emotion under that model's lexical readout.}
\label{fig:logitlens-shared}
\end{figure}

\subsection{Held-Out Generalization and Model-Size Ablation}
\label{app:gemma}

The three core models have residual width $4096$, and Qwen3-VL and InternVL3.5 use related language backbones. To test generalization beyond the models used to construct the consensus, we evaluate Gemma-3-12B and Gemma-3-27B. Neither variant contributes to consensus construction. Their different residual widths also require cross-width alignment before matched-vector cosine can be computed.

\begin{figure}[p]
\centering
\includegraphics[width=\linewidth]{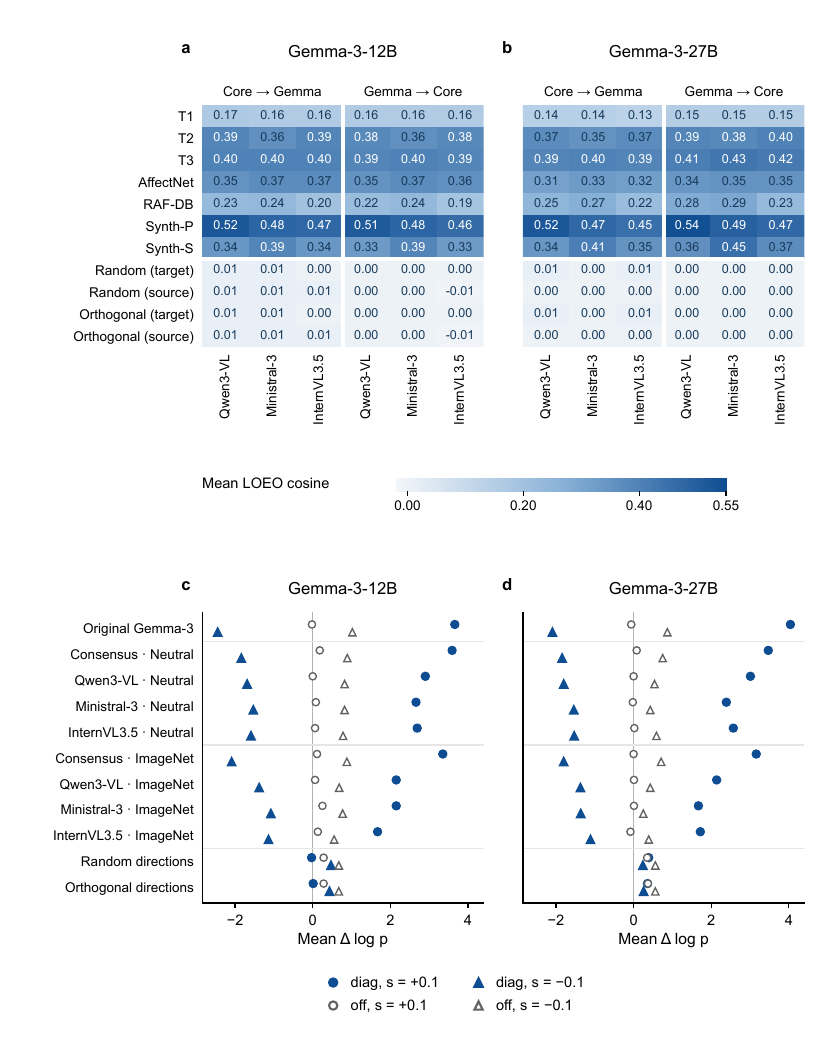}
\caption{Gemma-3 generalization at 12B and 27B. (a,b) Mean six-emotion LOEO cosine for seven sources and both directions with each core model. The last four rows are RAF-DB replacement controls. (c,d) AffectNet steering at $s=\pm0.1$. Filled/open markers show diag/off, and circles/triangles show positive/negative injection. Neutral denotes neutral-face mapping. Gemma-3 is excluded from consensus construction. Tables~\ref{tab:gemma-xmodel} and~\ref{tab:gemma-steer} give exact values.}
\label{fig:gemma-generalization}
\end{figure}

\paragraph{Held-out correspondence persists across unequal residual widths.}
We first explore whether a map fitted on five emotions predicts the sixth across unequal residual widths. Figure~\ref{fig:gemma-generalization}a,b and Table~\ref{tab:gemma-xmodel} report both directions between each Gemma-3 variant and the three core models, across seven sources. For 12B, LOEO cosine ranges from $0.16$ to $0.52$, with median $0.36$. All $42$ model--source comparisons exceed the permutation threshold ($z=2.37$--$3.70$), while all $24$ replacement controls remain below $z=1$. For reference, Ministral-3$\rightarrow$Qwen3-VL gives $0.13$--$0.45$ with median $0.34$ under the same held-out criterion (Table~\ref{tab:xmodel}).

The 27B model shows a similar pattern. LOEO cosine ranges from $0.13$ to $0.54$, again with median $0.36$, and all $42$ comparisons exceed the threshold ($z=2.40$--$3.78$). Its $24$ replacement controls remain below $z=2$, with cosine between $-0.003$ and $0.013$. Held-out correspondence therefore persists at both sizes, without a consistent gain from the larger model.

\begin{table}[t]
\centering
\small
\setlength{\tabcolsep}{3pt}
\caption{Mean six-emotion LOEO cosine between Gemma-3 and each core model in both directions. Unequal-width fits follow Appendix~\ref{app:metrics}. RAF-DB controls replace source or target vectors with matched-norm random or orthogonal directions. Source labels follow Table~\ref{tab:stimuli}.}
\label{tab:gemma-xmodel}
\begin{tabular}{@{}lrrrrrr@{}}
\toprule
 & \multicolumn{3}{c}{Source model $\to$ Gemma-3} & \multicolumn{3}{c}{Gemma-3 $\to$ target model} \\
\cmidrule(lr){2-4}\cmidrule(lr){5-7}
Source / control & Qwen3-VL & Ministral-3 & InternVL3.5 & Qwen3-VL & Ministral-3 & InternVL3.5 \\
\midrule
\multicolumn{7}{l}{\emph{Gemma-3-12B}} \\
T1 & $0.17$ & $0.16$ & $0.16$ & $0.16$ & $0.16$ & $0.16$ \\
T2 & $0.39$ & $0.36$ & $0.39$ & $0.38$ & $0.36$ & $0.38$ \\
T3 & $0.40$ & $0.40$ & $0.40$ & $0.39$ & $0.40$ & $0.39$ \\
\addlinespace[2pt]
AffectNet & $0.35$ & $0.37$ & $0.37$ & $0.35$ & $0.37$ & $0.36$ \\
RAF-DB & $0.23$ & $0.24$ & $0.20$ & $0.22$ & $0.24$ & $0.19$ \\
Synth-P & $0.52$ & $0.48$ & $0.47$ & $0.51$ & $0.48$ & $0.46$ \\
Synth-S & $0.34$ & $0.39$ & $0.34$ & $0.33$ & $0.39$ & $0.33$ \\
\addlinespace
\multicolumn{7}{l}{\emph{RAF-DB null controls}} \\
Random (target) & $0.01$ & $0.01$ & $0.00$ & $0.00$ & $0.00$ & $0.00$ \\
Random (source) & $0.01$ & $0.01$ & $0.01$ & $0.00$ & $0.00$ & $-0.01$ \\
Orthogonal (target) & $0.01$ & $0.01$ & $0.00$ & $0.00$ & $0.00$ & $0.00$ \\
Orthogonal (source) & $0.01$ & $0.01$ & $0.01$ & $0.00$ & $0.00$ & $-0.01$ \\
\midrule
\multicolumn{7}{l}{\emph{Gemma-3-27B}} \\
T1 & $0.14$ & $0.14$ & $0.13$ & $0.15$ & $0.15$ & $0.15$ \\
T2 & $0.37$ & $0.35$ & $0.37$ & $0.39$ & $0.38$ & $0.40$ \\
T3 & $0.39$ & $0.40$ & $0.39$ & $0.41$ & $0.43$ & $0.42$ \\
\addlinespace[2pt]
AffectNet & $0.31$ & $0.33$ & $0.32$ & $0.34$ & $0.35$ & $0.35$ \\
RAF-DB & $0.25$ & $0.27$ & $0.22$ & $0.28$ & $0.29$ & $0.23$ \\
Synth-P & $0.52$ & $0.47$ & $0.45$ & $0.54$ & $0.49$ & $0.47$ \\
Synth-S & $0.34$ & $0.41$ & $0.35$ & $0.36$ & $0.45$ & $0.37$ \\
\addlinespace
\multicolumn{7}{l}{\emph{RAF-DB null controls}} \\
Random (target) & $0.01$ & $0.00$ & $0.01$ & $0.00$ & $0.00$ & $0.00$ \\
Random (source) & $0.00$ & $0.00$ & $0.00$ & $0.00$ & $0.00$ & $0.00$ \\
Orthogonal (target) & $0.01$ & $0.00$ & $0.01$ & $0.00$ & $0.00$ & $0.00$ \\
Orthogonal (source) & $0.00$ & $0.00$ & $0.00$ & $0.00$ & $0.00$ & $0.00$ \\
\bottomrule
\end{tabular}
\end{table}

\paragraph{Consensus vectors correspond to the held-out Gemma-3 vectors.}
We next test the consensus vectors constructed without either Gemma-3 variant. We keep the three-model consensus fixed and learn a cross-width map from paired image activations, without using the six emotion vectors or their labels. Let $\mathbf G$ be the fixed shared activation template from Equation~\ref{eq:gpa}, and let $\widetilde{\mathbf X}_{\mathrm{Gemma}}$ contain centered activations from either Gemma-3 variant on the same neutral-face or ImageNet images. We use a thin singular value decomposition (SVD) to compute the map and transfer each consensus vector,
\begin{equation}
\begin{gathered}
\mathbf G^\top\widetilde{\mathbf X}_{\mathrm{Gemma}}
=\mathbf U\boldsymbol\Sigma\mathbf V^\top,
\qquad
\mathbf R_{\mathrm{Gemma}}=\mathbf U\mathbf V^\top,
\\
\bm v_e^{\mathrm{shared}\rightarrow\mathrm{Gemma}}
=\bar{\bm v}_e\mathbf R_{\mathrm{Gemma}},
\qquad e\in\mathcal E.
\end{gathered}
\label{eq:gemma-map}
\end{equation}
Gemma-3 activations are used only for this mapping step, and transferred vectors are normalized before steering.

Table~\ref{tab:gemma-steer} reports mean cosine with Gemma-3's own AffectNet vectors. Neutral-face and ImageNet maps give $0.63$ and $0.42$ for 12B, and $0.59$ and $0.39$ for 27B. The consensus therefore retains correspondence with models that did not contribute to its construction.

\paragraph{Consensus vectors retain causal effects in Gemma-3-12B.}
To test whether this correspondence also supports causal transfer, we inject Gemma-3's own vectors, the mapped consensus vectors, and vectors transferred from each core model separately. Figure~\ref{fig:gemma-generalization}c,d and Table~\ref{tab:gemma-steer} show the comparisons. For 12B at $s=+0.1$, the original vector raises the matched emotion's log-probability by $3.66$. Neutral-face and ImageNet consensus vectors give $3.59$ and $3.35$. Competing-emotion means are much smaller, although the mapped consensus does not suppress them on average. Negative injection lowers the matched emotion and raises competing emotions.

Under each alignment condition, the 12B consensus vectors produce larger positive matched-emotion effects than vectors transferred from any individual core model. For neutral-face alignment, the individual transfers give $2.66$--$2.90$, despite higher cosine with Gemma-3's own vectors ($0.70$--$0.75$) than the consensus ($0.63$). ImageNet alignment shows the same ordering of positive effects. These comparisons again show that similarity to the target model's vector does not fully predict steering strength.

\begin{table}[t]
\centering
\small
\setlength{\tabcolsep}{5pt}
\caption{Gemma-3 steering on AffectNet at image-token positions. Cosine compares mapped vectors with Gemma-3's own AffectNet vectors. Consensus construction excludes Gemma-3. Mapping uses paired activations without emotion-vector fitting. diag and off are mean $\Delta\log p$ for the injected emotion and the other five emotions.}
\label{tab:gemma-steer}
\begin{tabular}{@{}lrrrrr@{}}
\toprule
 &  & \multicolumn{2}{c}{$s=-0.1$} & \multicolumn{2}{c}{$s=+0.1$} \\
\cmidrule(lr){3-4}\cmidrule(lr){5-6}
Injected vector & Cosine & diag & off & diag & off \\
\midrule
\multicolumn{6}{l}{\emph{Gemma-3-12B}} \\
Original Gemma-3 vector & $1.00$ & $-2.45$ & $+1.02$ & $+3.66$ & $-0.02$ \\
\addlinespace
\multicolumn{6}{l}{\emph{Neutral-face alignment}} \\
Consensus vector & $0.63$ & $-1.84$ & $+0.89$ & $+3.59$ & $+0.18$ \\
Qwen3-VL vector & $0.75$ & $-1.69$ & $+0.82$ & $+2.90$ & $+0.00$ \\
Ministral-3 vector & $0.75$ & $-1.53$ & $+0.82$ & $+2.66$ & $+0.08$ \\
InternVL3.5 vector & $0.70$ & $-1.59$ & $+0.78$ & $+2.69$ & $+0.06$ \\
\addlinespace
\multicolumn{6}{l}{\emph{ImageNet alignment}} \\
Consensus vector & $0.42$ & $-2.09$ & $+0.88$ & $+3.35$ & $+0.11$ \\
Qwen3-VL vector & $0.46$ & $-1.38$ & $+0.68$ & $+2.15$ & $+0.06$ \\
Ministral-3 vector & $0.50$ & $-1.08$ & $+0.77$ & $+2.15$ & $+0.25$ \\
InternVL3.5 vector & $0.37$ & $-1.14$ & $+0.55$ & $+1.67$ & $+0.13$ \\
\addlinespace
\multicolumn{6}{l}{\emph{Null controls}} \\
Random directions & --- & $+0.47$ & $+0.67$ & $-0.03$ & $+0.28$ \\
Orthogonal directions & --- & $+0.43$ & $+0.67$ & $+0.01$ & $+0.28$ \\
\midrule
\multicolumn{6}{l}{\emph{Gemma-3-27B}} \\
Original Gemma-3 vector & $1.00$ & $-2.09$ & $+0.87$ & $+4.04$ & $-0.06$ \\
\addlinespace
\multicolumn{6}{l}{\emph{Neutral-face alignment}} \\
Consensus vector & $0.59$ & $-1.84$ & $+0.75$ & $+3.47$ & $+0.08$ \\
Qwen3-VL vector & $0.73$ & $-1.80$ & $+0.54$ & $+3.01$ & $+0.00$ \\
Ministral-3 vector & $0.72$ & $-1.54$ & $+0.43$ & $+2.39$ & $-0.02$ \\
InternVL3.5 vector & $0.66$ & $-1.53$ & $+0.59$ & $+2.57$ & $+0.02$ \\
\addlinespace
\multicolumn{6}{l}{\emph{ImageNet alignment}} \\
Consensus vector & $0.39$ & $-1.80$ & $+0.71$ & $+3.16$ & $+0.00$ \\
Qwen3-VL vector & $0.42$ & $-1.37$ & $+0.43$ & $+2.14$ & $+0.01$ \\
Ministral-3 vector & $0.43$ & $-1.36$ & $+0.25$ & $+1.67$ & $+0.01$ \\
InternVL3.5 vector & $0.34$ & $-1.11$ & $+0.39$ & $+1.72$ & $-0.08$ \\
\addlinespace
\multicolumn{6}{l}{\emph{Null controls}} \\
Random directions & --- & $+0.24$ & $+0.56$ & $+0.39$ & $+0.35$ \\
Orthogonal directions & --- & $+0.26$ & $+0.56$ & $+0.36$ & $+0.37$ \\
\bottomrule
\end{tabular}
\end{table}

\paragraph{The model-size ablation preserves the causal-transfer pattern.}
We repeat the causal comparison in Gemma-3-27B to test whether the transfer pattern holds at a larger model size. At $s=+0.1$, the original vector gives $(\mathrm{diag},\mathrm{off})=(+4.04,-0.06)$, while the consensus gives $(+3.47,+0.08)$ with neutral-face mapping and $(+3.16,+0.00)$ with ImageNet mapping. Both consensus variants again have larger positive matched-emotion effects than the corresponding single-model transfers. Negative injection reverses the matched-emotion effect, while the random and orthogonal controls lack this selective response. Thus, correspondence and causal transfer persist at both sizes. Because width, depth, and the evaluation subsets also differ, this comparison tests robustness across sizes rather than isolating parameter count alone.

\section{Limitations, Future Work, and Broader Impact}
\label{app:lim}

\subsection{Limitations}
\label{app:limitations}

Although we find shared structure and transferable causal effects in emotion representations across sources, modalities, and architectures, our study has several limitations. First, the core analyses focus on three approximately $8$B-scale VLMs. Although the held-out Gemma-3 experiments extend the analysis to 12B and 27B, generalization to substantially larger models and more diverse model families remains to be tested. Second, we focus on Ekman's six basic emotions because they provide a common emotion set across the image sources used here. The observed organization remains sensitive to stimulus source and may differ for more complex emotions, continuous affective dimensions, or emotion categories defined in different cultural contexts. Finally, this work focuses on understanding how VLMs represent and use emotion concepts. We do not test whether these findings can be used to improve task performance or reduce emotion-related hallucinations.

\subsection{Future Work}
\label{app:future}

In future work, we plan to test whether the relationships among emotion vectors and their causal effects extend to larger VLMs and more diverse model families. We also aim to extend the analysis to video and speech in multimodal large language models (MLLMs). Beyond the six basic emotions, we plan to study more complex emotions, continuous affective labels, and data from different cultural contexts. By comparing a broader range of stimulus sources, we hope to better understand which input properties drive differences in emotion representations. We will also explore whether the emotion vectors and shared subspace can improve emotion understanding or reduce emotion-related hallucinations. To evaluate these applications, we will measure changes in task accuracy and hallucination rates and check for unwanted effects on unrelated tasks.

\subsection{Broader Impact and Ethics}
\label{app:broader}

We study internal emotion representations in VLMs, without claiming to infer a person's true emotional state from facial appearance. Our emotion categories define controlled conditions for studying model representations. They do not provide a complete account of human affect. Model behavior can depend on dataset composition, cultural context, and stimulus source. Using these systems in consequential settings could therefore amplify existing biases or encourage overconfident interpretations of human emotion. These steering techniques could also be used to manipulate affect-related outputs. We therefore view them primarily as tools for understanding and auditing multimodal models and encourage careful validation, privacy protection, and human oversight in downstream applications.

\section{Ethics Statement}
\label{app:ethics}

This study explores emotion representations and their causal effects in VLMs using existing datasets and synthetic stimuli. We use existing facial-expression datasets rather than collecting new data from human participants. However, real face images raise privacy concerns even when they come from established research datasets. We therefore do not redistribute AffectNet or RAF-DB images, and access to these data remains subject to the original providers' terms.

Emotion labels provide controlled categories for our experiments, but they do not establish a person's true emotional state or capture the full range of human emotion. Cultural and demographic differences may also affect both dataset labels and model behavior. Our findings therefore should not be taken as evidence that these models can reliably assess individuals in sensitive settings such as mental health, education, or employment. In addition, emotion steering could be used to manipulate model outputs. We discuss these risks and the need for privacy protection, careful validation, and human oversight in Appendix~\ref{app:broader}.

\section{Reproducibility Statement}
\label{app:reproducibility}

Appendix~\ref{sec:setup} documents the model configurations, stimulus sources and splits, generation prompts, activation extraction, emotion-vector construction, steering protocol, alignment procedures, and similarity measures. Appendices~\ref{app:sim}--\ref{app:demix} provide detailed results and controls for the main analyses. We will release the code, experiment configurations, stimulus sample lists, synthetic-generation seeds, and extracted emotion vectors described in Appendix~\ref{app:compute}. For datasets that cannot be redistributed, we will provide sample identifiers so that researchers with access can reconstruct the experimental subsets.

\section{LLM Usage Statement}
\label{app:LLM}

We used LLMs to assist with language polishing. In addition, we used generative models to create the synthetic stories, synthetic facial portraits, and synthetic emotion-evoking scenes in CMES. The data sources, generation models, and prompts are documented in Appendices~\ref{app:stimuli} and~\ref{app:prompts}.

\end{document}